\documentclass[journal]{IEEEtran}
\usepackage{graphics,graphicx}
\usepackage{amsmath,amssymb,amsthm}
\newtheorem{myDef}{Definition}
\usepackage{multirow}
\usepackage{cases}
\usepackage{array}
\usepackage{algorithm,algorithmic}
\usepackage{color,colortbl,xcolor}
\usepackage[graphicx]{realboxes}
\usepackage{booktabs,url}
\usepackage[tight,footnotesize,sf,SF]{subfigure}
\usepackage{bigstrut}

\newcommand{\fref}[1]{Fig. \ref{#1}}
\newcommand{\sref}[1]{Section \ref{#1}}
\newcommand{\tref}[1]{TABLE \ref{#1}}
\newcommand{\eref}[1]{Eq. (\ref{#1})}

\begin{document}
\title{Multimodal Multi-objective Optimization: Comparative Study of the State-of-the-Art}

\author{Wenhua Li,
Tao Zhang,
Rui Wang, \emph{Senior Member, IEEE},
Shengjun Huang,
and Jing Liang \emph{Senior Member, IEEE}

\thanks{This work was supported by the National Science Fund for Outstanding Young Scholars (62122093), the National Natural Science Foundation of China (72071205, 71901210) and the Scientific Key Research Project of the National University of Defense Technology (ZZKY-ZX-11-04).}
\thanks{Wenhua Li is with the College of Systems Engineering, National University of Defense Technology, Changsha, China, 410073, e-mail: liwenhua@nudt.edu.cn.}
\thanks{Tao Zhang, Rui Wang and Shengjun Huang are with the College of Systems Engineering, National University of Defense Technology, Changsha, China, and Hunan Key Laboratory of Multi-energy System Intelligent Interconnection Technology, Changsha, 410073, China}
\thanks{Jing Liang is the School of Electrical Engineering, Zhengzhou University, Zhengzhou, China}
\thanks{Corresponding author: Rui Wang (email: ruiwangnudt@gmail.com). }}
\maketitle

\begin{abstract}
Multimodal multi-objective problems (MMOPs) commonly arise in real-world problems where distant solutions in decision space correspond to very similar objective values. To obtain more Pareto optimal solutions for MMOPs, many multimodal multi-objective evolutionary algorithms (MMEAs) have been proposed. For now, few studies have encompassed most of the recently proposed representative MMEAs and made a comparative comparison. In this study, we first review the related works during the last two decades. Then, we choose 12 state-of-the-art algorithms that utilize different diversity-maintaining techniques and compared their performance on existing test suites. Experimental results indicate the strengths and weaknesses of different techniques on different types of MMOPs, thus providing guidance on how to select/design MMEAs in specific scenarios. 
\end{abstract}

\begin{IEEEkeywords}
Multimodal multi-objective optimization, Evolutionary computation, Comparative study, Review
\end{IEEEkeywords}

\IEEEpeerreviewmaketitle

\section{Introduction}
\label{sec_introduction}
Many real-world engineering problems consider optimizing more than one objective. In general, there is a conflict between objectives which means no solution can obtain the best performance on all objectives. Such problems are recognized as multi-objective optimization problems (MOPs) \cite{deb2014multi,li2020reinvestigation,li2019multi}. Without loss of generality, an MOP can be expressed as follows:

\begin{equation}
\begin{gathered}
\text{Minimize}~~F(\mathbf{x}) =\{f_1(\mathbf{x}),f_2(\mathbf{x}),\cdots, f_m(\mathbf{x}) \}, \\
s.t.~~~~ \mathbf{x} = (x_1,x_2,\ldots, x_n) \in \Omega,
\end{gathered}
\label{eq_MOP}
\end{equation}
where $\Omega$ denotes the search space, $m$ is the number of objectives, and $\mathbf{x}$ is a decision vector that consists of $n$ decision variables $x_i$. A solution, $\mathbf{x_a}$, is considered to Pareto dominate another solution, $\mathbf{x_b}$, $iff$ $\forall i=1,2,...,m, f_i(\mathbf{x_a}) \leq f_i(\mathbf{x_b})$ and $\exists j=1,2,...,m, f_j(\mathbf{x_a}) < f_j(\mathbf{x_b})$. Furthermore, a Pareto optimal solution is a solution that is not Pareto dominated by any other solution. The set of Pareto optimal solutions is called a Pareto set (PS). The image of the PS is known as the Pareto front (PF).

To address MOPs, many multi-objective evolutionary algorithms (MOEAs) have been proposed and verified over many wide-acceptable benchmark problems. In general, the aim of MOEAs is to obtain a solution set that approximates the known true Pareto front most. This aim contains two parts: the convergence to the true Pareto front and obtaining an evenly distributing solution set. In order to address these issues, most of the existing MOEAs adopt the convergence-first strategy and the crowding-distance-based second-selection approach to select a new population after offspring generation, which is also known as the environmental selection strategy. 

\begin{figure}[tbph]
	\begin{center}
		\includegraphics[width=3in]{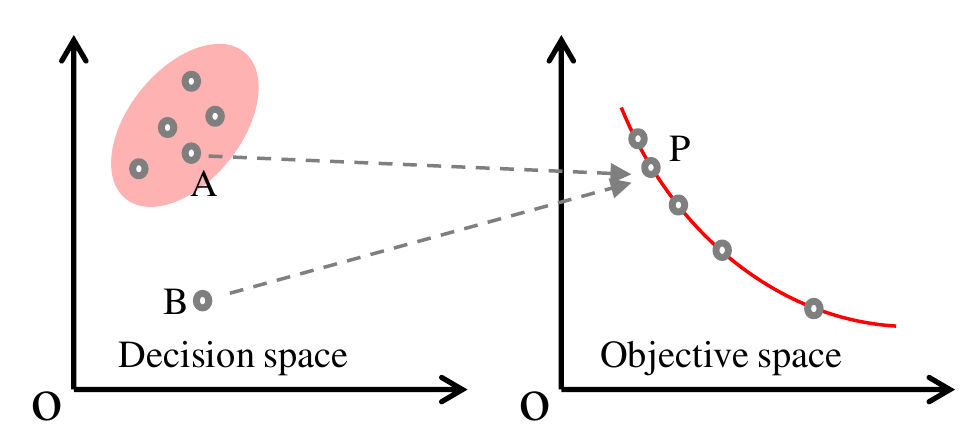}
		\caption{Illustration of MMOPs and the difficulties in solving MMOPs.}
		\label{fig_mmop}
	\end{center}
\end{figure}

In real-world engineering optimization, there arises a kind of MOP in that multiple different solutions share the same or similar objective values, termed multimodal multi-objective problems (MMOPs). \fref{fig_mmop} shows a two-variable two-objective MMOP, where $A$ and $B$ are distant in the decision space but share the same objective values. The aim of solving MMOPs is to obtain as many Pareto optimal (global and local) solutions as possible. The benefits of solving MMOPs are listed as follows: (1) Finding multiple local or global optimal solutions can help reveal the underlying nature of the problem, thus helping decision-makers (DMs) better understand the problem and conduct analysis of the problem. (2) Multiple alternative solutions can provide DMs with more choices. For manufacturers, multiple alternative solutions mean multiple different scenarios. (3) Finding multiple alternative solutions helps to find robust solutions. (4) Fast switching between multiple candidate solutions helps to solve dynamic optimization problems. (5) Retaining multiple optimal solutions can increase the diversity of solutions and help the algorithm jump out of the local optimal area.

\begin{figure}[tbph]
	\begin{center}
		\includegraphics[width=3in]{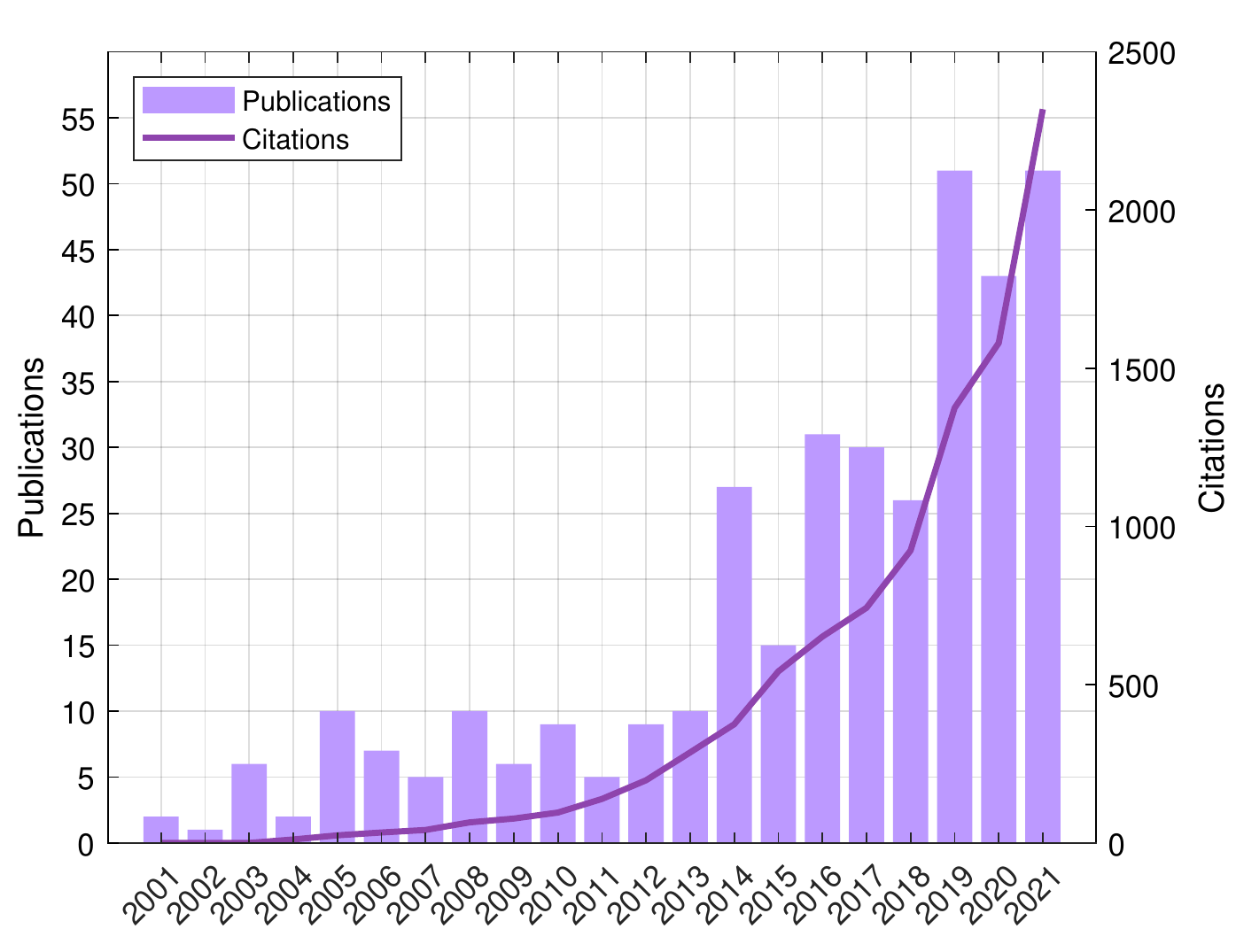}
		\caption{Times cited and publications of multimodal multi-objective optimization over time from the Web of Science Core Collection during 2001 to 2021.}
		\label{fig_trends}
	\end{center}
\end{figure}

\fref{fig_trends} shows the research trends of research in MMOPs from 2001 to 2021, from which we can see that research in this field raises more and more attention. Many approaches have been proposed in the most recent three years. Compared to MOPs, MMOPs are much more challenging. To be specific, for traditional MOEAs, the convergence-first strategy weakens the diversity of solutions in the decision space and thus dents the exploration ability of the MOEA. For the problem with more than one global PSs, the PS locates on the steep landscape is likely to be removed during the evolution. Another challenge is to balance the distribution of solutions both in the objective and decision spaces. Overall, traditional MOEAs face serious challenges in solving MMOPs, and developing novel approaches is important for this field.

Over the last two decades, a number of multi-objective multimodal evolutionary algorithms (MMEAs) have been proposed to solve MMOPs. Most of the primitive MMEAs perform the environmental selection by calculating the crowding distance in the decision space to enhance the diversity of solutions, e.g., alternative crowding distance used in Omni-optimizer \cite{deb2005omni,deb2008omni}. Another popular way is to utilize the niching mechanism, e.g., double-niched evolutionary algorithm (DNEA) \cite{liu2018double}, decision space-based niching NSGA-II (DN-NSGAII) \cite{liang2016multimodal} and multi-objective particle swarm optimization using ring topology and special crowding distance (MO\_Ring\_PSO\_SCD) \cite{yue2017multiobjective}. Such methods have been proved effective in the previously proposed MMOP test suites like MMF \cite{yue2017multiobjective} problems. In \cite{liu2019handling}, Liu et.al proposed the IDMP test suites, which are harder to be solved since obtaining different PSs needs different numbers of function evaluations. Experimental results show that the previous works are unable to obtain all PSs on IDMPs. Then, many enhanced diversity-maintaining strategies were proposed, e.g., the convergence-penalized density (CPD) method in CPDEA \cite{liu2019handling}, weighted-indicator in MMEA-WI \cite{li2021weighted} and zoning search with adaptive resource allocating (ZS-ARA) method \cite{fan2021zoning}. Moreover, it's reported in \cite{liang2019problem,li2022hierarchy} that MMOPs with local Pareto fronts (MMOPLs) are more common in real-world problems and normal MMOPs are special cases of MMOPLs. To address this issue, several approaches were proposed to solved MMOPLs, like $\epsilon$-dominance in $P_{Q,\epsilon}$-MOEA \cite{schutze2011computing}, multifront archive update method in DNEA-L \cite{liu2019searching}, dual clustering methods in MMOEA/DC \cite{lin2020multimodal} and hierarchy ranking method in HREA \cite{li2022hierarchy}.

In 2019, Tanabe et al. \cite{tanabe2019review} made a review of evolutionary multimodal multi-objective optimization and discussed several open issues about the existing test suites and performance metrics. In addition, Yue et al. \cite{yue2021review} made a recent review that mainly focuses on difficulties in dealing with MMOPs and properties of the existing diversity-maintaining methods, which was written in Chinese. As we can see from \fref{fig_trends}, many pieces of research have been conducted during the last three years. Moreover, to our best knowledge, there is no work that systematically made a comparative comparison of the existing MMEAs. To this end, in this paper, we first review existing MMOPs test suites for MMOPs and discuss their features. Then, we choose 12 representative MMEAs with different diversity-maintaining mechanisms and compare their performances on different test suites in detail. These comprehensive comparative results show not only the advantages and disadvantages of different diversity-maintaining techniques but also the challenges that existed in different test suites. The contributions of this work can be summarized as follows:

\begin{itemize}
\item Motivated by previous works, we conclude a more accurate and comprehensive definition of the multimodal multi-objective optimization problem.
\item A comprehensive review of the existing MMOP test suites is conducted. We compare their properties and make a discussion about the open issues and future studies.
\item We first conduct a comparative study on the existing MMEAs. 12 state-of-the-art MMEAs and 53 test problems (in four groups) are chosen for comparison. The performances in terms of $IGD$, $IGDX$, and $1/PSP$ are analyzed. The diversity-maintaining techniques, searching behaviors, and computational complexity are studied.
\item Based on the experimental results, the strengths and weaknesses of different techniques, limitations of the indicators, suggestions for further studies, and other conclusions are discussed in this work. Researchers can find suggestions for selecting and designing MMEAs.
\end{itemize}

The rest of this paper is organized as follows. \sref{sec_review} describes existing MMOPs test suites and performance metrics for MMEAs, followed by \sref{sec_mmea} that introduces 12 state-of-the-art MMEAs in detail. \sref{sec_setting} illustrates the experimental settings, followed by \sref{sec_result}, which analyses the experimental results in detail. In \sref{sec_discuss}, we further show the overall performance comparison results of the 12 algorithms over all the test suites and analyze the limitations of existing test suites. \sref{sec_conclusion} presents our conclusions and some possible paths for future research.

\section{Preliminarily study}
\label{sec_review}
It's reported in \cite{tanabe2019review} that, although MMOPs have been addressed for more than ten years, the definition of an MMOP is still controversial. By combining definitions from the existing work, we define an MMOP as follows:
\begin{myDef}
\label{def_1}
For a Pareto optimal solution $\mathbf{x}$ of an MOP, if there exists a distant solution $\mathbf{y}$ satisfying $\left \| f(\mathbf{x})-f(\mathbf{y}) \right \| \leq \delta$ ($\delta$ is a small positive value), then the MOP is called as an MMOP.
\end{myDef}
\begin{myDef}
\label{def_2}
Two solutions $\mathbf{x}$ and $\mathbf{y}$ are said as distant solutions if $\left \| \mathbf{x}-\mathbf{y} \right \| \geq \theta$ ($\theta$ is a positive value provided by decision maker).
\end{myDef}

To be specific, most primitive works consider the situation when $\delta=0$, which means there is no local PS but several global PSs corresponding to the same global PF. $\delta>0$ indicates that there exists local PS and local PF, termed multimodal multi-objective problems with local Pareto fronts (MMOPLs) \cite{li2022hierarchy}. 

\subsection{Benchamrk test suits}
\label{sec_testsuite}
To examine the performance of MMEAs, a number of MMOPs test suites have been designed. Deb \cite{deb1999multi} proposed Omni-test problems, where the number of decision variables and the number of PSs is adjustable. After that, SYM-PART test suites \cite{rudolph2007capabilities} were proposed by rotating and distortion operation. In addition, TWO-ON-ONE \cite{preuss2006pareto}, SSUF problems \cite{liang2016multimodal}, HPS problems \cite{zhang2017benchmark} and Polygon problems \cite{ishibuchi2011many} are also proposed to exam the diversity-maintaining performance of algorithms. However, the number of decision variables is small and not scalable for Polygon problems. To this end, the Multi-polygon problems \cite{ishibuchi2019scalable,peng2020decomposition} are proposed with adjustable dimensions of decision variables and objectives.

In 2017, Yue et al. \cite{yue2017multiobjective} designed eight MMF test problems by translation and symmetry operation, which are widely adopted as the benchmark problems for MMOPs. Based on the same idea, a novel scalable test suite is proposed \cite{yue2019novel} that contains both local and global PSs, which is also used as the benchmark for the 2019 IEEE CEC multimodal multi-objective optimization competition \cite{liang2019problem}. In 2018, Liu et al. \cite{liu2018multi} designed the MMMOP test suite, which is scalable both in the objective and decision space. In 2021, Tian et al. \cite{tian2021multi} proposed a large-scale sparse MMOP test suite motivated by the optimal architecture design problem of the convolutional neural network (CNN).

Most of the above-mentioned test suites assume that the difficulties in searching for different PSs are the same. However, this may not be true for real-world problems, where the landscape may be complex and irregular. To this end, Liu et al. \cite{liu2019handling} designed the IDMP test suite with 2-4 objectives and variables. The main property of IDMP is that for a point on the PF, solutions close to one equivalent Pareto optimal solution are more likely to dominate solutions close to another equivalent Pareto optimal solution. Experimental results show that the primitive MMEAs are unable to obtain all PSs on IDMP test problems. Based on IDMP, Li et al. \cite{li2022hierarchy} designed the IDMP\_e test suites by introducing other single-objective multimodal functions, which contain several local PSs and PFs with adjustable quality.

As for the real-world problems, Ishibuchi et al. \cite{ishibuchi2011many} generated a multimodal four-objective location planning problem from a real-world map considering the distances to the nearest elementary school, junior high school, railway station, and convenience store. However, there are only two decision variables. In addition, Yue et al. \cite{yue2019multimodal} found that the multi-objective feature selection problem is a typical MMOP. Moreover, Liang et al. \cite{liang2020problem} proposed a multimodal multiobjective path planning test suite and launch a competition \cite{yao2022multimodal,jin2021multi}. However, since the encoded length of a path is discrete and indeterminate, the existing MMEAs can not be directly used to solve these problems.

\begin{table}
\caption{Properties of MMOP test suites, where D, M, and P denote the number of decision variables, objectives, and equivalent PSs, respectively. Local indicates if there are local PSs.}
\begin{tabular}{lccccc}
\hline
\textbf{Test Suites} & \textbf{D} & \textbf{M} & \textbf{P} & \textbf{Local} & \textbf{PF shape} \bigstrut\\
\hline
\textbf{Omni-test \cite{deb1999multi}} & Any   & 2     & Any   & No    & Convex \\
\textbf{SYM-PART \cite{rudolph2007capabilities}} & 2     & 2     & 9     & No    & Convex \\
\textbf{TWO-ON-ONE \cite{preuss2006pareto}} & 2     & 2     & 2     & No    & Convex \\
\textbf{Polygon \cite{ishibuchi2011many}} & 2     & Any   & Any   & No    & Convex \\
\textbf{Multi-polygon \cite{ishibuchi2019scalable}} & Any   & Any   & Any   & No    & Convex \\
\textbf{MMF \cite{yue2017multiobjective}} & 2     & 2     & 2     & No    & Both \\
\textbf{Problems in \cite{yue2019novel}} & Any   & Any   & Any   & Yes   & Both \\
\textbf{MMMOP \cite{liu2018multi}} & Any   & 2   & 4   & No    & Concave \\
\textbf{SMMOP \cite{tian2021multi}} & Any   & Any   & Any   & No    & Convex \\
\textbf{IDMP \cite{liu2019handling}} & 2-4   & 2-4   & 2-4   & No    & Convex \\
\textbf{IDMP\_e \cite{li2022hierarchy}} & 2-3   & 2-3   & Any   & Yes   & Convex \\
\hline
\textbf{Location planning \cite{ishibuchi2011many}} & 2   & 4   & -   & No   & Convex \\
\textbf{Feature selection \cite{yue2019multimodal}} & Any   & 2   & -   & Yes   & Convex \\
\textbf{Path planning \cite{liang2020problem}} & -   & 2-7   & -   & No   & Convex \\
\hline
\end{tabular}%
\label{tab_testsuites}
\end{table}

Generally speaking, a comprehensive MMOP test suite should have the following properties: The dimension of the decision space can be extended, the number of objectives can be adjusted, the true PSs and PFs are known, the PSs and PFs have various shapes, the number of PSs is scalable, and the local PSs and the global PSs coexist. To be specific, the overall properties of the existing MMOP test suites are listed in \tref{tab_testsuites}. As we can see, the test suite proposed in \cite{yue2019novel} is considered a comprehensive benchmark for now. It's scalable both in objectives and decision variables with local PSs. However, many existing test problems are relatively simple and the required function evaluations for these problems are small compared to real-world problems. For the existing MMEAs, there is no suitable benchmark problem that can both examine the abilities on maintaining the diversity and converging to the PF. In addition, many real-world problems are discrete or mix-integer. It's hard to check the performance of the existing MMEAs on discrete optimization problems for now.

\subsection{Performance metrics}
\label{sec_metrics}
Different from traditional MOPs, the aim of solving MMOPs is to obtain as many Pareto optimal solutions as possible. Thus, the diversity of solutions in the decision space is important. To this end, motivated by Inverted Generation Distance ($IGD$) \cite{zitzler2003performance}, the Inverted Generation Distance in the Decision Space ($IGDX$) \cite{zhou2009approximating} is proposed. Specifically, for an obtained solution set $\mathbf{X}$, the $IGD$ and $IGDX$ can be calculated as:

\begin{equation}
IGD(\mathbf{X})=\frac{1}{|\mathbf{X^*} |} \sum_{\mathbf{y} \in \mathbf{X^*}} 
\min_{\mathbf{x} \in \mathbf{X}} \{ED(f(\mathbf{x}),f(\mathbf{y})) \},
\end{equation}
\begin{equation}
IGDX(\mathbf{X})=\frac{1}{|\mathbf{X^*} |} \sum_{\mathbf{y} \in \mathbf{X^*}} 
\min_{\mathbf{x} \in \mathbf{X}} \{ED(\mathbf{x},\mathbf{y}) \},
\end{equation}
where $ED(\mathbf{x},\mathbf{y})$ is the Euclidean distance between $\mathbf{x}$ and $\mathbf{y}$. $\mathbf{X}$ and $\mathbf{X^*}$ denote the obtained solution set and a set of a finite number of Pareto optimal solutions uniformly sampled from the true PS, respectively.

$IGDX$ is a representative metric that evaluates the performance of an MMEA in finding solutions with high quality and finding diverse solutions in the decision space. However, $IGDX$ can not measure the diversity in the objective space. Thus, many works adopted to use both $IGD$ and $IGDX$ to overall evaluate the performance of an MMEA. In addition, Cover Rate ($CR$) and Pareto Sets Proximity ($PSP$) \cite{yue2017multiobjective} are proposed to reflect the similarity between the obtained PSs and the true PSs, where $CR$ is a modification of the maximum spread ($MS$) \cite{tang2012hybrid}, $PSP$ is a combination of $CR$ and $IGDX$, which can be expressed as:

\begin{equation}
\label{equ_psp}
PSP(\mathbf{X}) = \frac{CR(\mathbf{X})}{IGDX(\mathbf{X})},
\end{equation}
\begin{equation}
CR(\mathbf{X})=(\prod_{i=1}^D \sigma_i)^{\frac{1}{2D}},
\end{equation}
\begin{equation}
\sigma _i=(\frac{\text{min}(x_i^{*,max},x_i^{max})- 
\text{max}(x_i^{*,min},x_i^{min})}{x_i^{*,max}-x_i^{*,min}})^2,
\end{equation}
where $x_i^{*,max}$ and $x_i^{max}$ are the maximum values of the $i$-th variable in the PS and the obtained solutions respectively. $\sigma _i=1$ when $x_i^{*,max}= x_i^{*,min}$; $\sigma _i=0$ when 
$x_i^{*,max} \leq x_i^{min}$ or $x_i^{max} \leq x_i^{*,min}$.

The larger $PSP$ is the better performance of the solution set. Since the best $PSP$ value is infinite large, it's hard to evaluate the distance between the evaluated solution set and the reference solution set. Then in \cite{yue2019novel}, $1/PSP$ is used as a new indicator.

The above-mentioned metrics evaluate the solutions' quality only in the decision space. Inverted Generational Distance-Multi-modal ($IGDM$) \cite{liu2018multi} is proposed,  which can measure not only the convergence performance but also the diversity performances both in the objective and the decision spaces. However, it needs a parameter defined by the user.

To sum up, the convergence and diversity quality in the decision space of a solution set can be well evaluated by $IGDX$ and $PSP$. However, the quality in the objective space is not evaluated properly. Thus, $IGD$ and HyperVolume ($HV$) are also popular indicators for evaluating the performance of MMEAs. For now, the existing performance metrics need the information of the true PF and PS, which is hard for real-world problems. Moreover, a parameter-free indicator that can measure the diversity and convergence quality of a solution set both in the objective and the decision space is needed.

\section{Compared multimodal multi-objective evolutionary algorithms}
\label{sec_mmea}
In this part, the detail information of twelve chosen MMEAs are introduced, namely, Omni-optimizer \cite{deb2005omni,deb2008omni}, DN-NSGAII \cite{liang2016multimodal}, MO\_Ring\_PSO\_SCD \cite{yue2017multiobjective}, MO\_PSO\_MM \cite{Jing2018A}, DNEA \cite{liu2018double}, Tri-MOEA\&TAR \cite{liu2018multi}, DNEA-L \cite{liu2019searching}, CPDEA \cite{liu2019handling}, MP-MMEA \cite{tian2021multi}, MMOEA/DC \cite{lin2020multimodal}, MMEA-WI \cite{li2021weighted} and HREA \cite{li2022hierarchy}.

\textbf{Omni-optimizer} is considered the most representative MMEA. It's proposed to obtain multi-optima both for single and multi-objective problems. There are several strategies utilized in Omni-optimizer. First, the Latin hypercube is used to uniformly generate the initial population. Second, a nearest neighbor-based strategy is proposed to choose two individuals that take part in tournament selection (known as restricted selection). Third, a two-tier fitness assignment scheme is adopted in which the primary fitness is computed using the phenotypes (objectives and constraint values). And the secondary fitness is computed using both phenotypes and genotypes (decision variables).

\textbf{DN-NSGAII} introduces a niching method and a selection operator to create the mating pool and select offspring, respectively. The procedure of the niching method can be roughly described as follow. A solution and a constant number of solutions are randomly chosen. Then, the current solution and the solution with the smallest Euclidean distance to the current solution are selected. Repeat the above steps until the mating pool is full. In the original paper, the authors also proposed SS-UF1 and S-UF3 test problems.

\textbf{MO\_Ring\_PSO\_SCD} is another representative MMEA that received much attention. The personal best archive (PBA) and the neighborhood best archive (NBA) are first established. Then, ring topology \cite{li2009niching} is used to induce multiple niches. In addition, motivated by Omni-optimizer, special crowding distance (SCD) is proposed to maintain the diversity of solutions both in the decision and objective space. In this paper, the MMF test problems and the $PSP$ indicator are proposed to examine the performance of MMEAs.

\textbf{MO\_PSO\_MM} introduced a self-organizing mechanism, which is updated simultaneously during the evolution, to find the distribution structure of the population and build the neighborhood in the decision space. Then, the solutions which are similar to each other can be mapped into the same neighborhood. In addition, special crowding distance \cite{yue2017multiobjective} is utilized to maintain the diversity of solutions. The effectiveness of MO\_PSO\_MM is mainly verified by the MMF test suite.

\textbf{DNEA} introduced a niche sharing method to diversify the solution set in both the objective and decision spaces. Through the double-niching method, a solution that is very close to others in the objective (decision) space but far away from others in the decision (objective) space still has a chance to be selected. The performance and behavior of DNEA were verified by Polygon-based problems.

\textbf{Tri-MOEA\&TAR} proposed to use two archives (convergence and diversity archives) and the recombination strategy to solve MMOPs. To be specific, the decision variable analysis method is first performed to find the convergence-related and diversity-related decision variables. Then, this information is passed to the two archives to ensure better convergence and diversity quality. Finally, the recombination strategy is used to obtain a large number of multiple Pareto optimal solutions. The MMMOP test suite was proposed in this work.

\textbf{DNEA-L} is proposed to solve MMOPLs based on DNEA \cite{liu2018double}. A multi-front archive update method is utilized to obtain both global and local PSs. To be specific, a neighborhood size is first calculated. Then, solutions dominated by their neighbors are removed. The sizes of the first $K$ fronts are maintained by the double-sharing function proposed in DNEA. The effectiveness of DNEA-L is verified by Polygon-based problems with local Pareto optima.

\textbf{CPDEA} is the first work that considers the imbalanced searching difficulties for different PSs. Convergence-penalized density method is proposed to help explore the whole decision space during the evolution. Thus, the population will not prematurely converge to some equivalent Pareto optimal solutions which are easy to find. In addition, the double $k$-nearest neighbor method is utilized to further improve the distribution both in the decision and objective space. Moreover, the IDMP test suite was proposed to better examine the ability of MMEAs in maintaining diversity.

\textbf{MP-MMEA} first considers solving large-scale MMOPs. Multiple subpopulations are introduced to obtain equivalent PSs, which are maintained according to the proposed subpopulation similarities. Thus, for each subpopulation, the diversity in the decision space does not need to be measured. An adaptively updated guiding vector is utilized to distinguish the search directions of different subpopulations. In addition, the proposed benchmark problems SMMOP1-SMMOP8 are adopted to verify the performance of MP-MMEA.

\textbf{MMOEA/DC} considered obtaining both global and local PSs by utilizing the double clustering method, namely, the neighborhood-based clustering method (NCM) in the decision space and the hierarchical clustering method (HCM) in the objective space. In addition, the harmonic averaged distance is utilized to evaluate the crowding distance of solutions in the decision space. The performance of MMOEA/DC is verified by the novel MMF test suites \cite{yue2019novel}.

\textbf{MMEA-WI} adopted a weighted-indicator to evaluate the potential convergence quality of a solution, which is derived from IBEA \cite{phan2013r2}. The weighted-indicator for a solution is calculated by summing the fitness of other solutions according to the distance. Then, solutions will gather around the PS. In addition, a convergence archive is introduced to improve the convergence ability and maintain the uniformity of solutions. The effectiveness of MMEA-WI is mainly verified by IDMP test suites.


\textbf{HREA} adopted a local convergence indicator to evaluate the local convergence of solutions. Then, both global and local PSs can be preserved during the evolution. To control the quality of the obtained local PF, a hierarchy ranking method is proposed that can balance the convergence and diversity of solutions. In addition, based on IDMP, IDMP\_e is proposed which has the adjustable number of global and local PSs.

\begin{table*}[htb]
\centering
\caption{Overview of the compared twelve MMEAs (in chronological order)}
\begin{tabular}{llllp{32em}}
\hline
\textbf{Algorithm} & \textbf{Framework} & \textbf{Local} & \multicolumn{1}{m{32em}}{\textbf{Mechanism/Strategy}} \\
\hline

\textbf{Omni-optimizer \cite{deb2005omni}} & GA    & No    & \multicolumn{1}{m{32em}}{Latin hypercube sampling-based population; restricted mating selection and alternative crowding distance}  \\
\textbf{DN-NSGAII \cite{liang2016multimodal}} & GA    & No    & Niching in the decision space; crowding distance in the decision space \\
\textbf{MO\_Ring\_PSO\_SCD \cite{yue2017multiobjective}} & PSO   & No    & Ring topology; special crowding distance \\
\textbf{MO\_PSO\_MM \cite{Jing2018A}} & PSO   & No    & \multicolumn{1}{m{32em}}{Self-organizing method to find the neighborhood relation; special crowding distance} \\
\textbf{DNEA \cite{liu2018double}} & GA    & No    & Niche sharing method in both the objective and decision spaces \\
\textbf{Tri-MOEA\&TAR \cite{liu2018multi}} & Decompose-based & No   & \multicolumn{1}{m{32em}}{Diversity archive and convergence archive; clustering menthod and niching for objective and decision space respectively; decision variable analysis method} \\
\textbf{DNEA-L \cite{liu2019searching}} & GA    & Yes    & \multicolumn{1}{m{32em}}{Niche sharing method in both the objective and decision spaces; multi-front archive} \\
\textbf{CPDEA \cite{liu2019handling}} & One-by-one & No   & Convergence-penalized density method; double k-nearest neighbor method \\
\textbf{MP-MMEA \cite{tian2021multi}} & GA & No   & \multicolumn{1}{m{32em}}{Guide multiple subpopulations via adaptively updated guiding vectors; binary tournament selection; subpopulation similarity} \\
\textbf{MMOEA/DC \cite{lin2020multimodal}} & GA    & Yes    & \multicolumn{1}{m{32em}}{Neighborhood-based clustering method in the decision space; hierarchical clustering method in the objective space; harmonic averaged distance} \\
\textbf{MMEA-WI \cite{li2021weighted}} & Indicator-based & No   & Weighted-indicator; convergence archive \\
\textbf{HREA \cite{li2022hierarchy}} & GA    & Yes   & Local convergence quality; hierarchy ranking method; convergence archive \\
\hline
\end{tabular}%
\label{tab_mmea}
\end{table*}

Overall, the main differences and similarities of the twelve algorithms above can be summarized in \tref{tab_mmea}, which lists the framework and diversity-maintaining mechanisms of the selected MMEAs. As we can see, most of the MMEAs adopt crowding distance (or variants) in the decision space as the mating-selection and/or second-selection criteria. Genetic algorithm (GA) is the most used framework/optimizer, while particle swarm optimizer (PSO), differential evolution (DE), decomposed-based method, and indicator-based method are also successfully utilized. Primitive works considered more on the diversity in the decision space. However, since MMEA belongs to MOEA, it's also important to maintain the distribution of solutions in the objective space. Thus, more and more works take the crowding distance both in the decision and objective spaces into account.

\section{Setting of computational experiments}
\label{sec_setting}
\subsection{Benchmark and parameter specifications}
This work aims to comprehensively compare the performance of existing MMEAs. Therefore, according to our previous discussion, MMF (8 problems), IDMP (12 problems), Polygon-based problems (20 problems) and problems with local PFs (IDMP\_e and some of MMF, 13 problems) are selected as the test problems (53 problems in total). Specifically, the MMF test suite has complex PF shapes and is considered representative MMOPs, IDMP is used to test the diversity-maintaining ability since there are different difficulties in finding different PSs, and Polygon-based problems are adopted to examine the ability in solving many-objective and many-decision-variable MMOPs, where the number of objectives $M$ is set to 2, 3, 5, and 8. IDMP\_e and some of MMF are chosen to examine the MMEAs' ability in obtaining local PSs.

As for the algorithm's parameters, it's worth mentioning that, each algorithm has its preferred setting for common parameters (population size $N$ and function evaluations $FEs$), e.g, there is no need for Tri-MOEA\&TAR to set a large population size. Therefore, the performance of algorithms may vary on the parameter setting. For a widely-accepted and fair comparison, according to the previous works \cite{liang2019problem}, we set $N = 100*D$ and $FEs = 5000 * D$ respectively, where $D$ is the number of decision variables. For Polygon-based problems, we set $N$ to 200, 300, 400 and 400 when $M$ is 2, 3, 5 and 8, respectively. The simulated binary crossover (SBX) and polynomial mutation (PM) operators are employed to generate offspring except for MO\_Ring\_PSO\_SCD and MO\_PSO\_MM. In addition, other specific parameters are set according to the original papers, which are listed in \tref{tab_para}. It's worth mentioning that, experiments are conducted on a PC configured with an Intel i9-9900X @ 3.50 GHz and 64 G RAM. PlatEMO \cite{tian2017platemo} and framework proposed in IEEE CEC 2019 competition\footnote{The source codes and results can be found in \url{http://www5.zzu.edu.cn/ecilab/info/1036/1211.htm}} \cite{liang2019problem} are adopted. For the convenience of researchers in the multimodal multi-objective optimization field, we collect the existing source code of MMEAs and MMOP test suites and keep updating\footnote{The source code of algorithms and MMOP test suites used in this work can be access in \url{https://github.com/Wenhua-Li/ComparativeStudyofMMOP}}.

\begin{table}
\centering
\caption{Specific parameters setting for the compared twelve MMEAs}
\begin{tabular}{lp{18em}}
\hline
\textbf{Algorithm} & \textbf{Parameter setting} \\
\hline
\textbf{Omni-optimizer \cite{deb2005omni}} & $\epsilon=0.001$      \\
\textbf{DN-NSGAII \cite{liang2016multimodal}} & -     \\
\textbf{MO\_Ring\_PSO\_SCD \cite{yue2017multiobjective}} & $C_1=C_2=2.05$, $W=0.7298$    \\
\textbf{MO\_PSO\_MM \cite{Jing2018A}} & $C_1=C_2=2.05$, $W=0.7298$, $\eta_0=0.7$    \\
\textbf{DNEA \cite{liu2018double}} & $\sigma_{obj}=0.06$, $\sigma_{var}=0.02$     \\
\textbf{Tri-MOEA\&TAR \cite{liu2018multi}} & $\epsilon_{peak}=0.01$, $\sigma=0.05$, $p_{con}=0.2$  \\
\textbf{DNEA-L \cite{liu2019searching}} & $nb=3$, $K=3$, $N_{ns}=50$     \\
\textbf{CPDEA \cite{liu2019handling}} & $\eta=2$, $K=3$  \\
\textbf{MP-MMEA \cite{tian2021multi}} & -  \\
\textbf{MMOEA/DC \cite{lin2020multimodal}} & $\lambda=0.1$, $\beta=5$     \\
\textbf{MMEA-WI \cite{li2021weighted}} & $p=0.4$ \\
\textbf{HREA \cite{li2022hierarchy}} &   $\epsilon=0.2$, $\epsilon=0.5$ for IMDPM3\_e \\
\hline
\end{tabular}%
\label{tab_para}
\end{table}

\subsection{Performance evaluation}
As we discussed in \sref{sec_metrics}, we select $IGD$, $IGDX$ and $1/PSP$ as the performance metrics to comprehensively compare the performance of all MMEAs. In addition, to evaluate the overall performances, the non-parametric statistical test Friedman test \cite{friedman1940comparison} is adopted. Specifically, for each test problem, the results of 30 times independent runs of all MMEAs are used to calculate the average ranks ($r_i^j$, where $i$ and $j$ are indexes of algorithms and test problems) by the Friedman test. Then, for a test suite, these ranks are summed to calculate the overall average ranks, which indicate the performances of MMEAs in the specific test suite, shown as follows:
\begin{equation}
R_i = \frac{\sum_{j=1}^{J}r_i^j}{J}
\end{equation}
where $J$ is the number of test problems, e.g., $J=8$ for MMF test suites, $J=12$ for IDMP test suites. The smaller the $R_i$ is, the better performance of the $i$-th MMEA. 

\section{Results and discussion}
\label{sec_result}
To comprehensively compared the performances of MMEAs, four groups of MMOP test problems are adopted. It's worth mentioning that, all experiments are independently executed 30 times and the average results are used to present and analyze. However, due to page limitations, the detailed values of the running result are provided in the supplementary file, e.g., $IGD$, $IGDX$ and $1/PSP$. Each table (Table S-III to Table S-XIV) shows the average and variance over 30 runs with the best results highlighted. In addition, all experimental results can be obtained online. \tref{tab_avgrank} lists the detailed average rank values IGD, IGDX and 1/PSP for all algorithms. In addition, the intuitive exhibitions are provided with bar plots in the following discussion.

\begin{table*}
\caption{Average ranks of IGD, IGDX and 1/PSP for 12 compared MMEAs on four different MMOP test suites, where the best rank is highlighted with gray background.}
\resizebox{\linewidth}{!}{
\begin{tabular}{m{3em}<{\centering}m{3em}<{\centering}m{4em}<{\centering}m{4em}<{\centering}m{4em}<{\centering}m{4em}<{\centering}m{4em}<{\centering}m{4em}<{\centering}m{4em}<{\centering}m{4em}<{\centering}m{4em}<{\centering}m{4em}<{\centering}m{4em}<{\centering}m{4em}<{\centering}}
\hline
{\textbf{Problems}} &{\textbf{Indicators}}& {\textbf{Omni-optimizer}} & {\textbf{DN-NSGAII}} & {\textbf{MO\_Ring \_PSO\_SCD}} & {\textbf{MO\_PSO \_MM}} & {\textbf{DNEA}} & {\textbf{Tri-MO EA\&TAR}} & {\textbf{DNEA-L}} & {\textbf{CPDEA}} & {\textbf{MP-MMEA}} & {\textbf{MMOEA /DC}} & {\textbf{MMEA-WI}} & {\textbf{HREA}} \bigstrut\\

\hline
\multirow{3}[2]{*}{{\textbf{MMF}}} & {\textbf{IGD}} & {5.43 } & {8.30 } & {8.21 } & {3.87 } & {2.92 } & {8.61 } & {10.06 } & \cellcolor[rgb]{ .749,  .749,  .749}{2.67 } & {8.79 } & {4.71 } & {7.20 } & {7.24 } \bigstrut[t]\\
      & {\textbf{IGDX}} & {10.87 } & {11.05 } & {6.03 } & {3.30 } & {6.73 } & {9.53 } & {6.04 } & \cellcolor[rgb]{ .749,  .749,  .749}{2.22 } & {9.63 } & {4.43 } & {5.45 } & {2.72 } \\
      & {\textbf{1/PSP}} & {10.83 } & {11.03 } & {6.10 } & {3.35 } & {6.78 } & {9.63 } & {6.03 } & \cellcolor[rgb]{ .749,  .749,  .749}{2.16 } & {9.61 } & {4.38 } & {5.40 } & {2.71 } \bigstrut[b]\\
\hline
\multirow{3}[2]{*}{{\textbf{IDMP}}} & {\textbf{IGD}} & {7.58 } & {8.14 } & {11.26 } & {11.34 } & \cellcolor[rgb]{ .749,  .749,  .749}{1.23 } & {8.44 } & {1.93 } & {4.17 } & {5.92 } & {7.83 } & {6.09 } & {4.10 } \bigstrut[t]\\
      & {\textbf{IGDX}} & {9.26 } & {9.39 } & {7.01 } & {6.30 } & {9.53 } & {10.44 } & {3.94 } & {4.62 } & {7.86 } & {2.93 } & {3.98 } & \cellcolor[rgb]{ .749,  .749,  .749}{2.75 } \\
      & {\textbf{1/PSP}} & {8.88 } & {9.00 } & {6.86 } & {6.16 } & {9.92 } & {10.69 } & {3.95 } & {4.67 } & {8.07 } & {2.91 } & {4.01 } & \cellcolor[rgb]{ .749,  .749,  .749}{2.88 } \bigstrut[b]\\
\hline
\multirow{3}[2]{*}{{\textbf{MMOPL}}} & {\textbf{IGD}} & {8.40 } & {9.24 } & {7.56 } & {7.55 } & {6.90 } & {11.29 } & {3.48 } & {6.60 } & {5.88 } & {1.55 } & {8.01 } & \cellcolor[rgb]{ .749,  .749,  .749}{1.54 } \bigstrut[t]\\
      & {\textbf{IGDX}} & {9.46 } & {9.46 } & {6.66 } & {7.05 } & {8.52 } & {10.10 } & {3.39 } & {7.08 } & {5.90 } & {1.94 } & {7.33 } & \cellcolor[rgb]{ .749,  .749,  .749}{1.12 } \\
      & {\textbf{1/PSP}} & {9.04 } & {8.93 } & {5.72 } & {6.18 } & {9.89 } & {11.19 } & {3.35 } & {7.70 } & {5.63 } & {1.94 } & {7.31 } & \cellcolor[rgb]{ .749,  .749,  .749}{1.12 } \bigstrut[b]\\
\hline
\multicolumn{1}{c}{\multirow{3}[2]{*}{{\textbf{Polygon}}}} & {\textbf{IGD}} & {11.04 } & {10.99 } & {8.64 } & {9.63 } & {2.33 } & {7.96 } & {4.39 } & {2.94 } & {8.04 } & {4.80 } & \cellcolor[rgb]{ .749,  .749,  .749}{2.40 } & {4.86 } \bigstrut[t]\\
      & {\textbf{IGDX}} & {11.41 } & {11.21 } & {7.61 } & {8.74 } & {5.05 } & {8.08 } & {4.41 } & {5.01 } & {8.39 } & {3.09 } & \cellcolor[rgb]{ .749,  .749,  .749}{2.15 } & {2.88 } \\
      & {\textbf{1/PSP}} & {10.60 } & {10.63 } & {7.57 } & {8.57 } & {5.17 } & {8.69 } & \cellcolor[rgb]{ .749,  .749,  .749}{2.78 } & {4.91 } & {8.28 } & {2.42 } & {3.40 } & {4.99 } \bigstrut[b]\\
\hline
\multicolumn{1}{c}{\multirow{3}[2]{*}{{\textbf{Overall}}}} & {\textbf{IGD}} & {8.72 } & {9.47 } & {8.90 } & {8.63 } & \cellcolor[rgb]{ .749,  .749,  .749}{3.26 } & {8.95 } & {4.99 } & {3.96 } & {7.08 } & {4.65 } & {5.26 } & {4.11 } \bigstrut[t]\\
      & {\textbf{IGDX}} & {10.36 } & {10.34 } & {7.00 } & {6.95 } & {7.16 } & {9.33 } & {4.44 } & {4.97 } & {7.85 } & {2.92 } & {4.29 } & \cellcolor[rgb]{ .749,  .749,  .749}{2.38 } \\
      & {\textbf{1/PSP}} & {9.86 } & {9.90 } & {6.73 } & {6.65 } & {7.64 } & {9.90 } & {3.82 } & {5.09 } & {7.78 } & \cellcolor[rgb]{ .749,  .749,  .749}{2.66 } & {4.76 } & {3.21 } \bigstrut[b]\\
\hline

\end{tabular}%

\label{tab_avgrank}
}
\end{table*}

\subsection{Performance comparison on MMF problems}
\label{sec_mmfresult}
\begin{figure}[tbph]
	\begin{center}
		\includegraphics[width=3.5in]{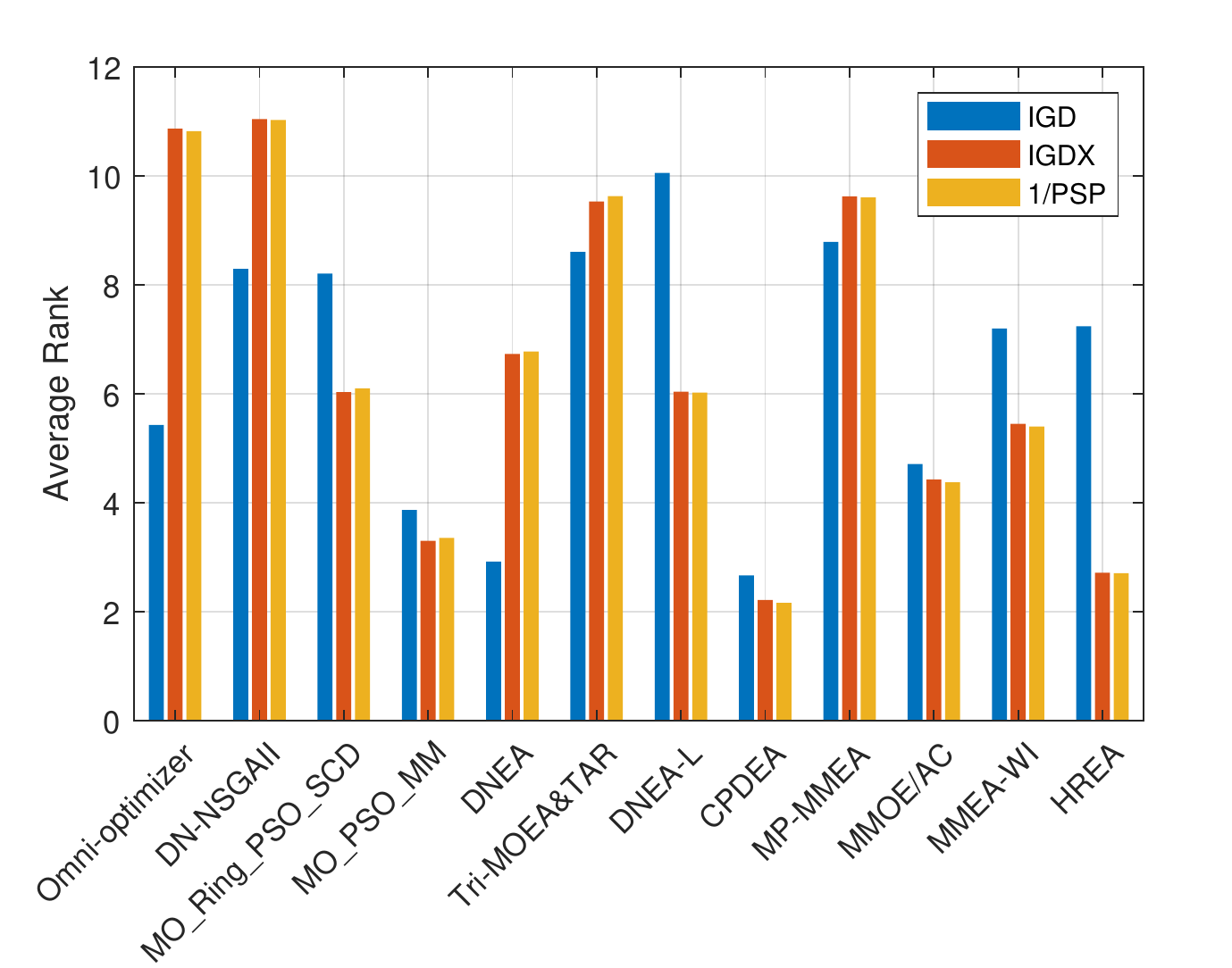}
		\caption{The average rank of all compared MMEAs on MMF test suite in terms of $IGD$, $IGDX$ and $1/PSP$.}
		\label{fig_avgrankmmf}
	\end{center}
\end{figure}

\fref{fig_avgrankmmf} presents the overall rank of all compared MMEAs on the MMF test suite, from which we can see that the average ranks of $IGDX$ and $1/PSP$ are very similar. Although there are minor differences in the values of these two indicators, they are highly consistent in the ranking of algorithm performance. Therefore, in subsequent content, we only discuss the results of $IGD$ and $IGDX$. In terms of $IGD$, CPDEA and DNEA perform better than other compared algorithms. As we can see from Table S-III, which lists the $IGD$ results, DNEA, CPDEA and MMOEA/DC win 5, 1 and 2 instances respectively. Another competitive MMEA is MO\_PSO\_MM, which performs a bit worse than DNEA. As for algorithms that consider local PSs (DNEA-L, MMOEA/DC and HREA), their performances are apparently worse than other MMEAs in terms of $IGD$. Further study shows that some of the obtained solutions can not reach the true PF. The aim to obtain local PS needs to simultaneously explore the whole decision space, which may be the reason for bad performance in terms of $IGD$.

As for $IGDX$ and $1/PSP$, CPDEA, HREA and MO\_PSO\_MM outperform other algorithms. As shown in Table S-IV and Table S-V, HREA, CPDEA and MMOEA/DC win 4, 2 and 2 instances over 8 MMF problems. CPDEA performs best on MMF test problems both in terms of $IGD$ and $IGDX$. The double $k$-nearest neighbor method used in CPDEA can measure the crowding distance in both the decision and objective spaces. HREA performs a bit worse than CPDEA. It adopts the local convergence quality to maintain the diversity which is effective in dealing with MMOPs. Although MO\_PSO\_MM can not obtain the best result for any MMF test problem, its overall performance is relatively strong and stable. The self-organizing method can help form a stable neighbor relationship that can help maintain diversity. Omni-optimizer, DN-NSGAII, Tri-MOEA\&TAR and MP-MMEA are the worst four MMEAs in terms of $IGDX$. To be specific, Omni-optimizer and DN-NSGAII are two primitive and representative MMEAs. Although the diversity-maintaining strategies are relatively weak, many later works are motivated by them. Tri-MOEA\&TAR used a decision variable analysis method, which may be inapplicable for the MMF test suite. MP-MMEA is designed especially for large-scale sparse MMOPs. Thus, its performance on two-variable simple MMOPs is relatively weak.

\begin{figure*}[htbp]
\centering
\subfigure[Omni-optimizer]{\includegraphics[width=1.1in]{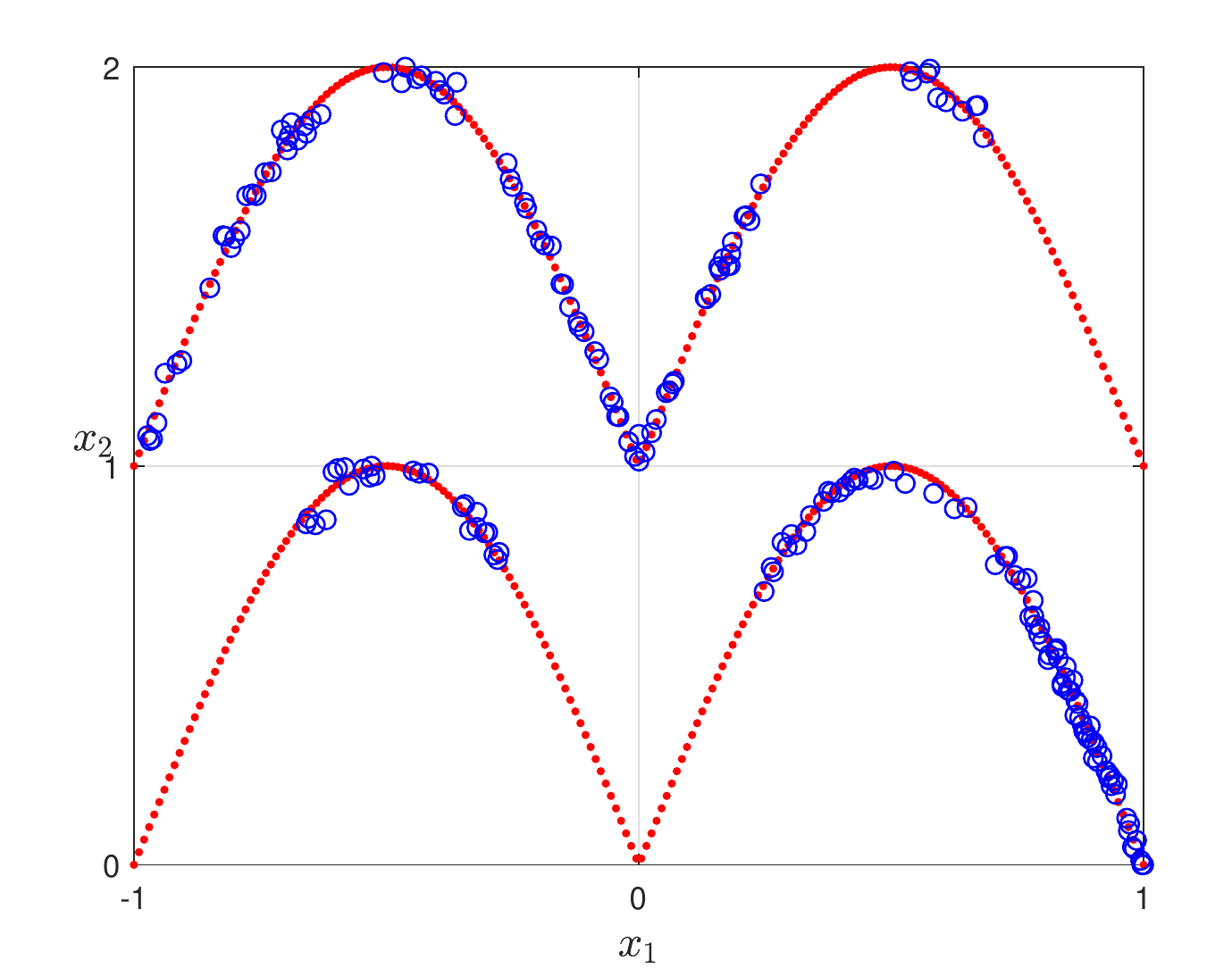}}
\subfigure[DN-NSGAII]{\includegraphics[width=1.1in]{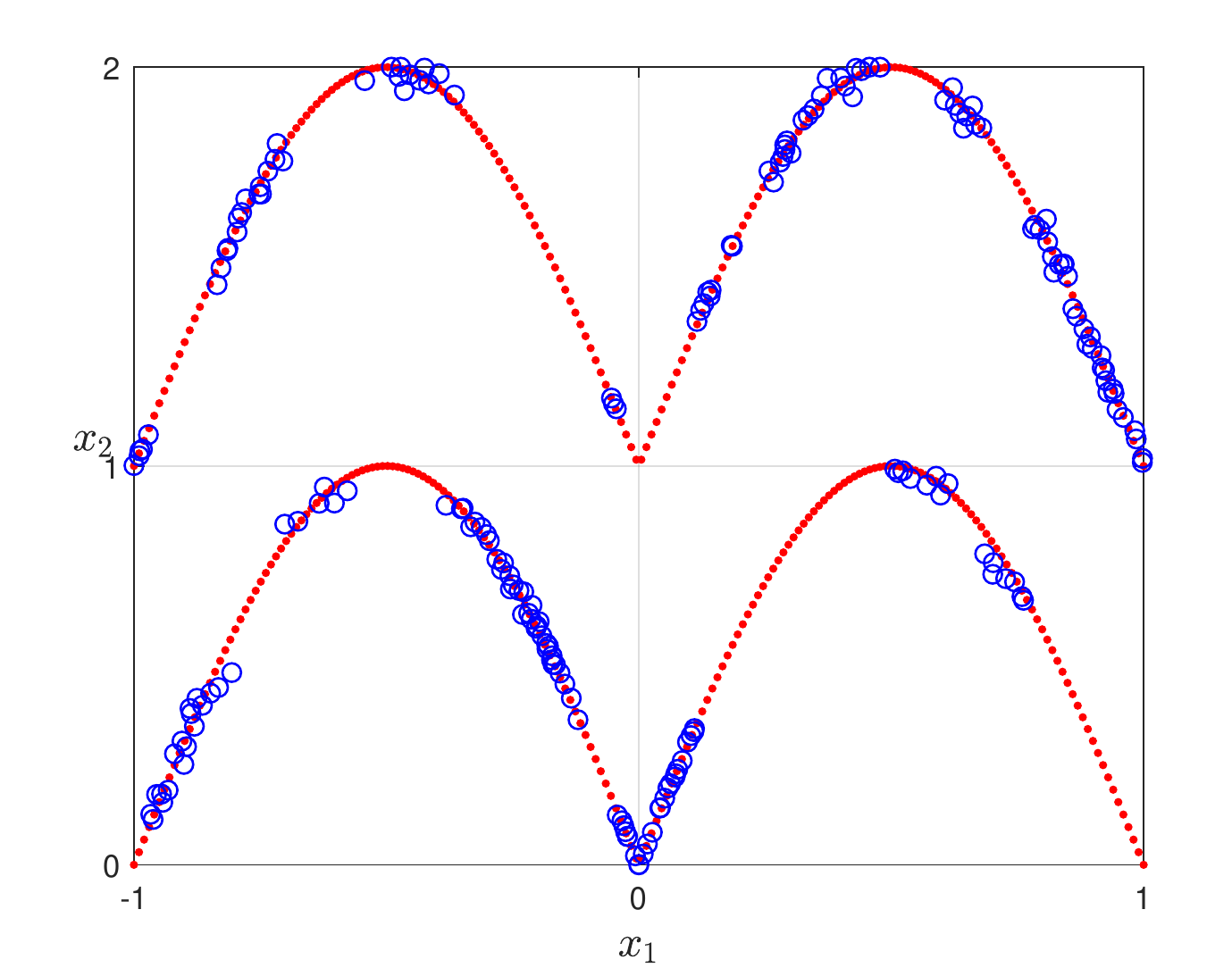}}
\subfigure[MO\_R\_PSO\_SCD]{\includegraphics[width=1.1in]{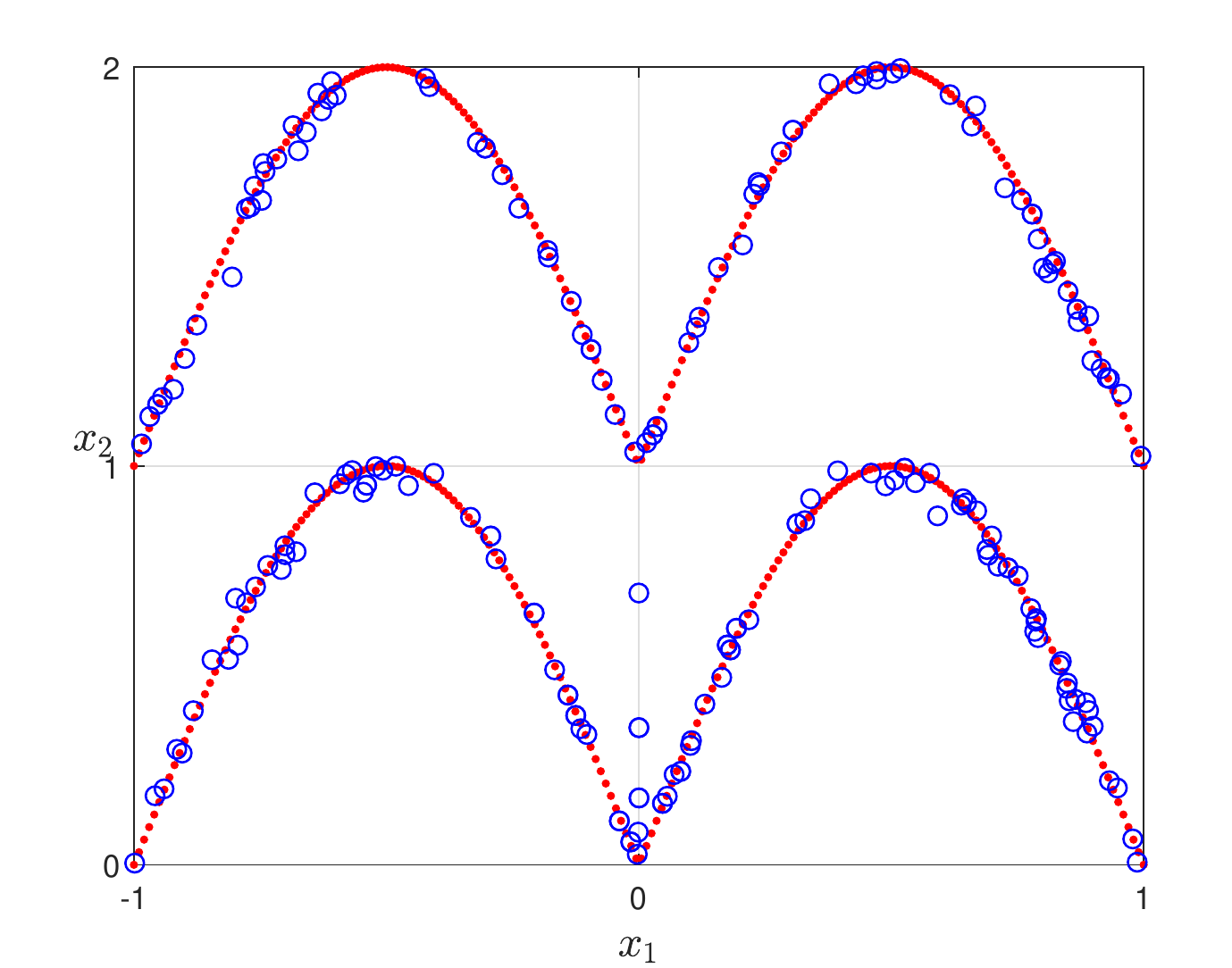}}
\subfigure[MO\_PSO\_MM]{\includegraphics[width=1.1in]{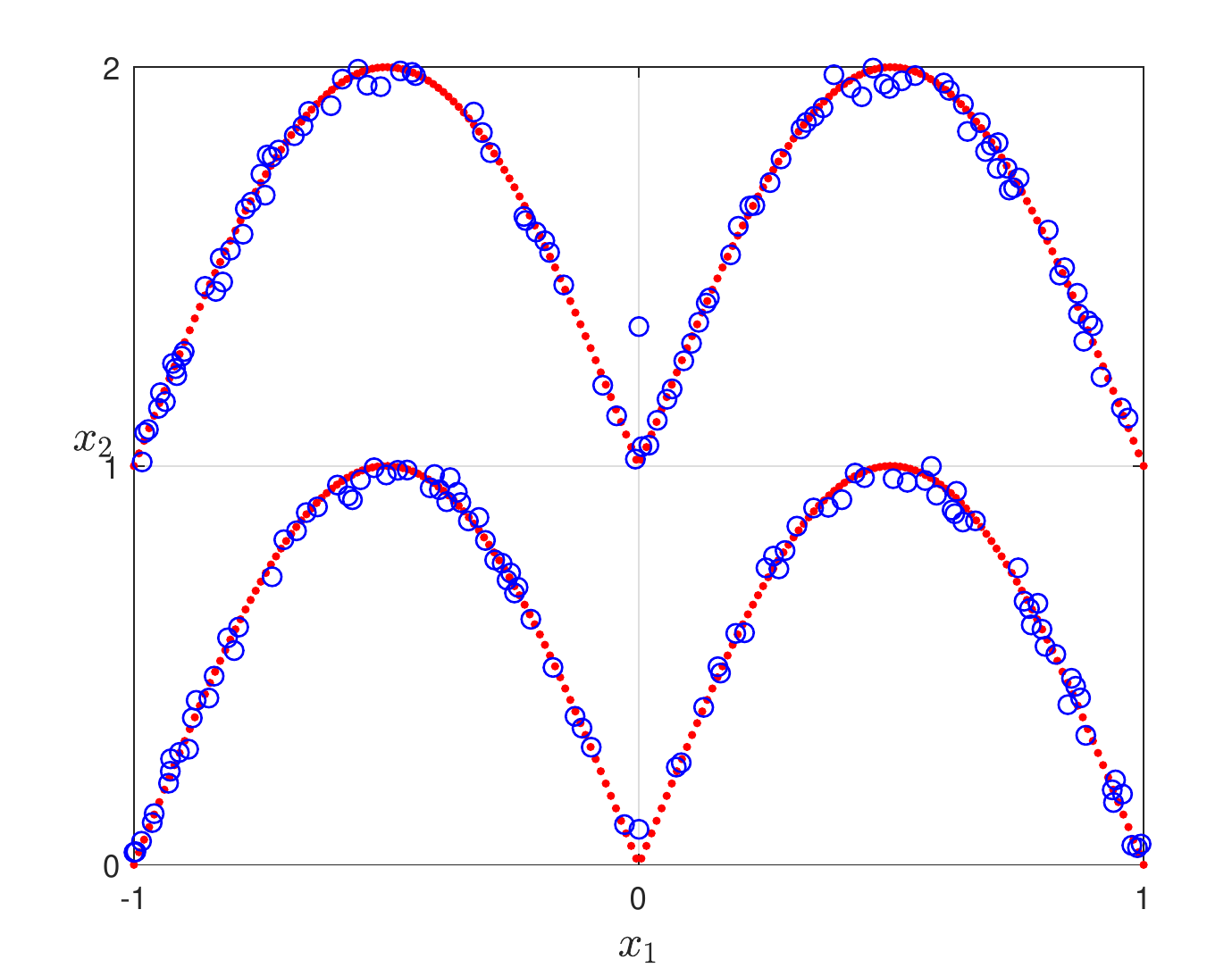}}
\subfigure[DNEA]{\includegraphics[width=1.1in]{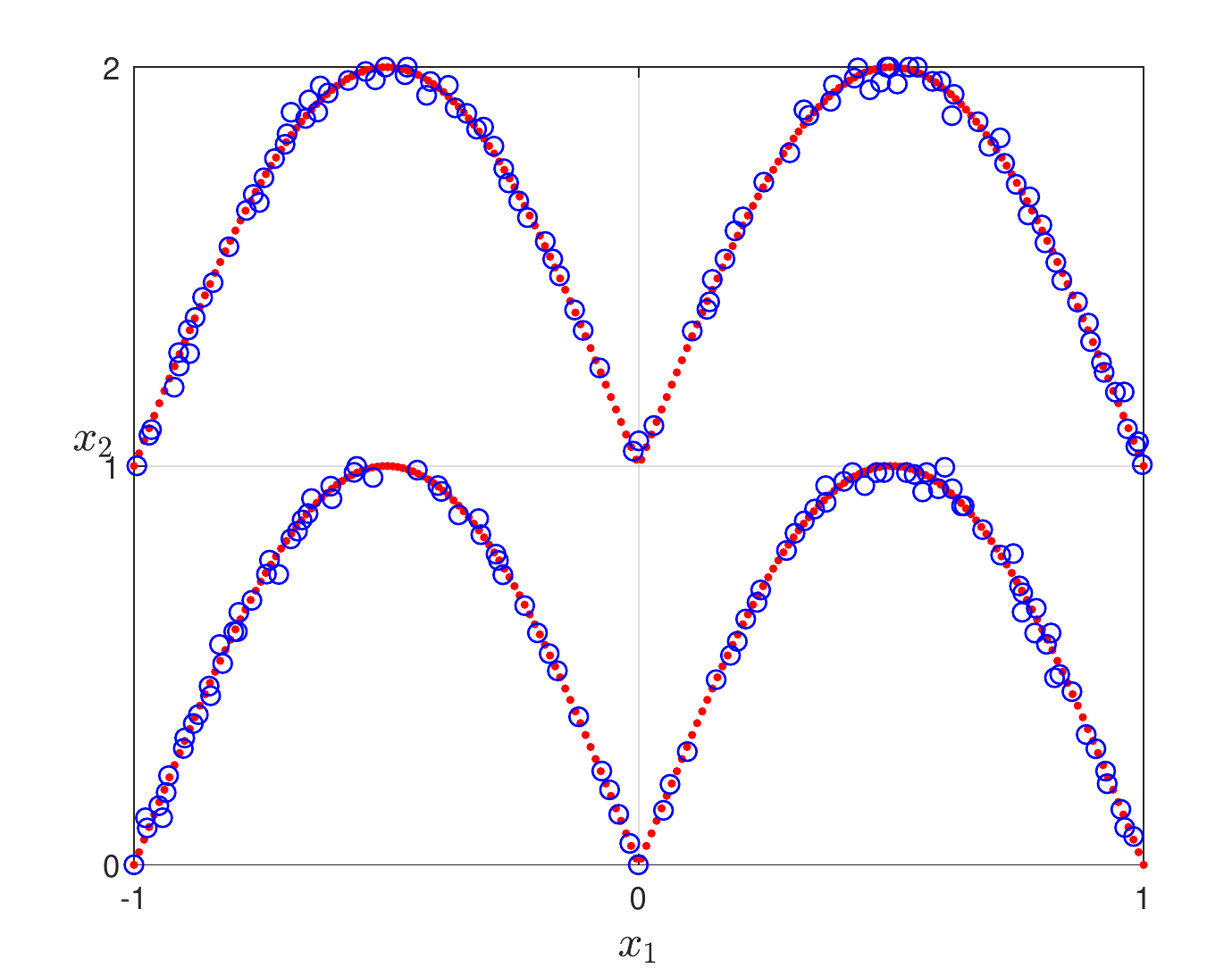}}
\subfigure[Tri-MOEA\&TAR]{\includegraphics[width=1.1in]{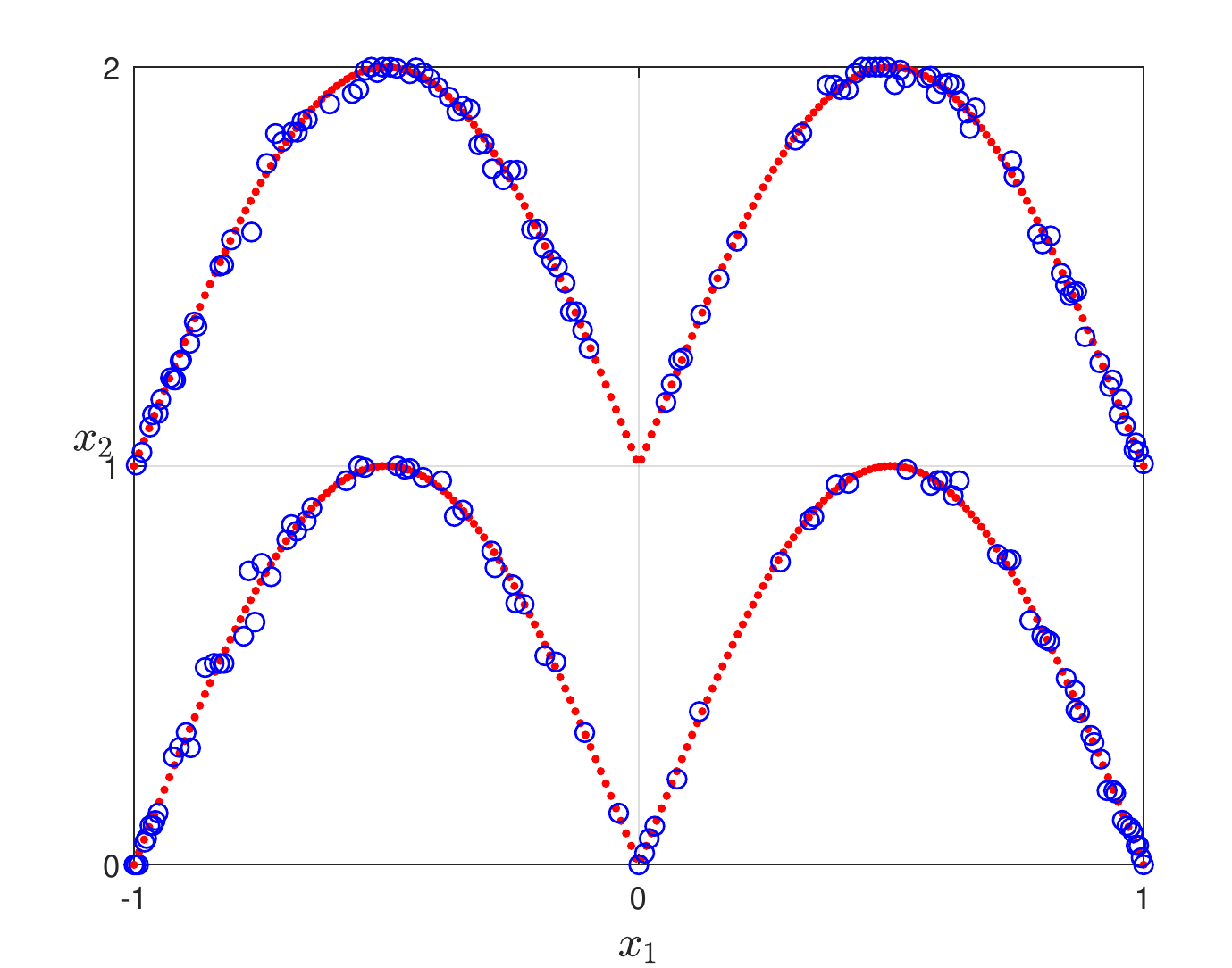}}
\subfigure[DNEA-L]{\includegraphics[width=1.1in]{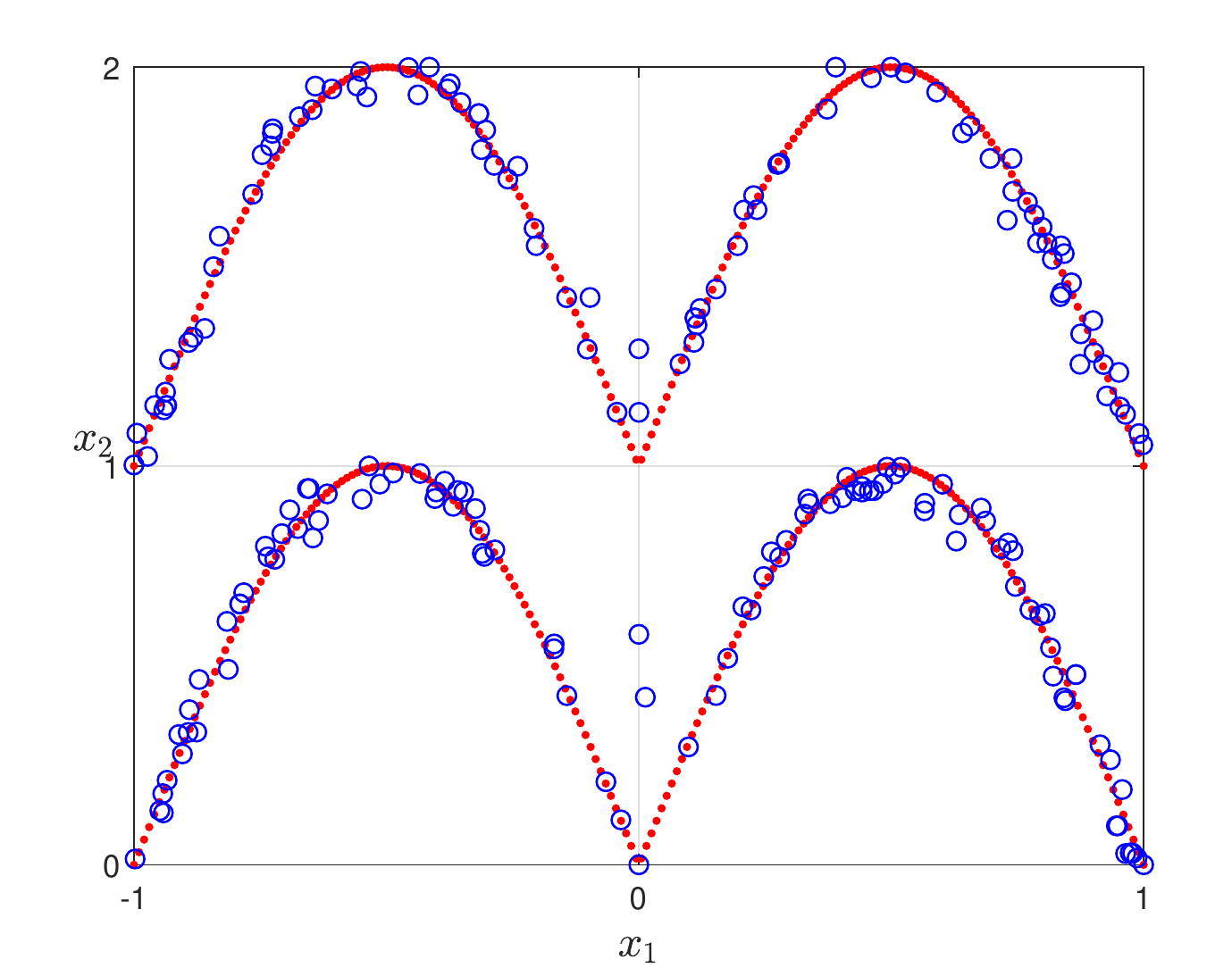}}
\subfigure[CPDEA]{\includegraphics[width=1.1in]{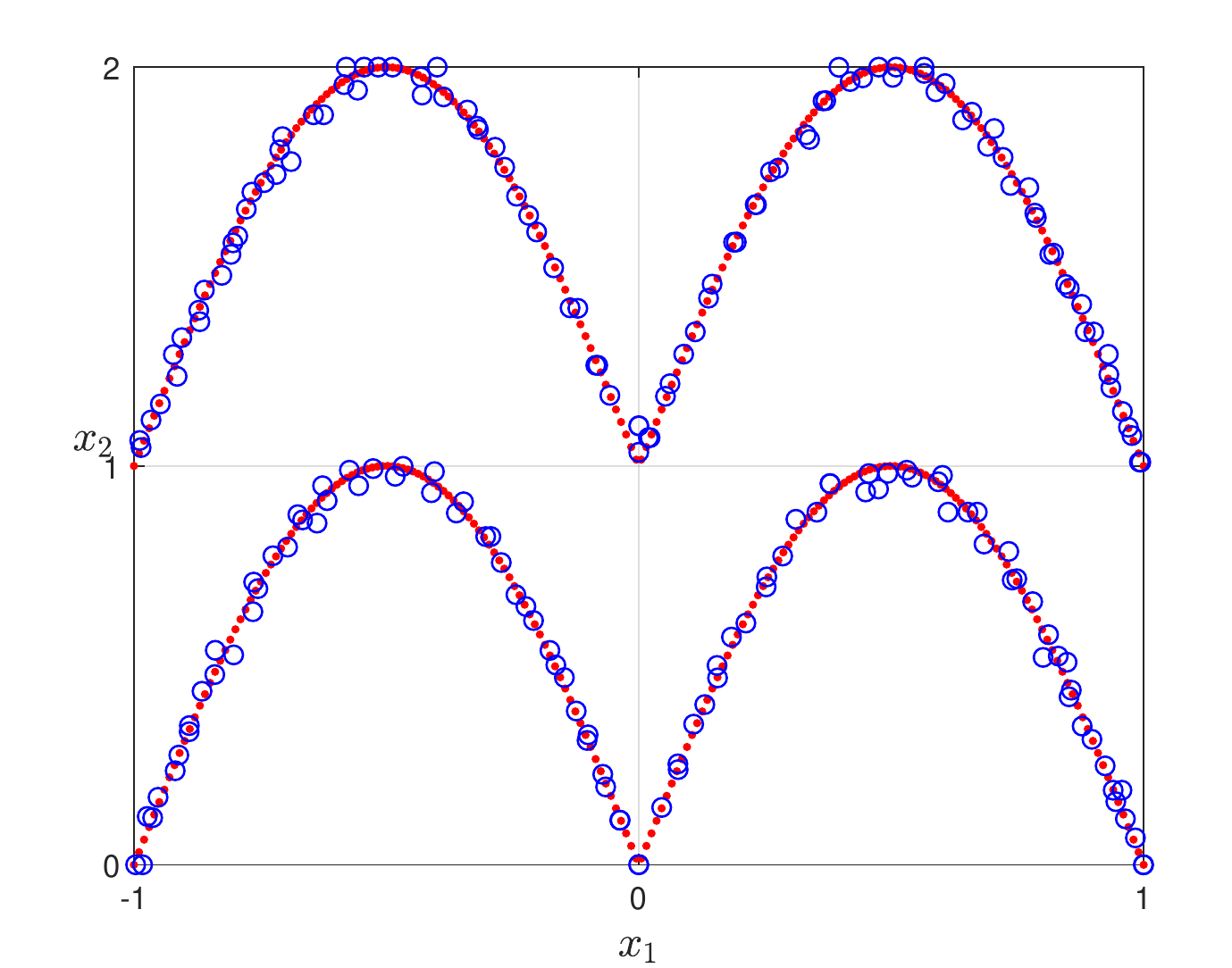}}
\subfigure[MP-MMEA]{\includegraphics[width=1.1in]{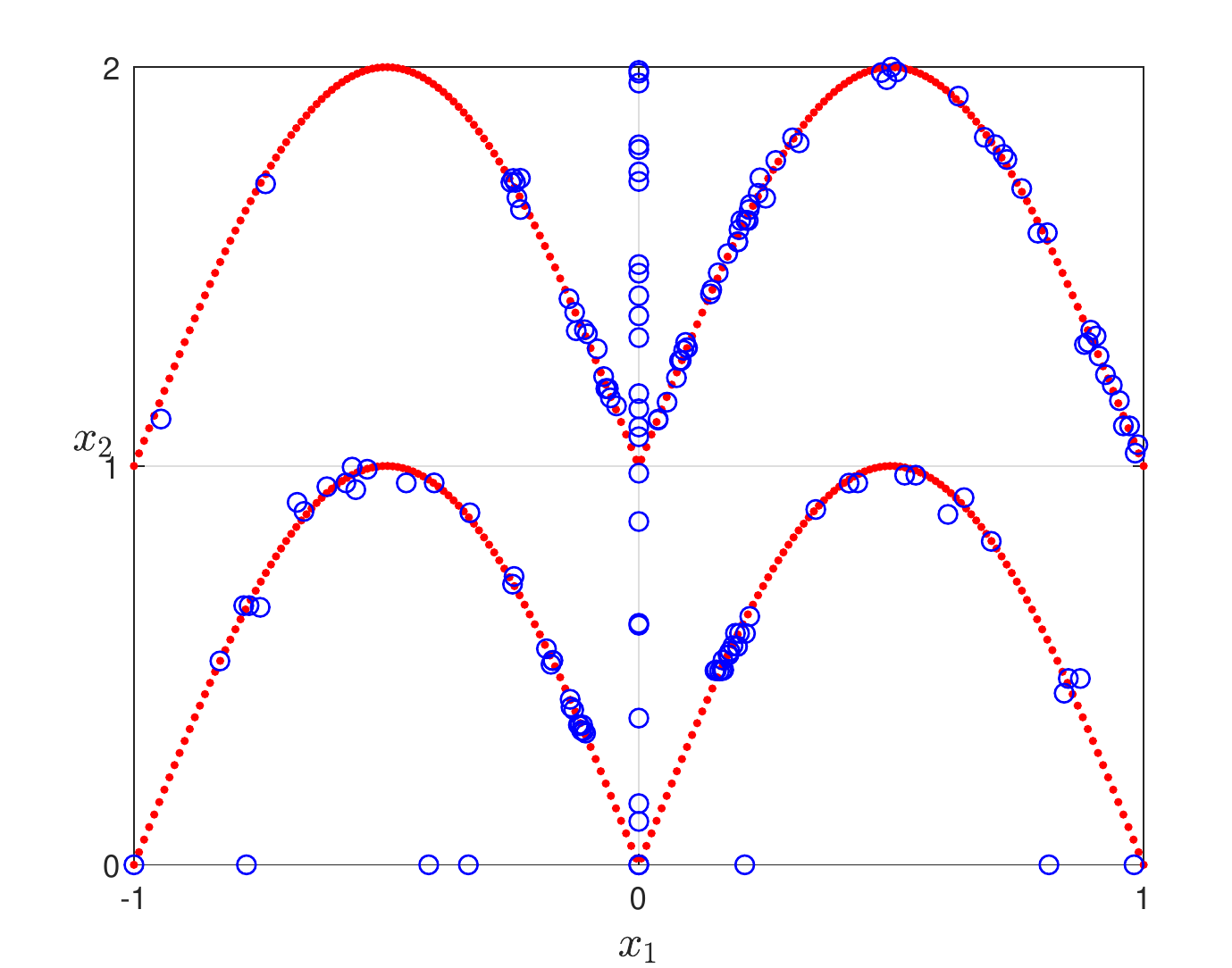}}
\subfigure[MMOEA/DC]{\includegraphics[width=1.1in]{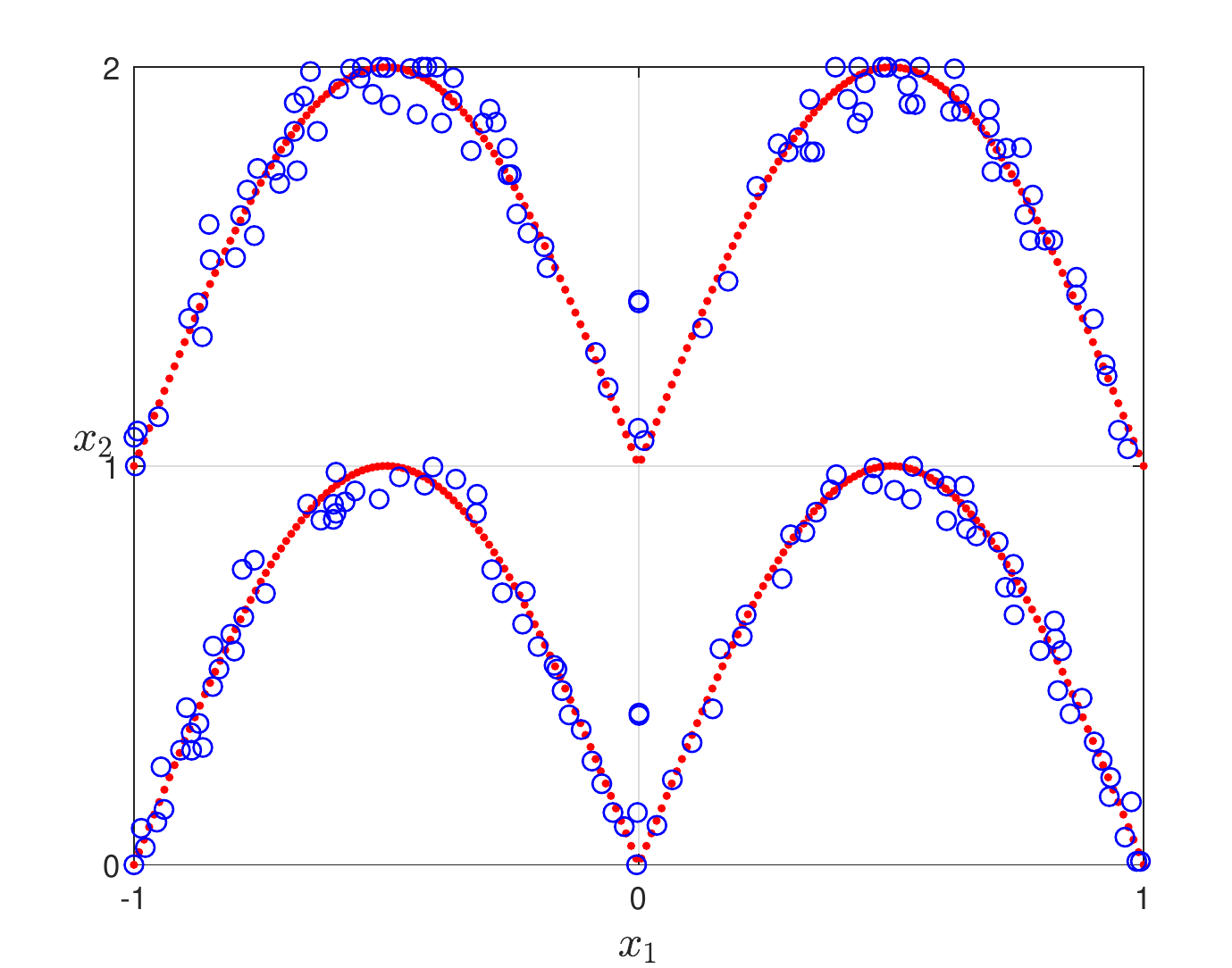}}
\subfigure[MMEA-WI]{\includegraphics[width=1.1in]{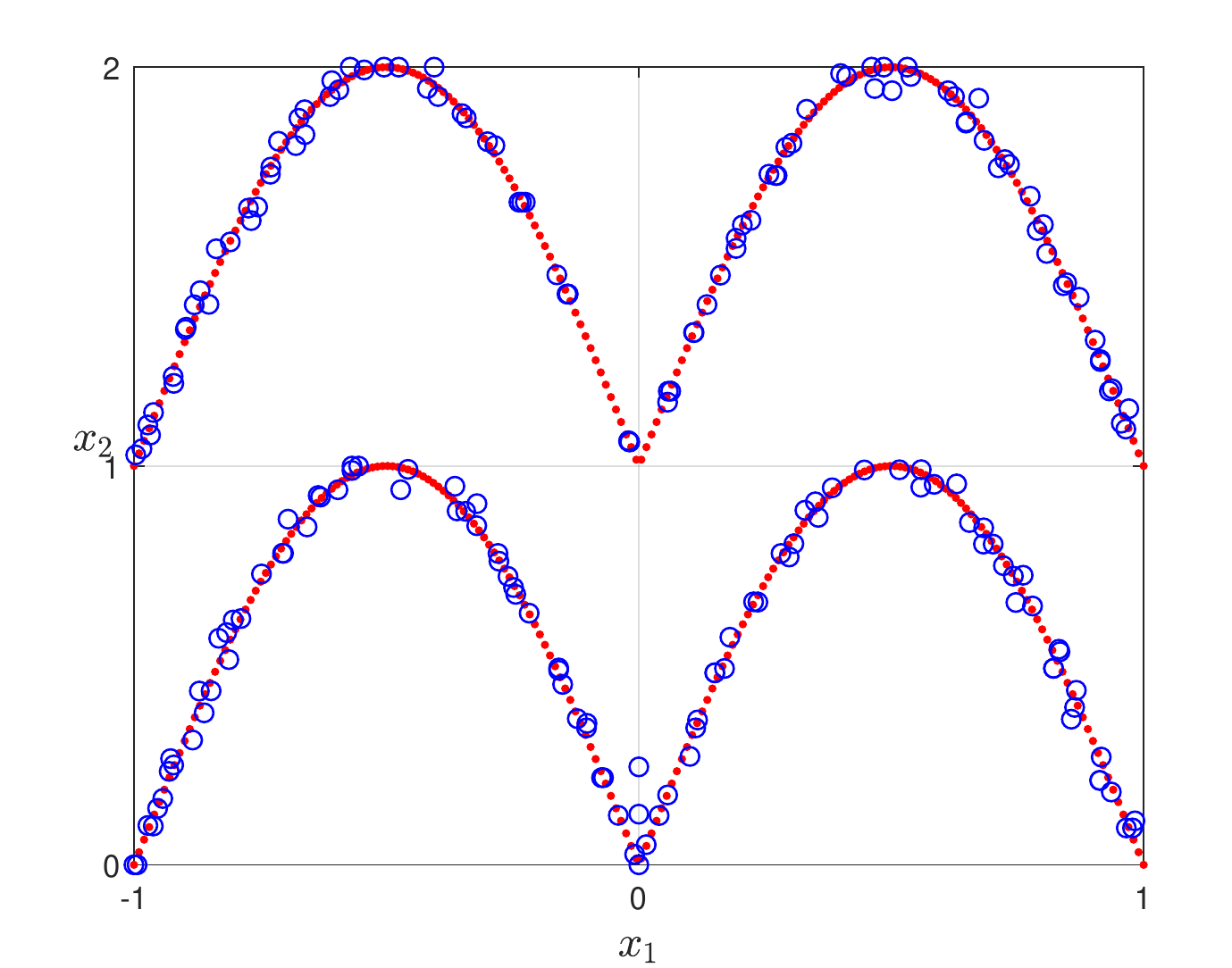}}
\subfigure[HREA]{\includegraphics[width=1.1in]{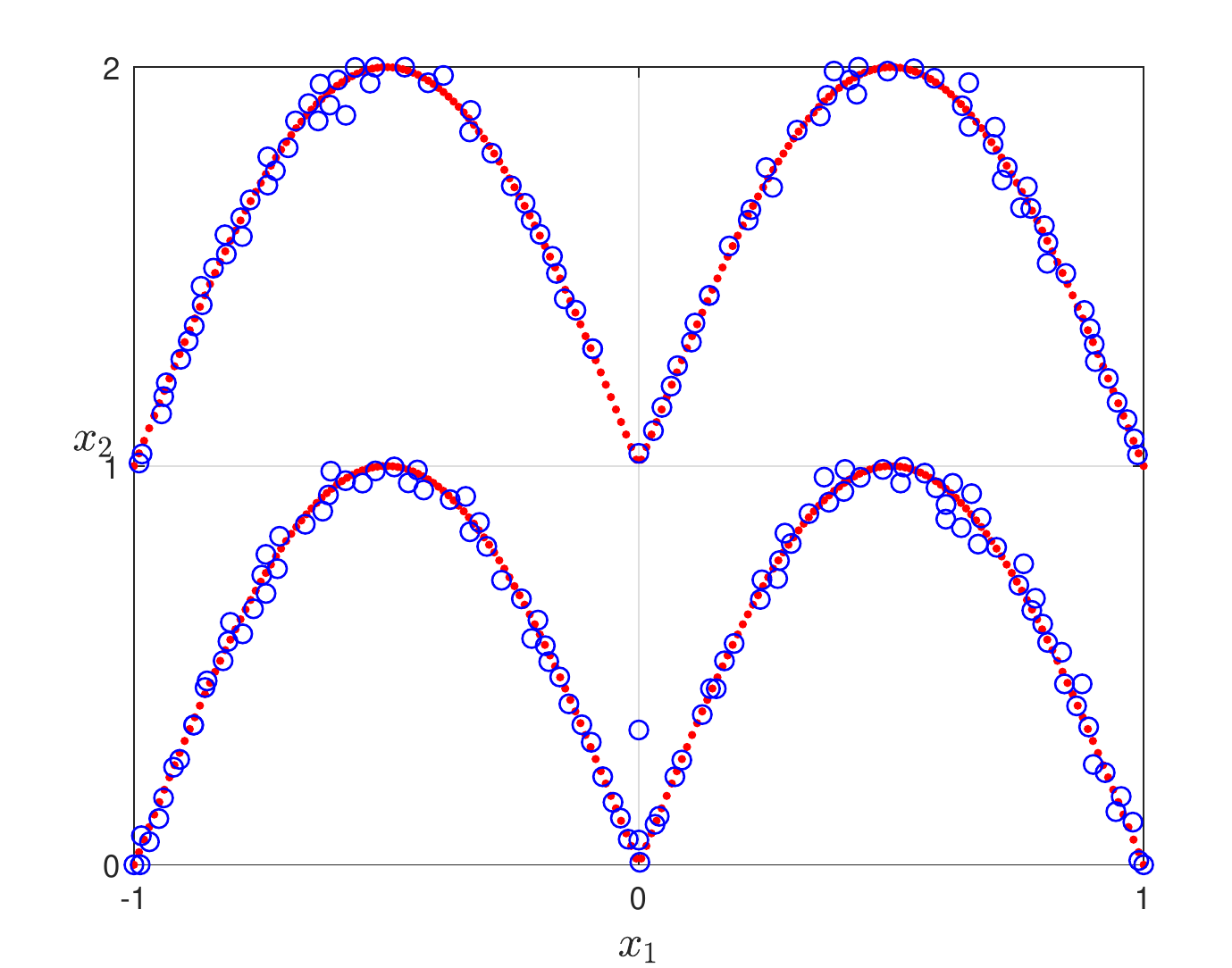}}
\subfigure[Omni-optimizer]{\includegraphics[width=1.1in]{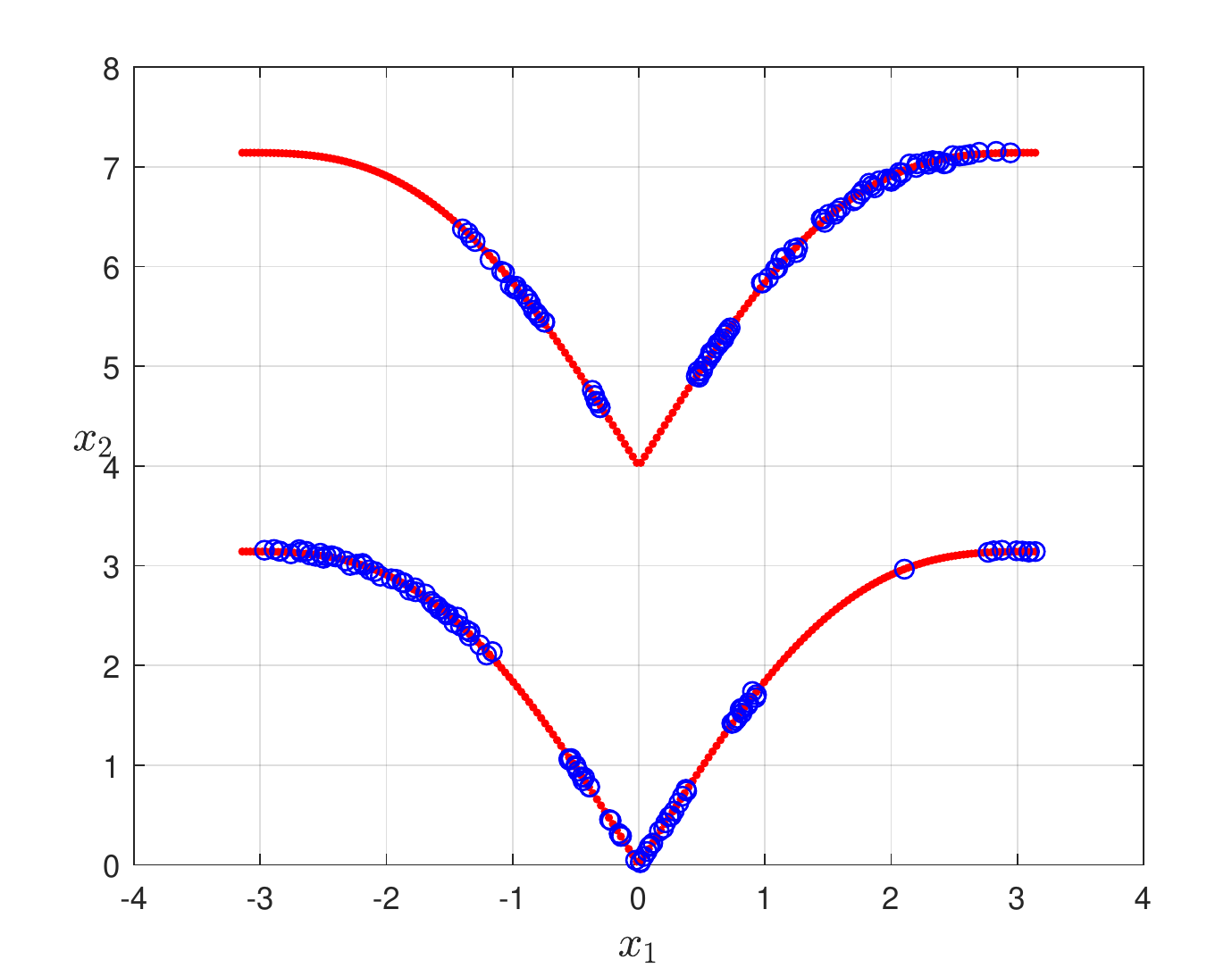}}
\subfigure[DN-NSGAII]{\includegraphics[width=1.1in]{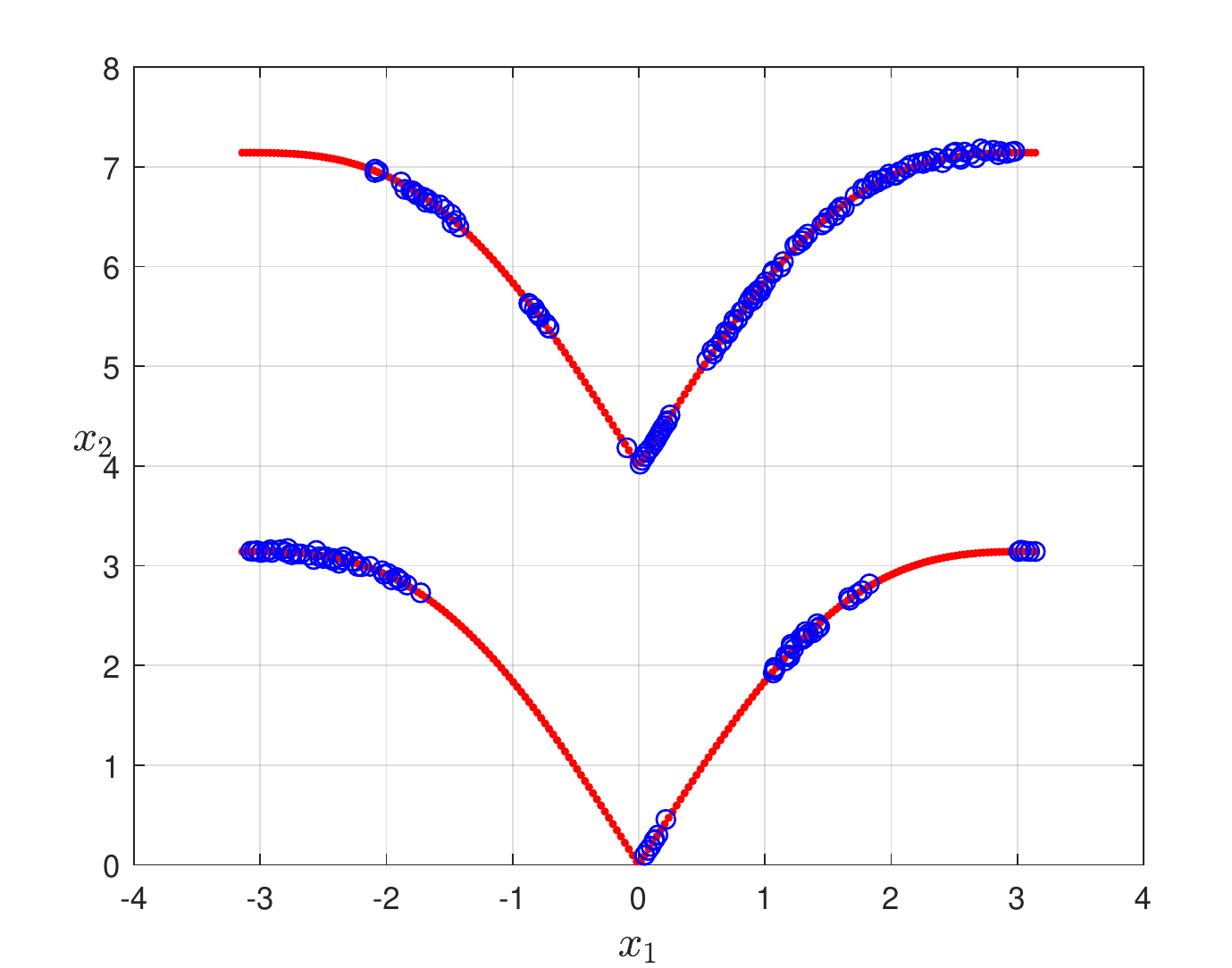}}
\subfigure[MO\_R\_PSO\_SCD]{\includegraphics[width=1.1in]{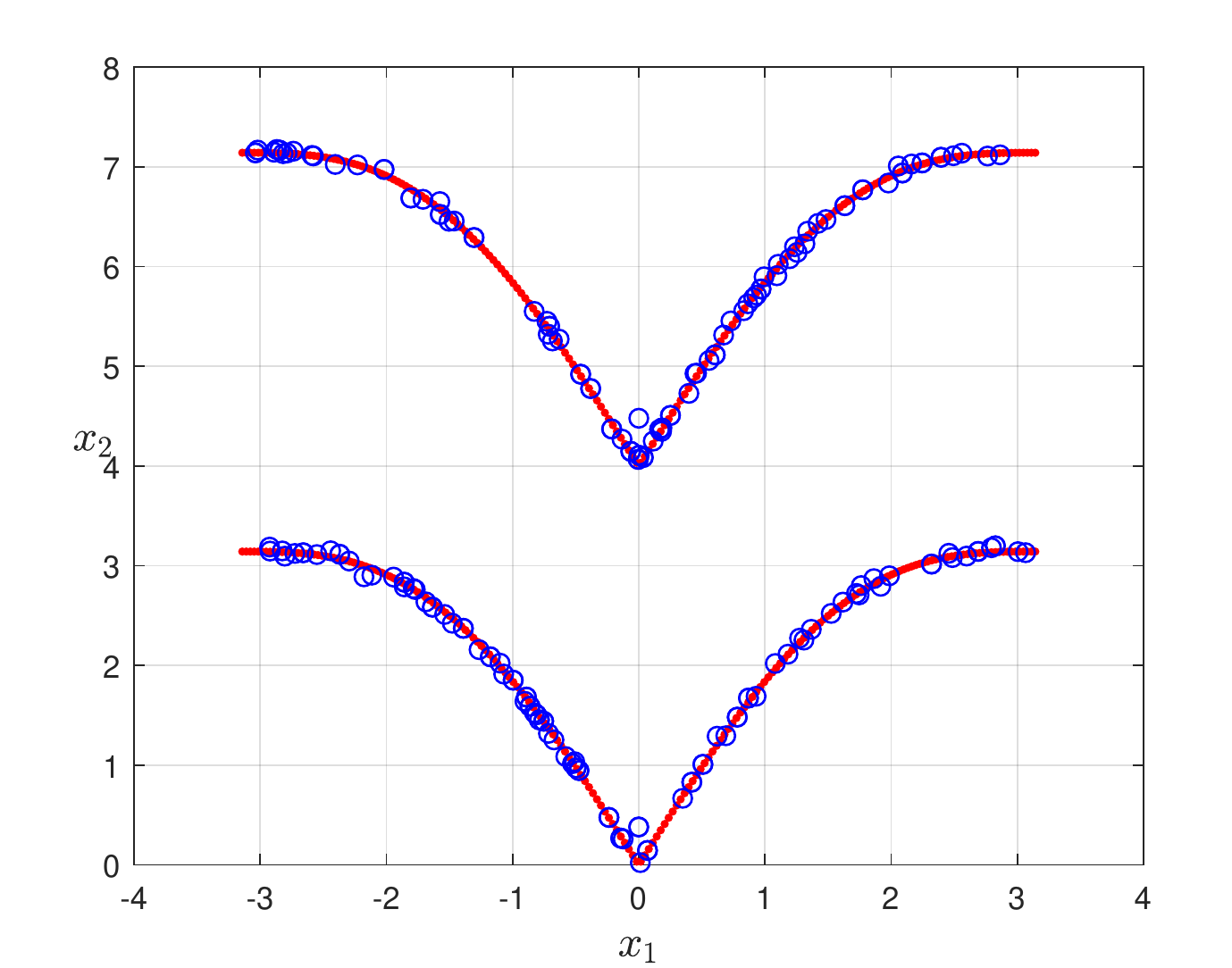}}
\subfigure[MO\_PSO\_MM]{\includegraphics[width=1.1in]{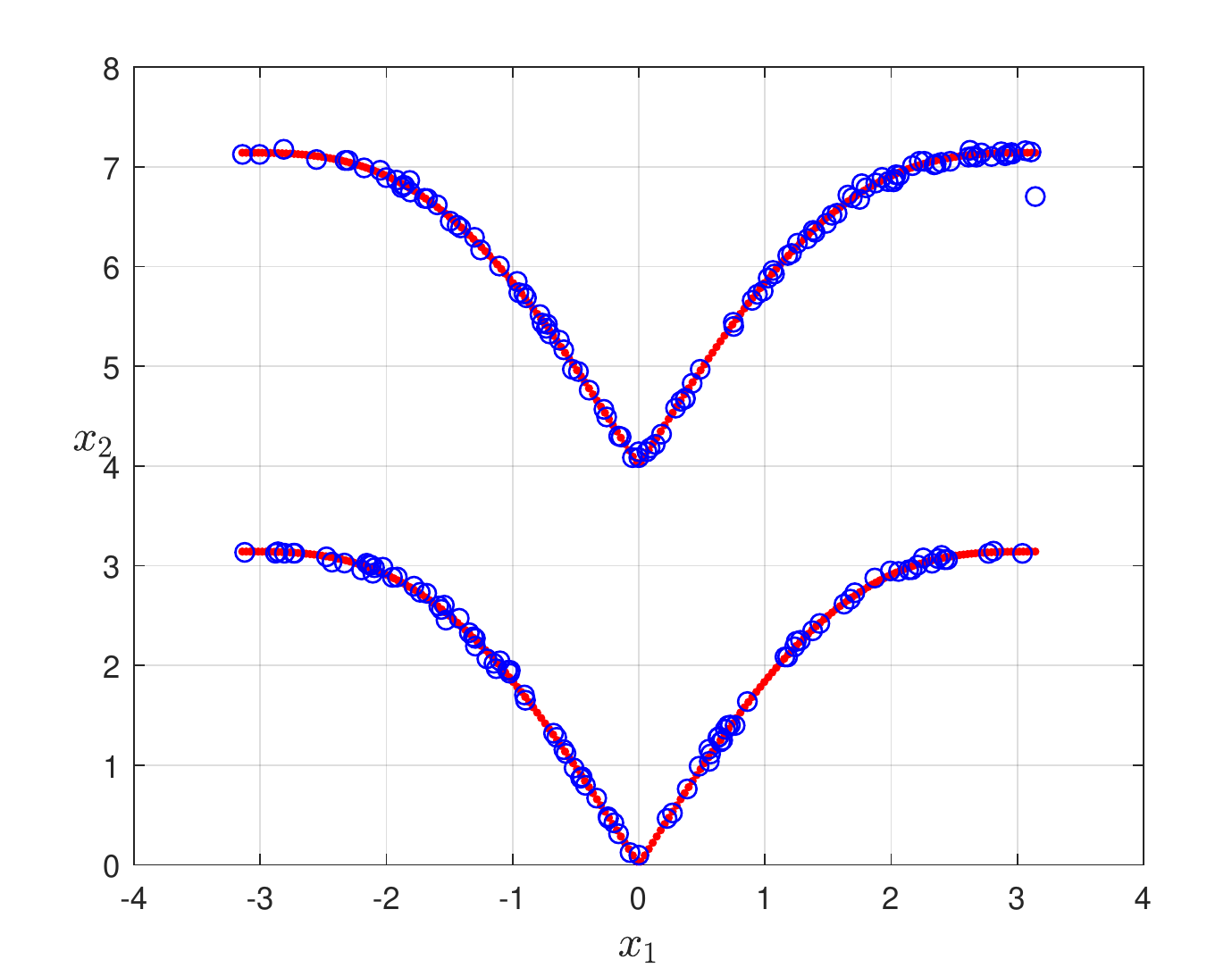}}
\subfigure[DNEA]{\includegraphics[width=1.1in]{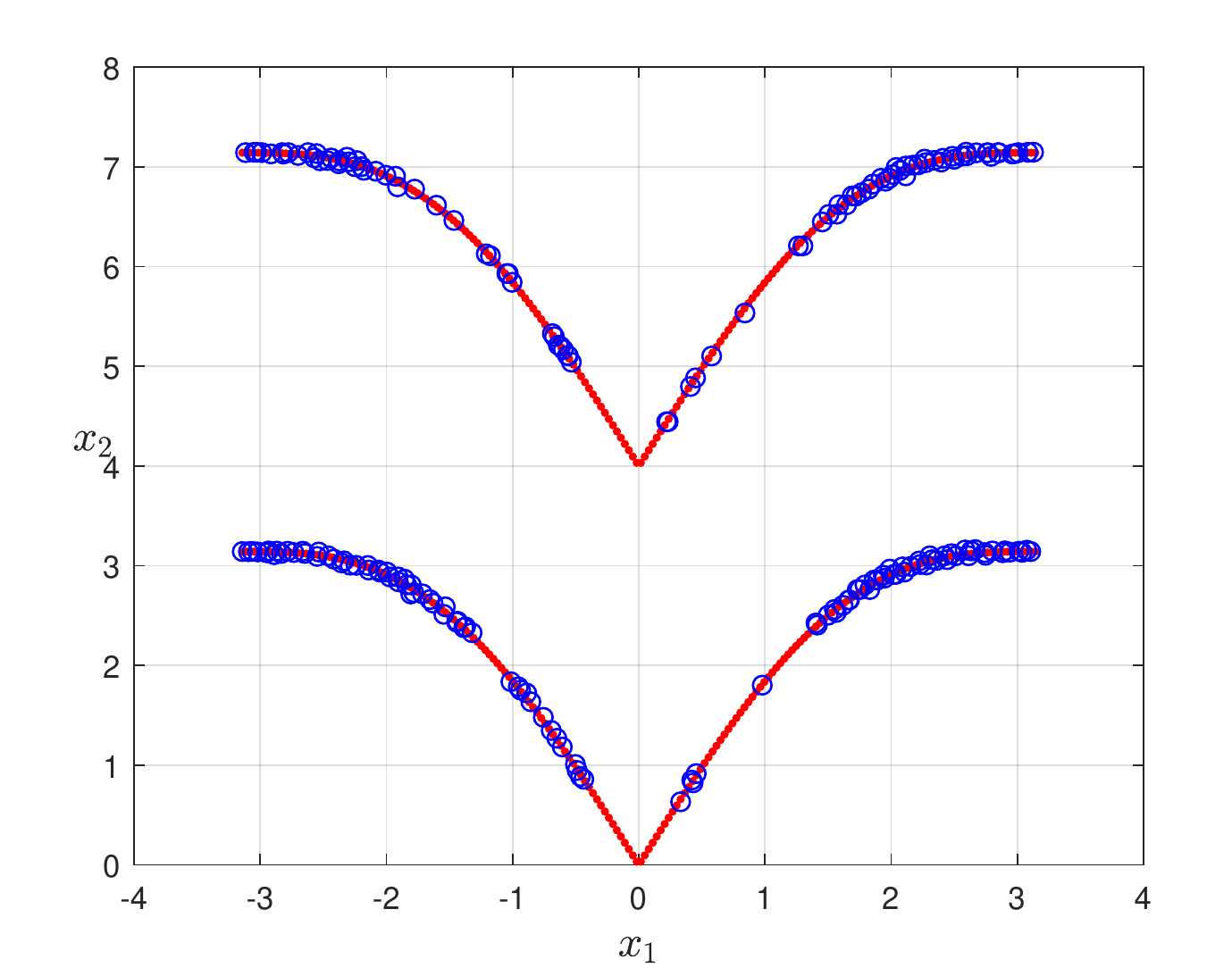}}
\subfigure[Tri-MOEA\&TAR]{\includegraphics[width=1.1in]{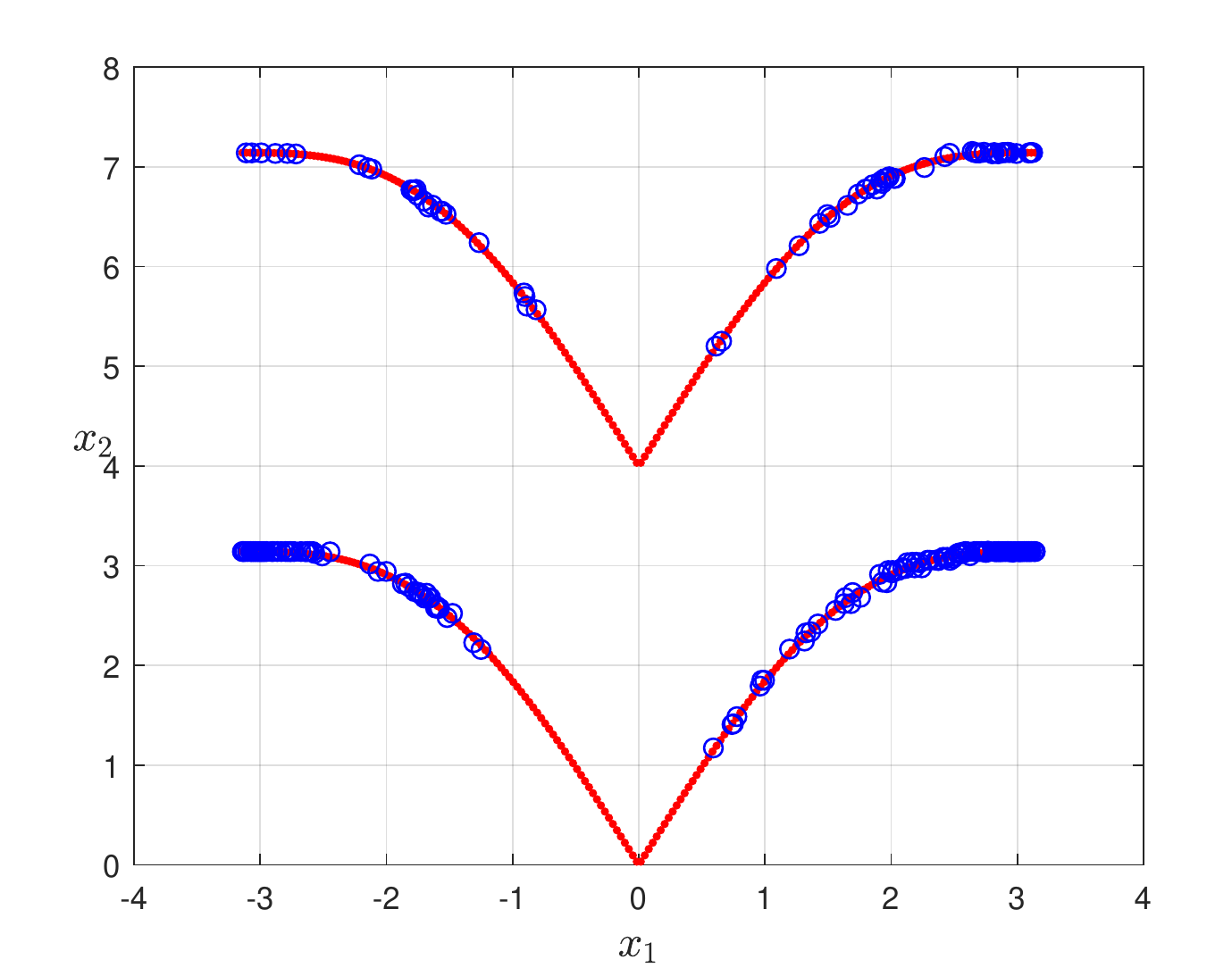}}
\subfigure[DNEA-L]{\includegraphics[width=1.1in]{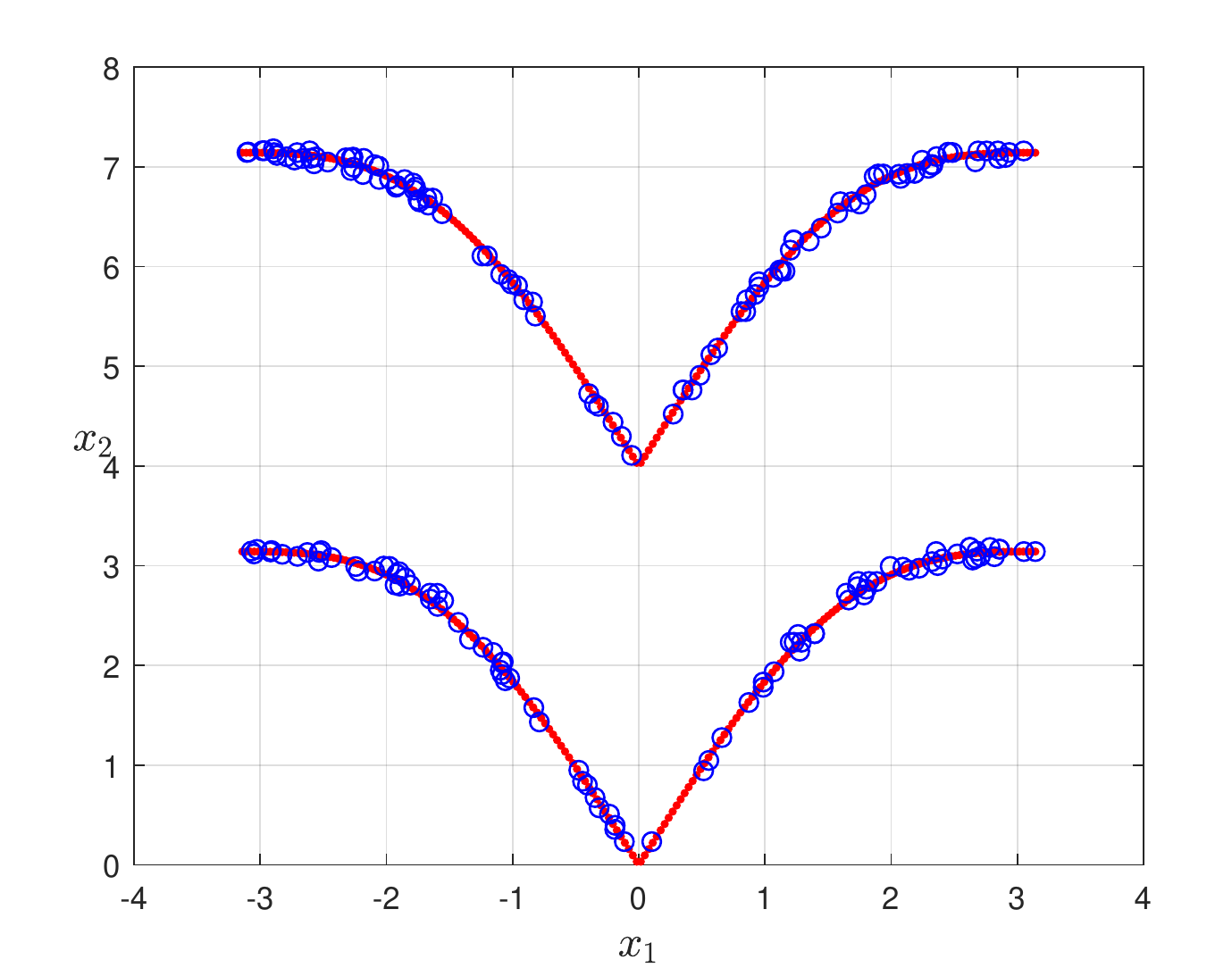}}
\subfigure[CPDEA]{\includegraphics[width=1.1in]{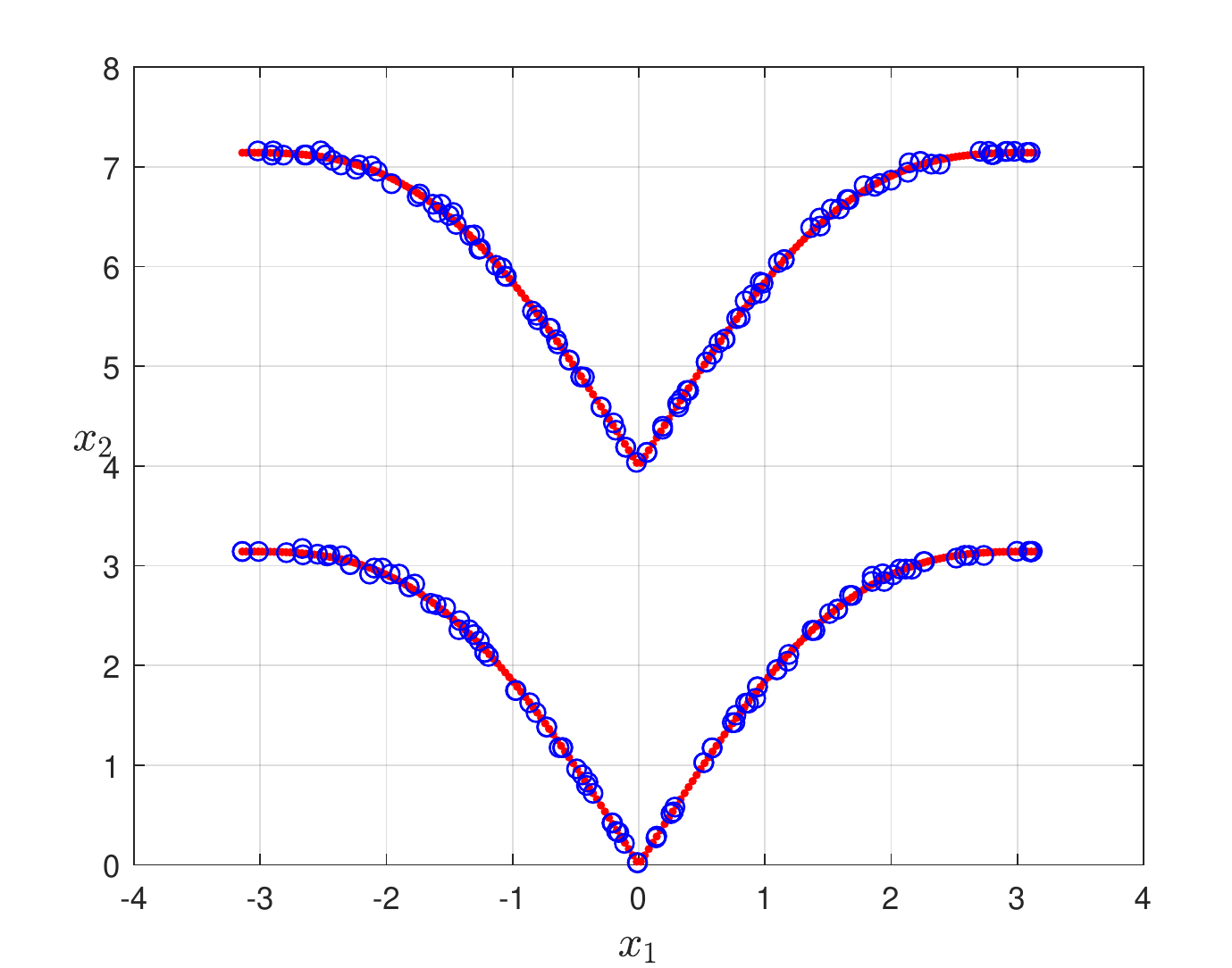}}
\subfigure[MP-MMEA]{\includegraphics[width=1.1in]{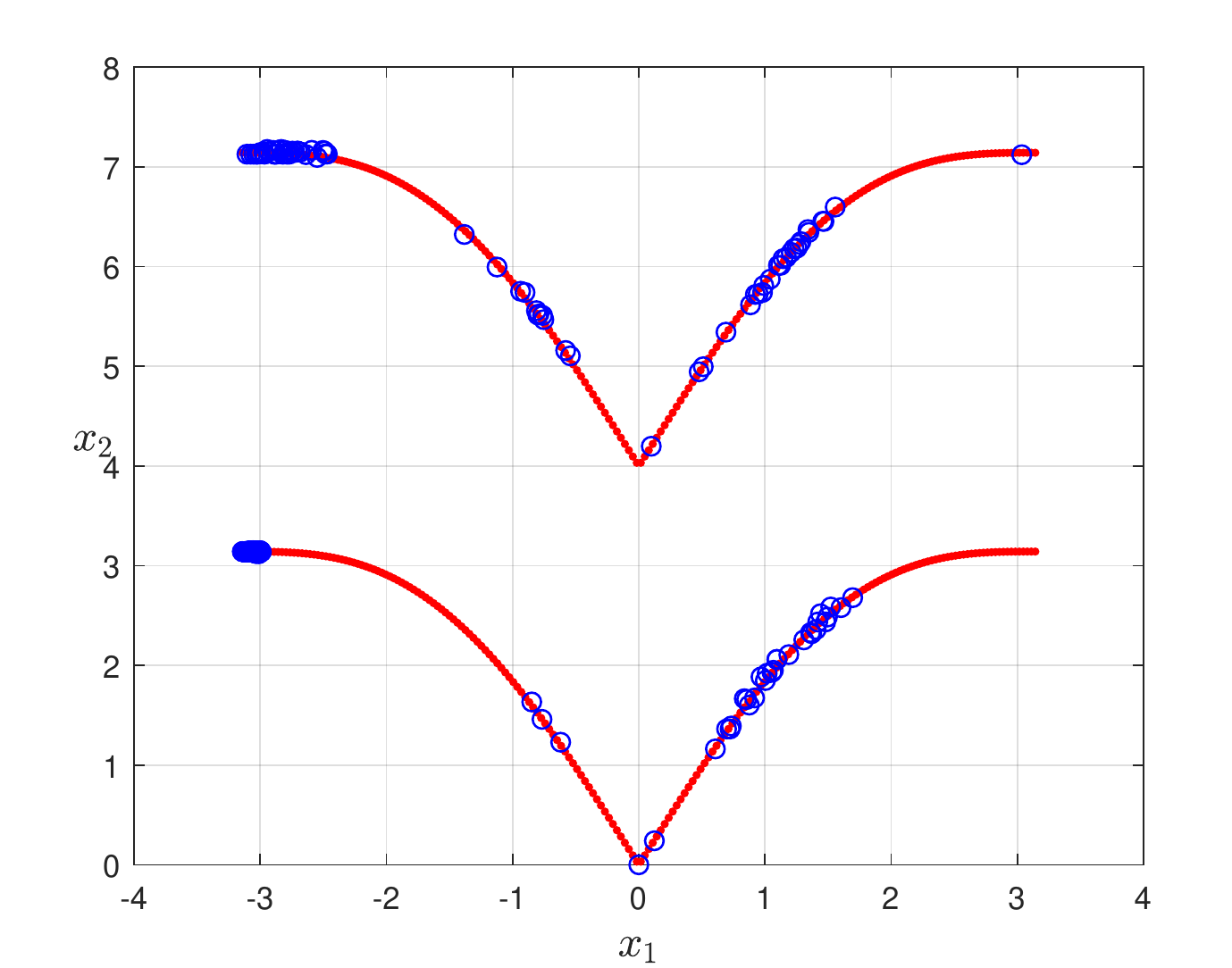}}
\subfigure[MMOEA/DC]{\includegraphics[width=1.1in]{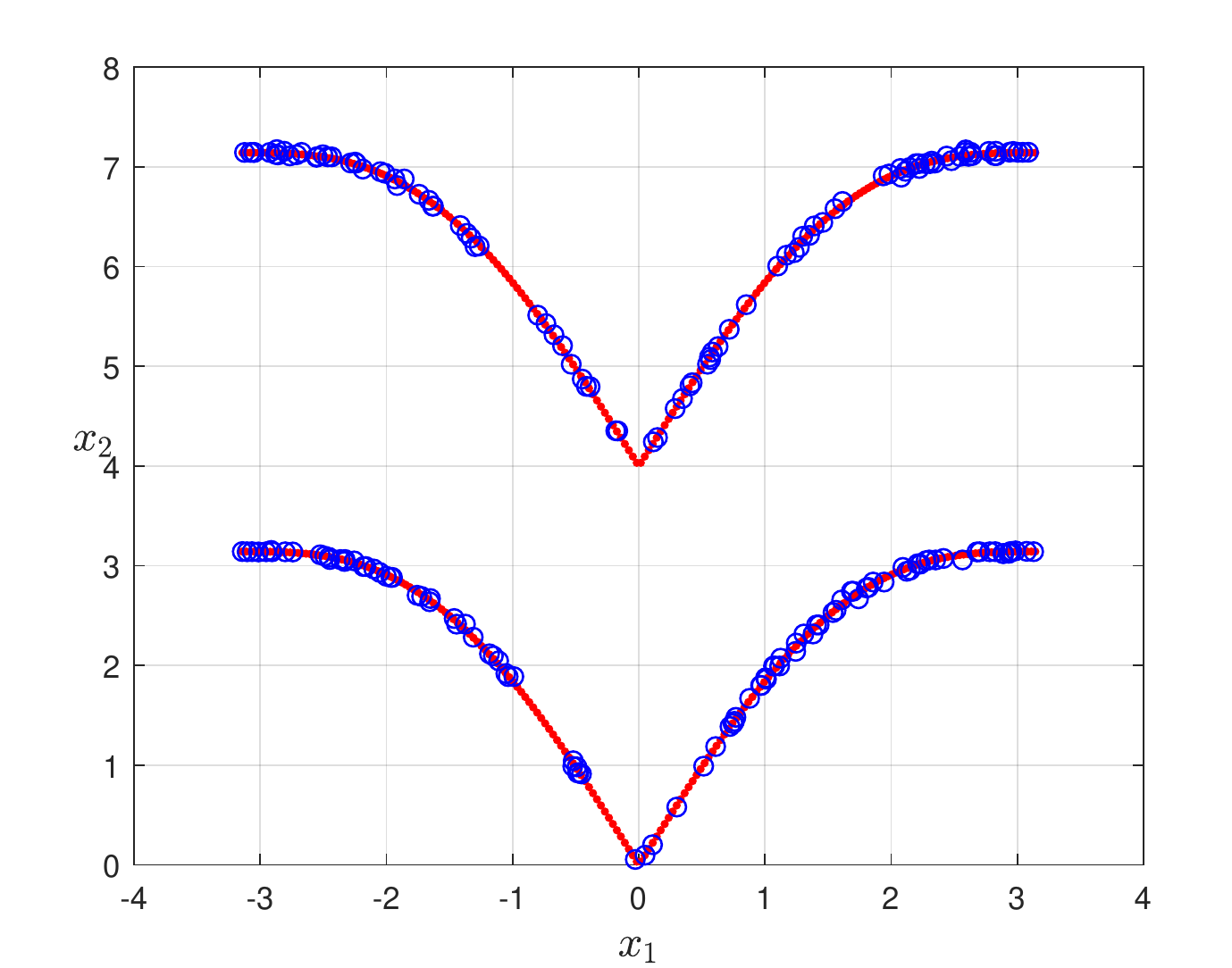}}
\subfigure[MMEA-WI]{\includegraphics[width=1.1in]{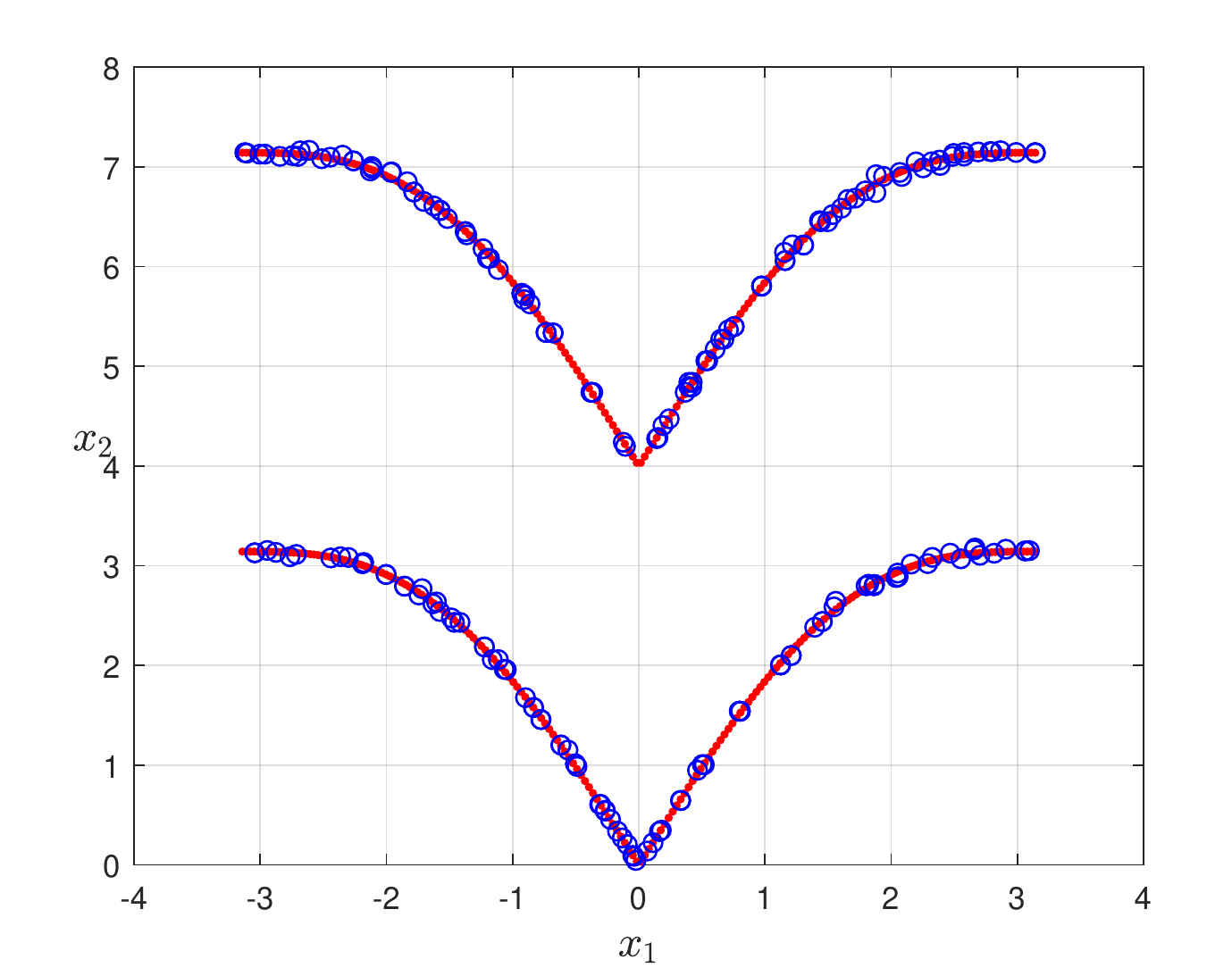}}
\subfigure[HREA]{\includegraphics[width=1.1in]{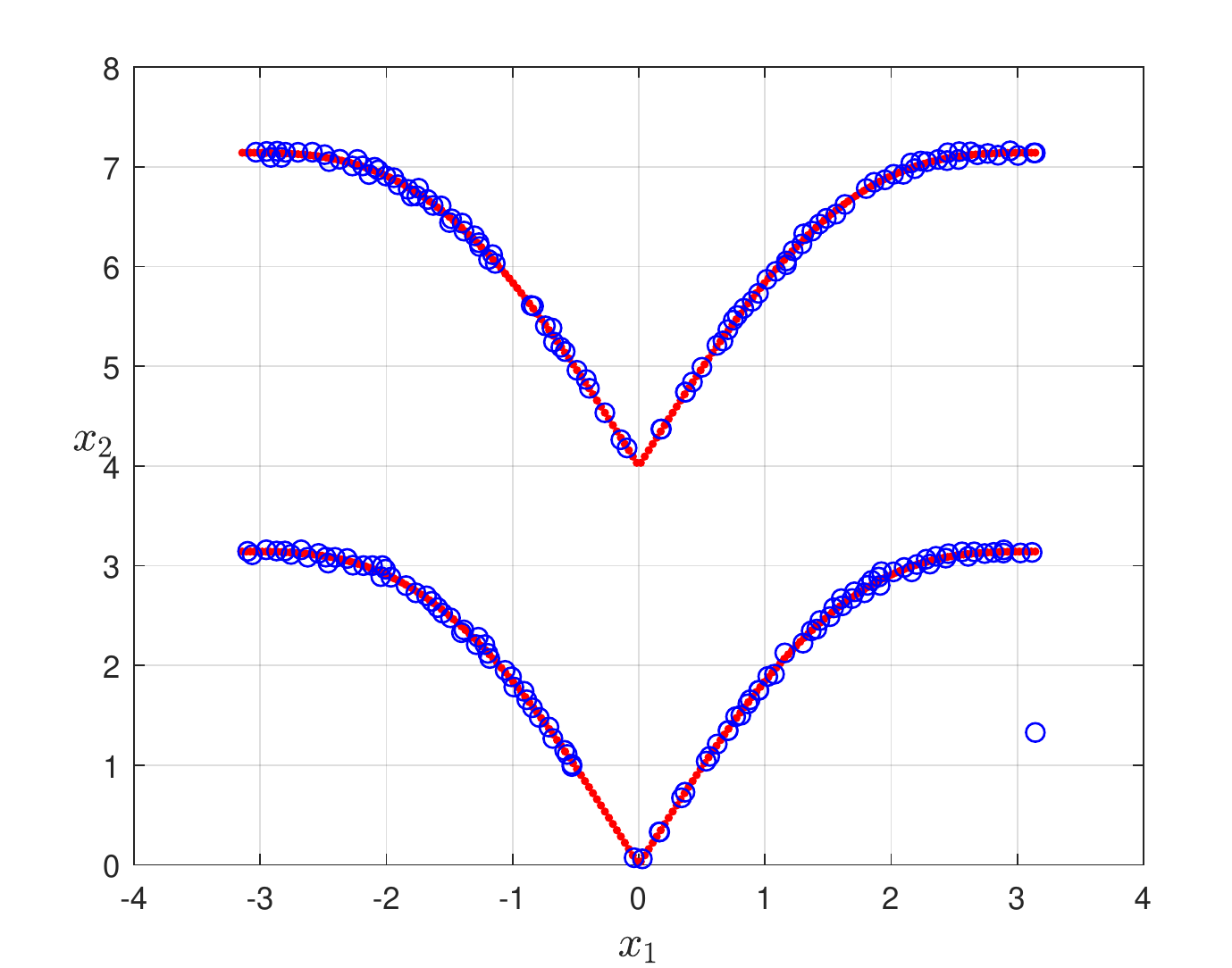}}
\caption{The distribution of solutions obtained by all algorithms (MO\_R\_PSO\_SCD is the short name for MO\_Ring\_PSO\_SCD) in the decision spaces on MMF4 (the first two rows) and MMF8 (the last two rows), where the red points and blue circles are true PS and obtained solutions respectively.}
\label{fig_mmfresult}
\end{figure*}

To further analyze the performance of all algorithms, \fref{fig_mmfresult} shows the final distribution of solutions on MMF4 and MMF8. As we can see, solutions obtained by MO\_PSO\_MM, CPDEA, MMEA-WI and HREA are more evenly distributed in the decision space. The same situation can be observed in other test problems. Readers can find more information in the supplementary file. To sum up, CPDEA is the best algorithm for the MMF test suite, while MO\_PSO\_MM and HREA are competitive.

\subsection{Performance comparison on IDMP problems}
\label{sec_idmpresult}
The main property of the IDMP test suite is that the difficulties of searching different PSs are different. Thus, normal MOEAs are more likely to converge to the easy-searching PS. Therefore, IDMP is a more accurate test suite to examine the diversity-maintaining ability of algorithms. 

\begin{figure}[t]
	\begin{center}
		\includegraphics[width=3.5in]{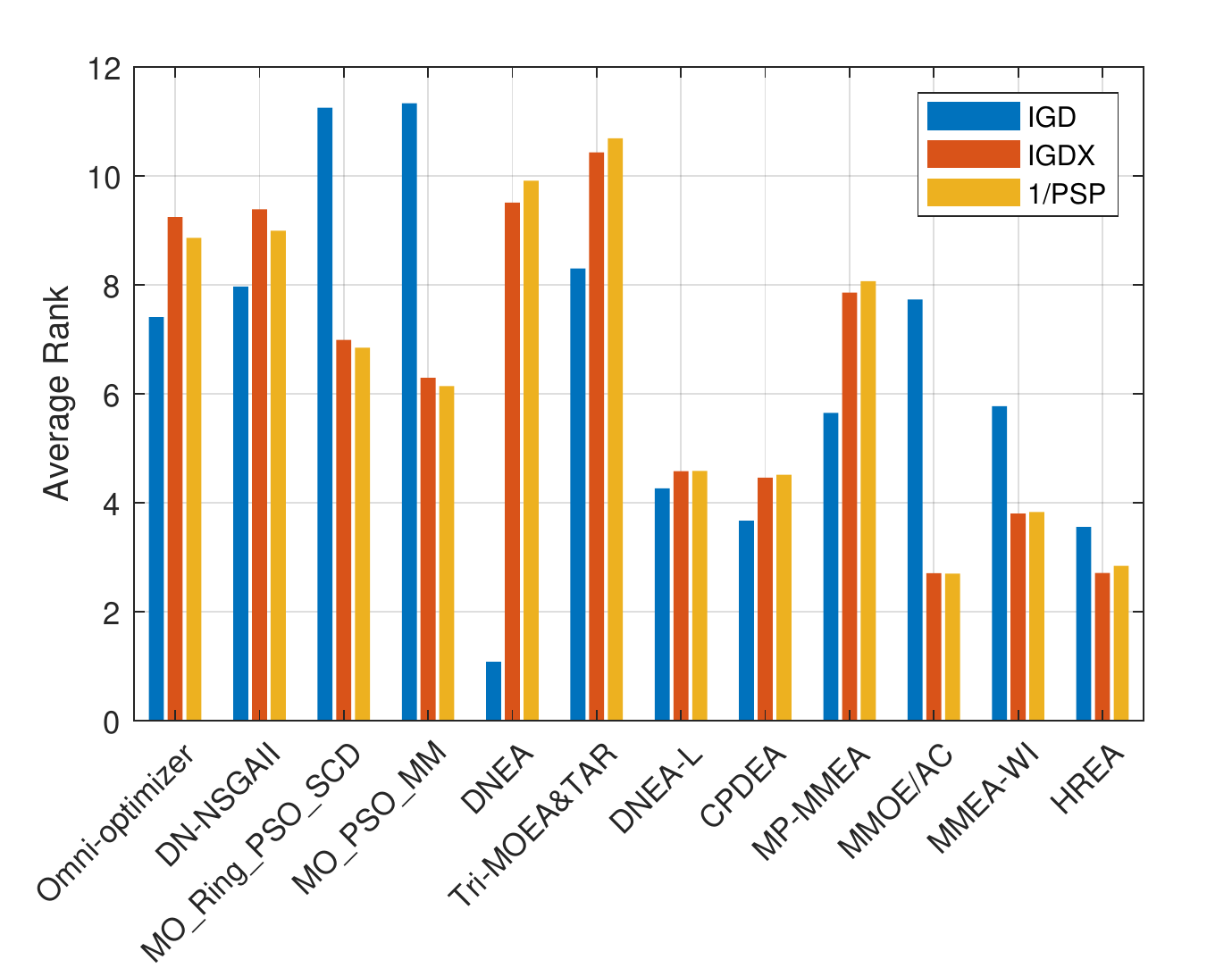}
		\caption{The average rank of all compared MMEAs on IDMP test suite in terms of $IGD$, $IGDX$ and $1/PSP$.}
		\label{fig_avgrankidmp}
	\end{center}
\end{figure}

The overall performance rank is presented in \fref{fig_avgrankidmp}, from which we can see that in terms of $IGDX$ and $1/PSP$, HREA and MMOEA/DC are in the first echelon, followed by DNEA-L, MMEA-WI and CPDEA. As we can see, algorithms considering local PSs (DNEA-L, MMOEA/DC and HREA) show overwhelming superiority in such problems. They can easily obtain all PSs for most of the IDMP test problems, except for some problems with 4 objectives, while other primitive algorithms can only obtain some of the PSs. In general, algorithms proposed before DNEA-L perform poorly on the IDMP test suite. Primitive works did not consider the situation that different difficulties in searching for different PSs. The same conclusion can be found in \cite{liu2019handling}. Table S-VI, Table S-VII and Table S-VIII list the detailed $IGD$, $IGDX$ and $1/PSP$ results of all MMEAs on IDMP. In terms of $IGDX$ and $1/PSP$, HREA, MMOEA/DC, MMEA-WI and MO\_PSO\_MM win 5, 5, 1 and 1 instances respectively. It's interesting that MO\_PSO\_MM shows a great result on IDMPM2T4. Compared to other IDMP test problems, the difference in difficulty between searching for different PSs is smaller in IDMPM2T4. Thus, MO\_PSO\_MM can stably obtain all PSs on IDMPM2T4.

As for $IGD$, DNEA shows its dominance since it wins all 12 test problems. That really means DNEA is an effective and competitive MOEA for solving IDMP problems. As a comparison, MO\_PSO\_MM and MO\_Ring\_PSO\_SCD receive the worst average ranks. Results in Table S-V show that these two algorithms perform the worst on almost all test problems. The diversity in the decision space is overemphasized for these two algorithms. Another two poorly performing algorithms are Tri-MOEA\&TAR and MP-MMEA, in which variable analysis method is adopted to improve the convergence ability.

\fref{fig_avgrankidmp} presents the final distribution of solutions in the decision space. Intuitively, solutions obtained by DNEA-L, CPDEA, MMOEA/DC, MMEA-WI and HREA distribute more evenly. To sum up, HREA and MMOEA/DC are the two best algorithms for IDMP and primitive proposed algorithms are unable to obtain all PSs. Algorithms with a local-PS-maintaining strategy have a stronger ability in finding and obtaining PSs.

\begin{figure*}[htbp]
\centering
\subfigure[Omni-optimizer]{\includegraphics[width=1.1in]{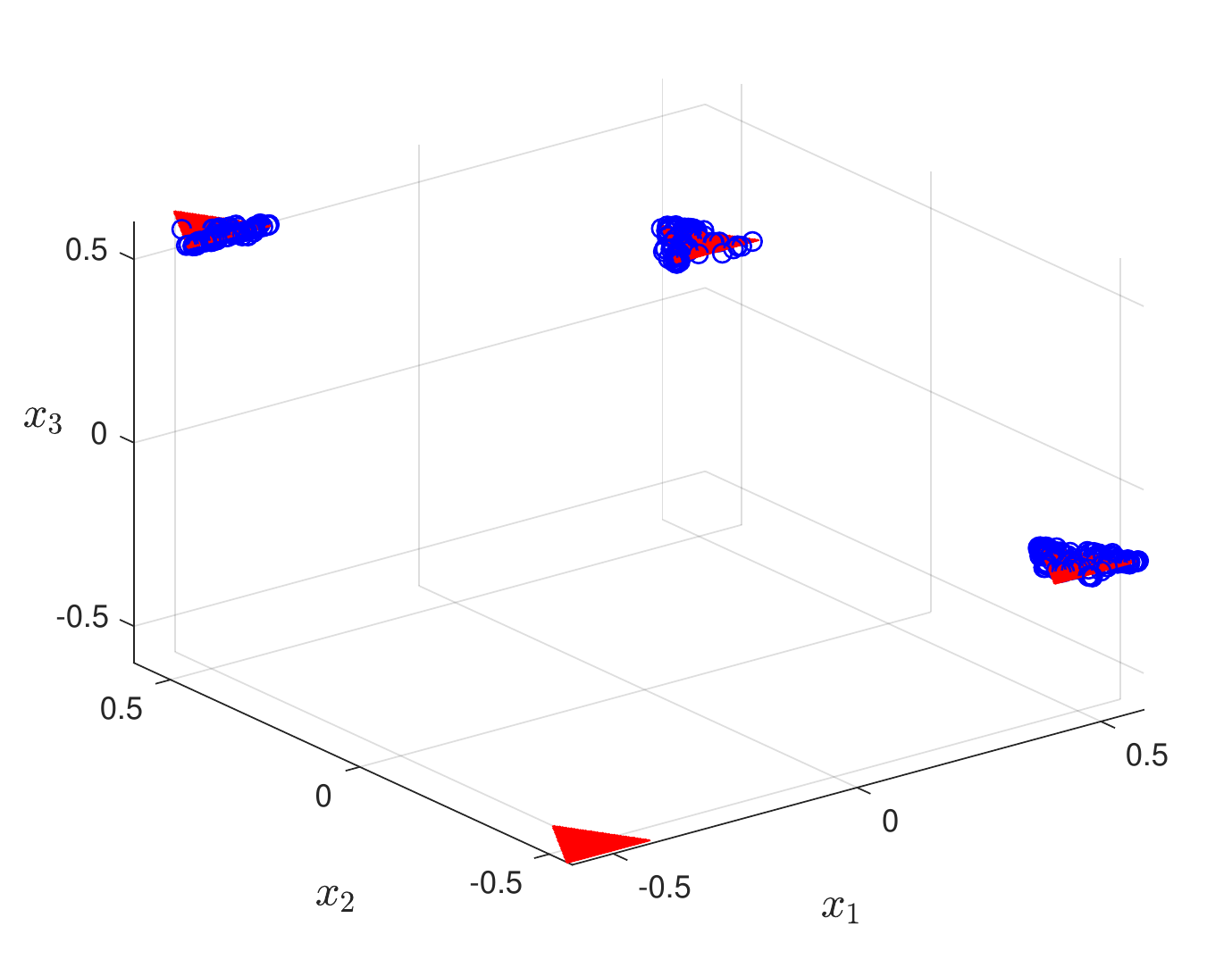}}
\subfigure[DN-NSGAII]{\includegraphics[width=1.1in]{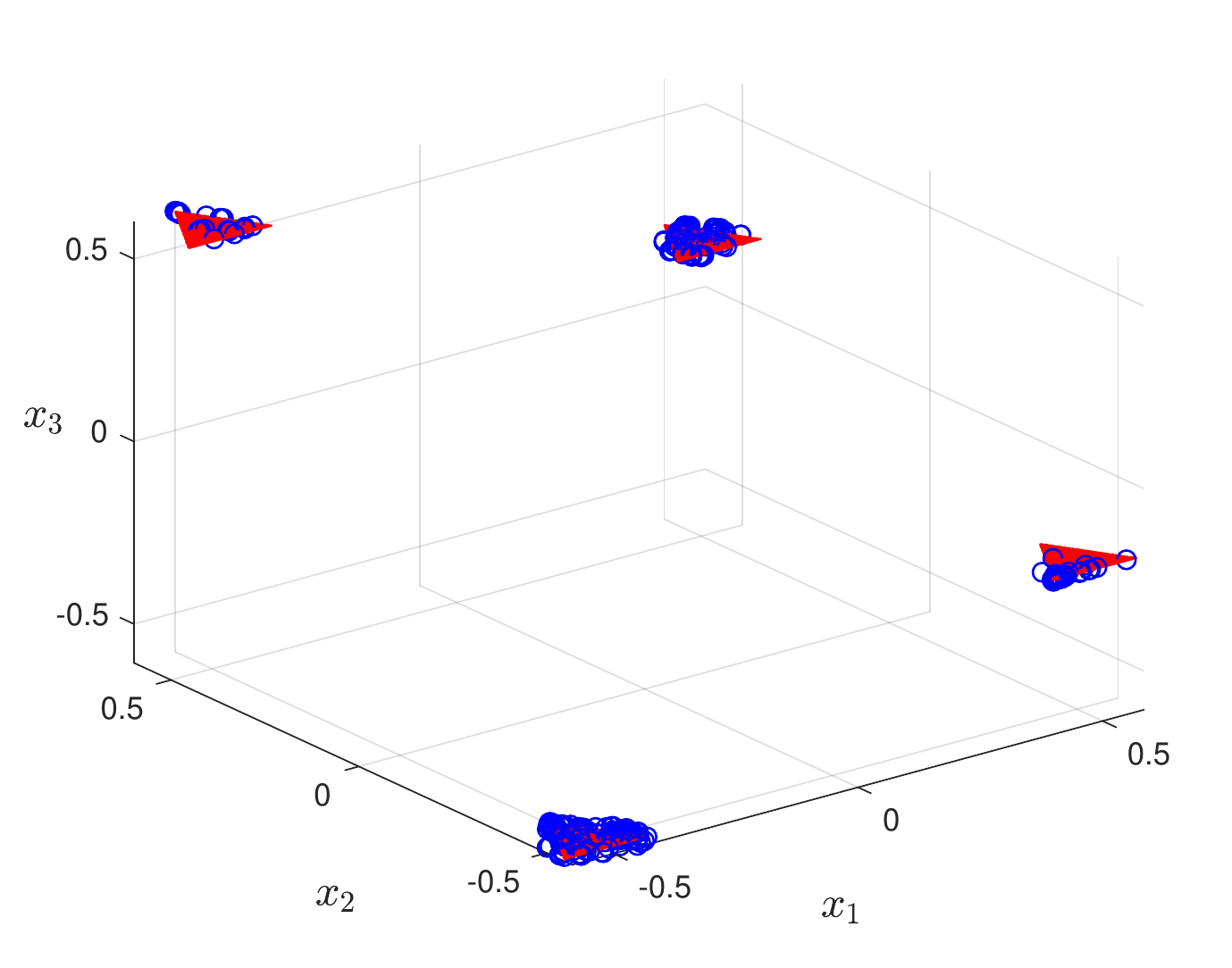}}
\subfigure[MO\_R\_PSO\_SCD]{\includegraphics[width=1.1in]{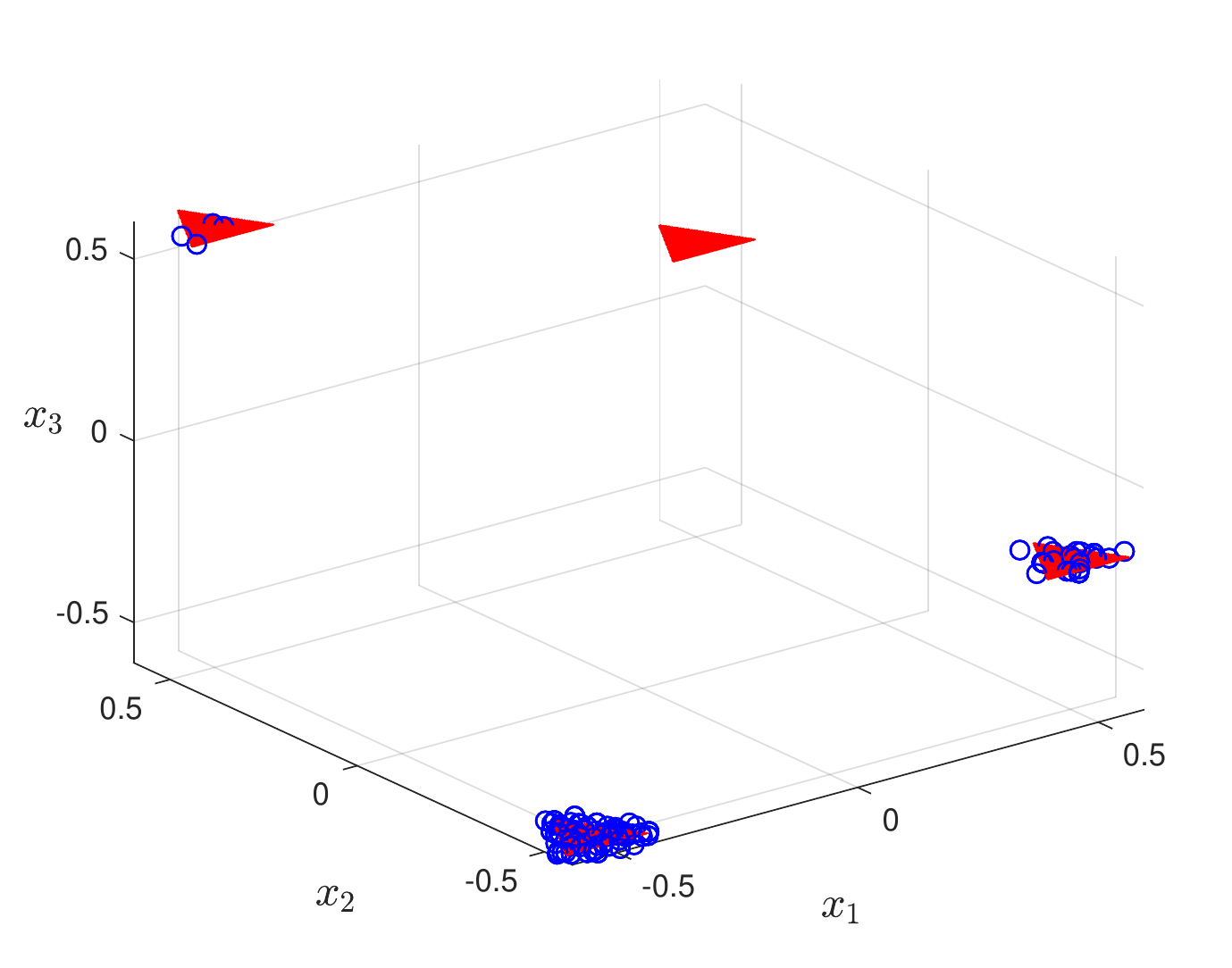}}
\subfigure[MO\_PSO\_MM]{\includegraphics[width=1.1in]{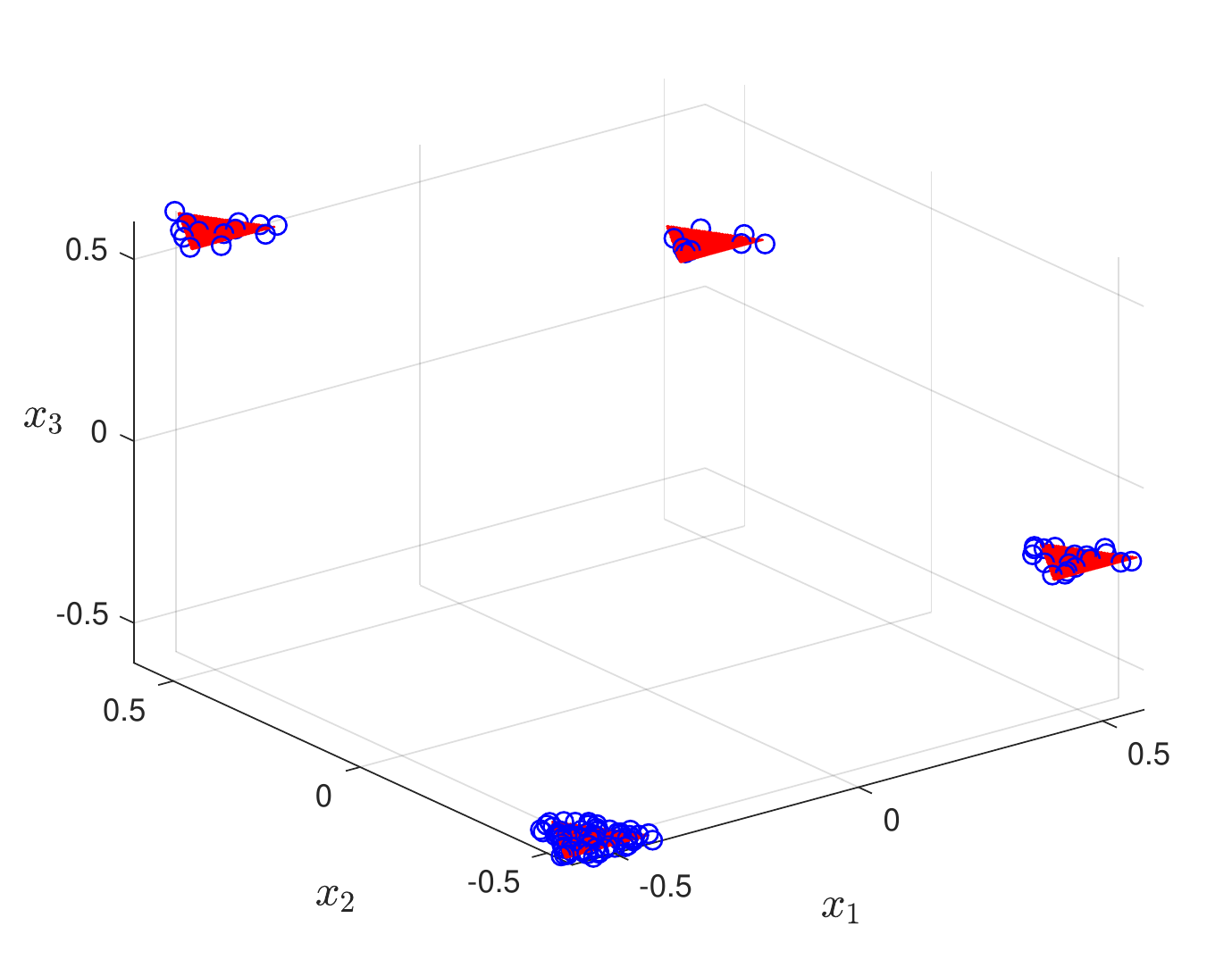}}
\subfigure[DNEA]{\includegraphics[width=1.1in]{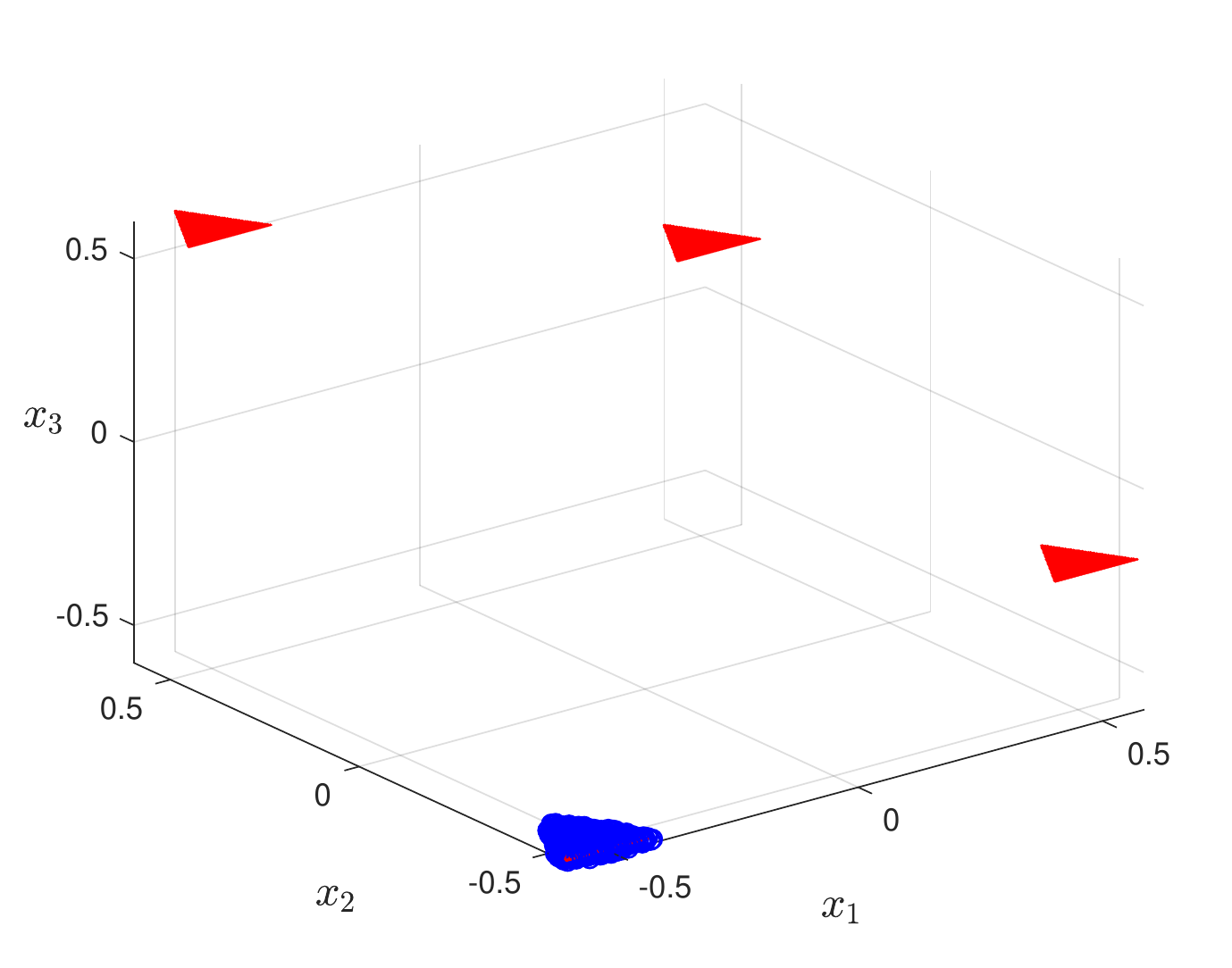}}
\subfigure[Tri-MOEA\&TAR]{\includegraphics[width=1.1in]{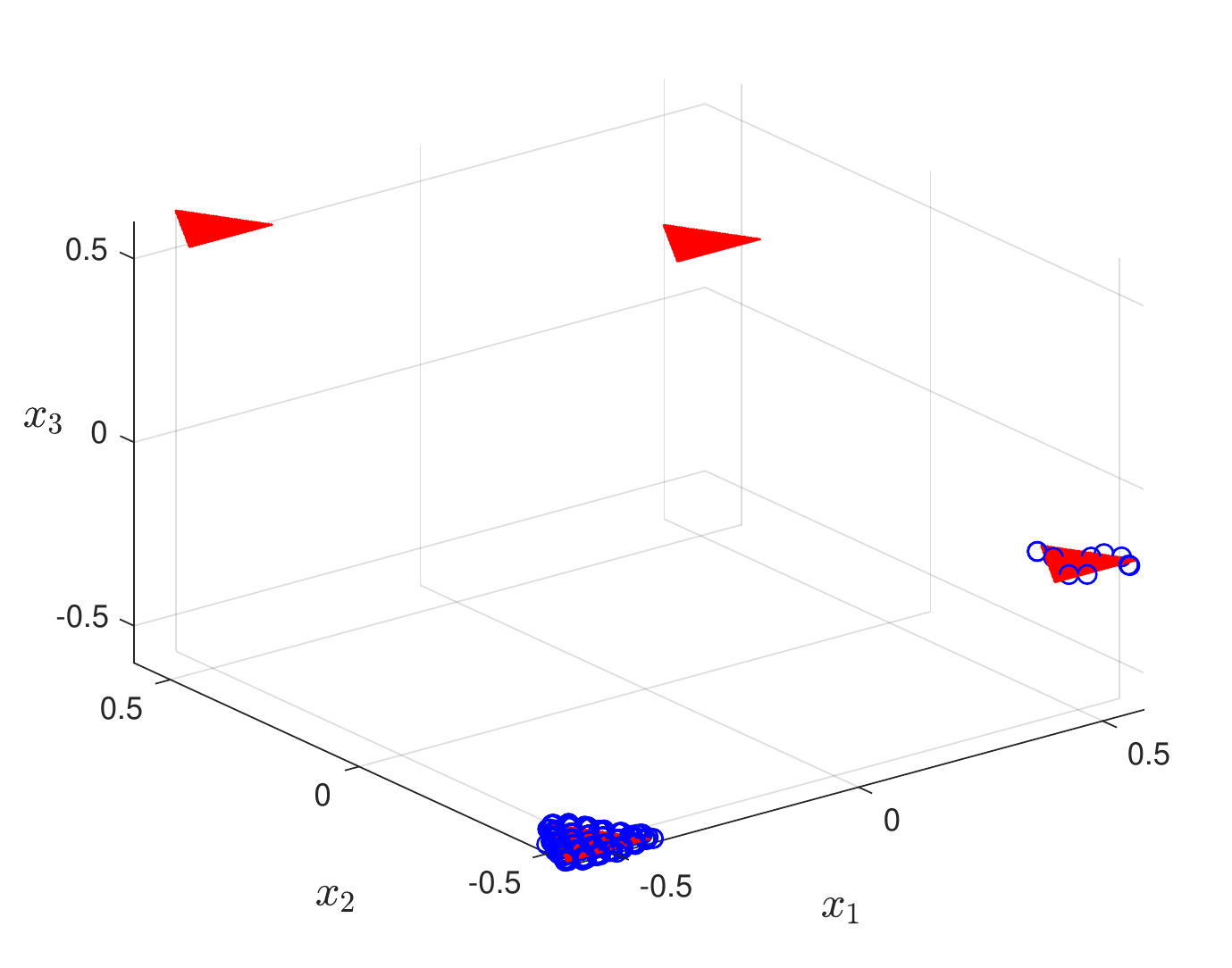}}
\subfigure[DNEA-L]{\includegraphics[width=1.1in]{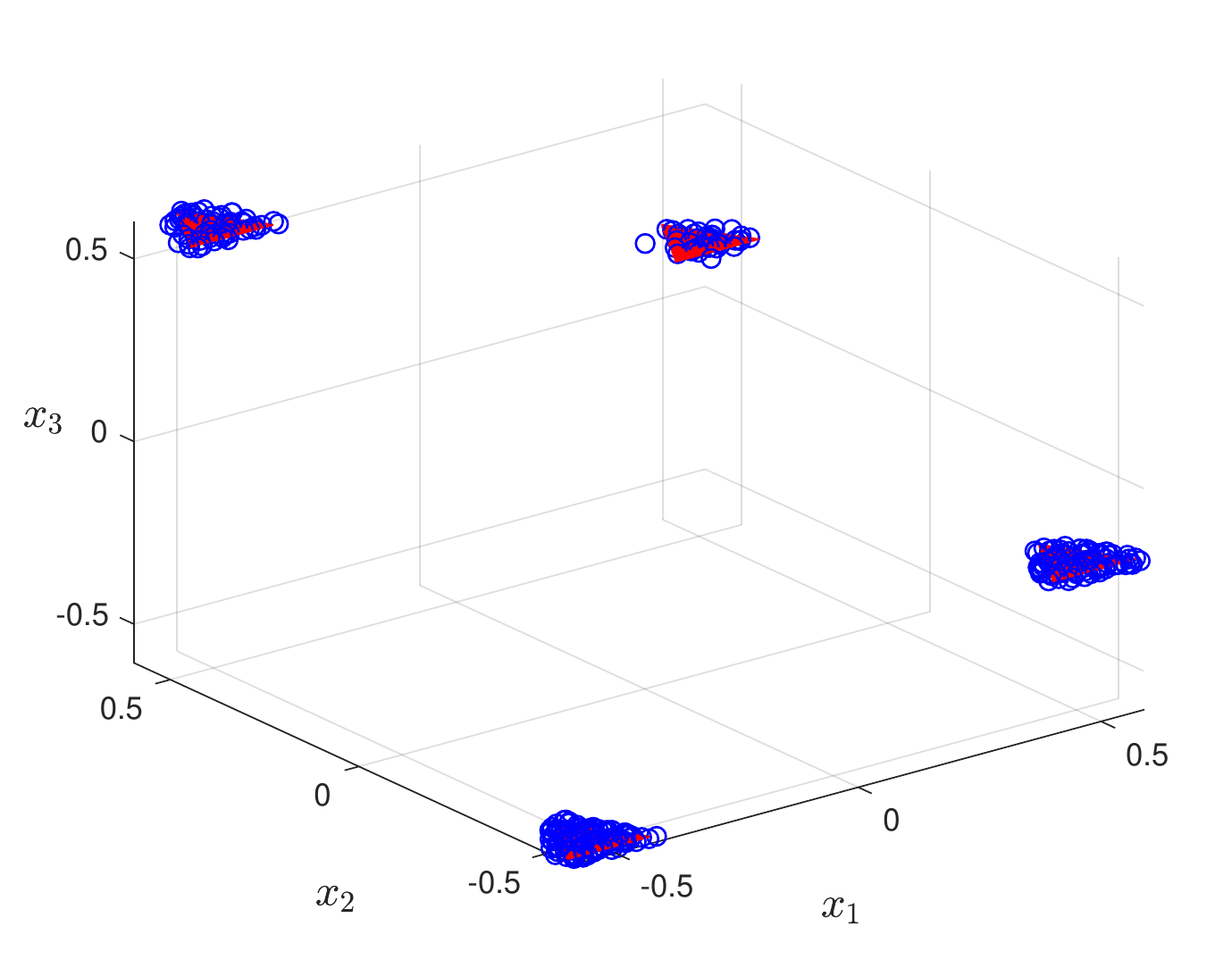}}
\subfigure[CPDEA]{\includegraphics[width=1.1in]{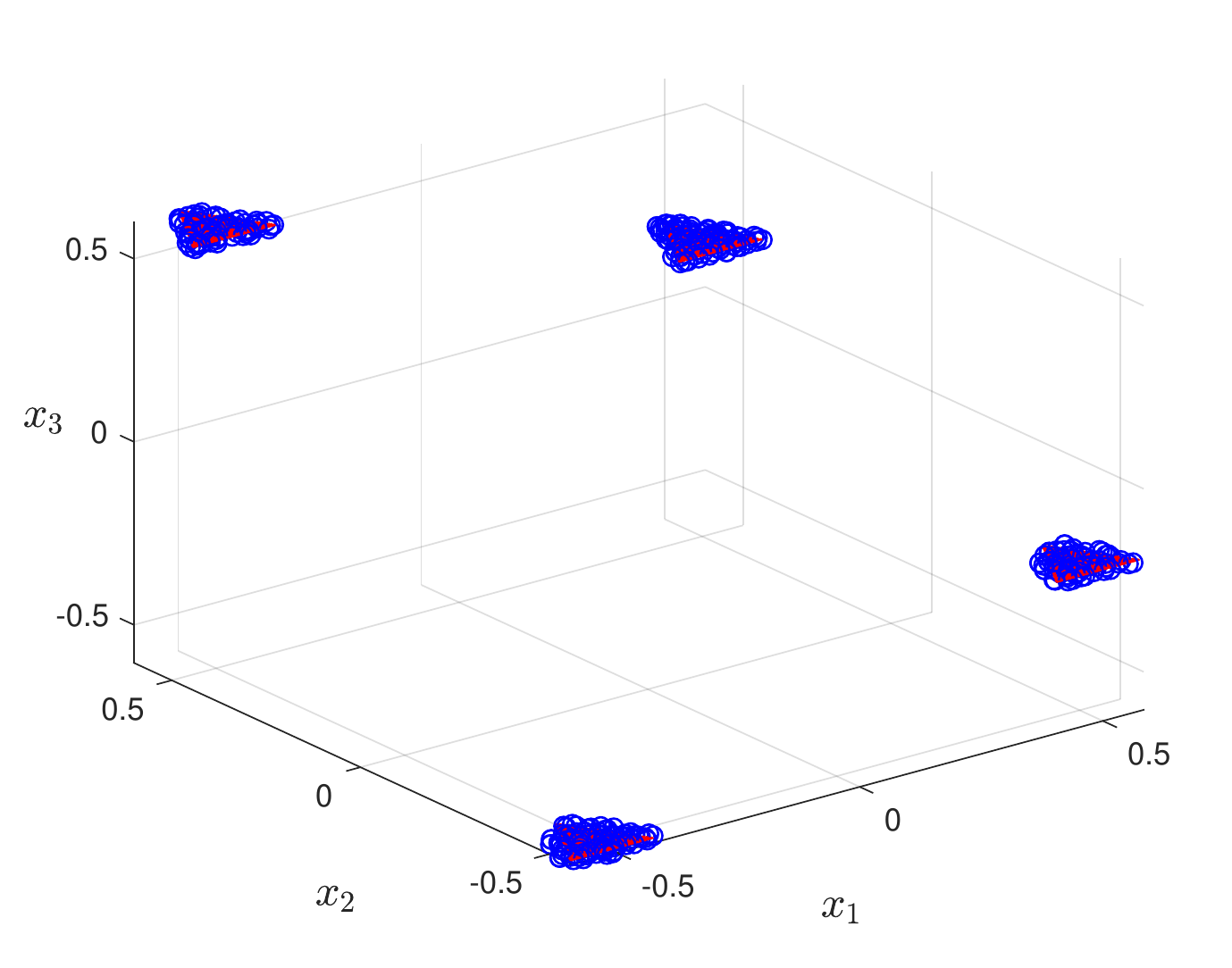}}
\subfigure[MP-MMEA]{\includegraphics[width=1.1in]{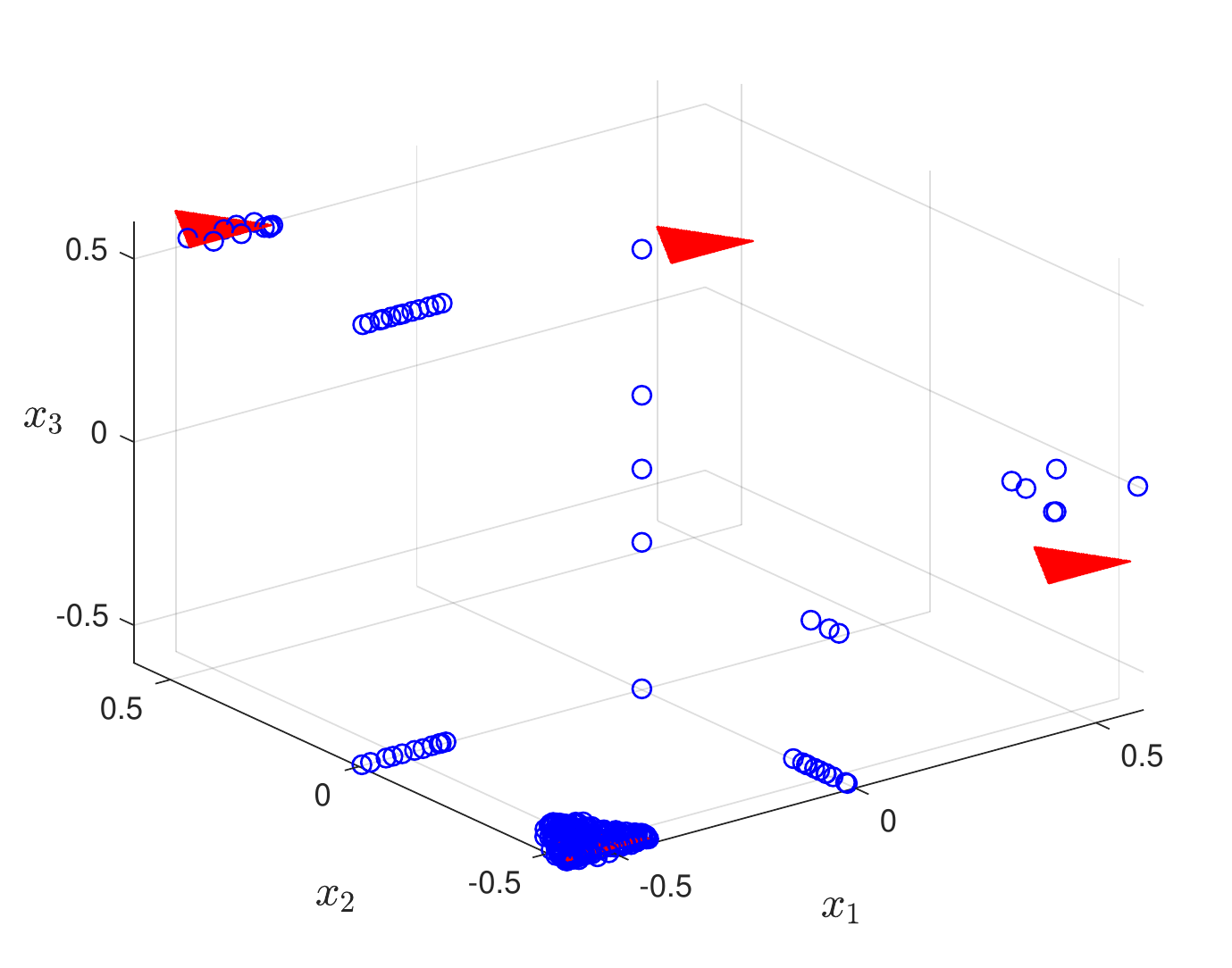}}
\subfigure[MMOEA/DC]{\includegraphics[width=1.1in]{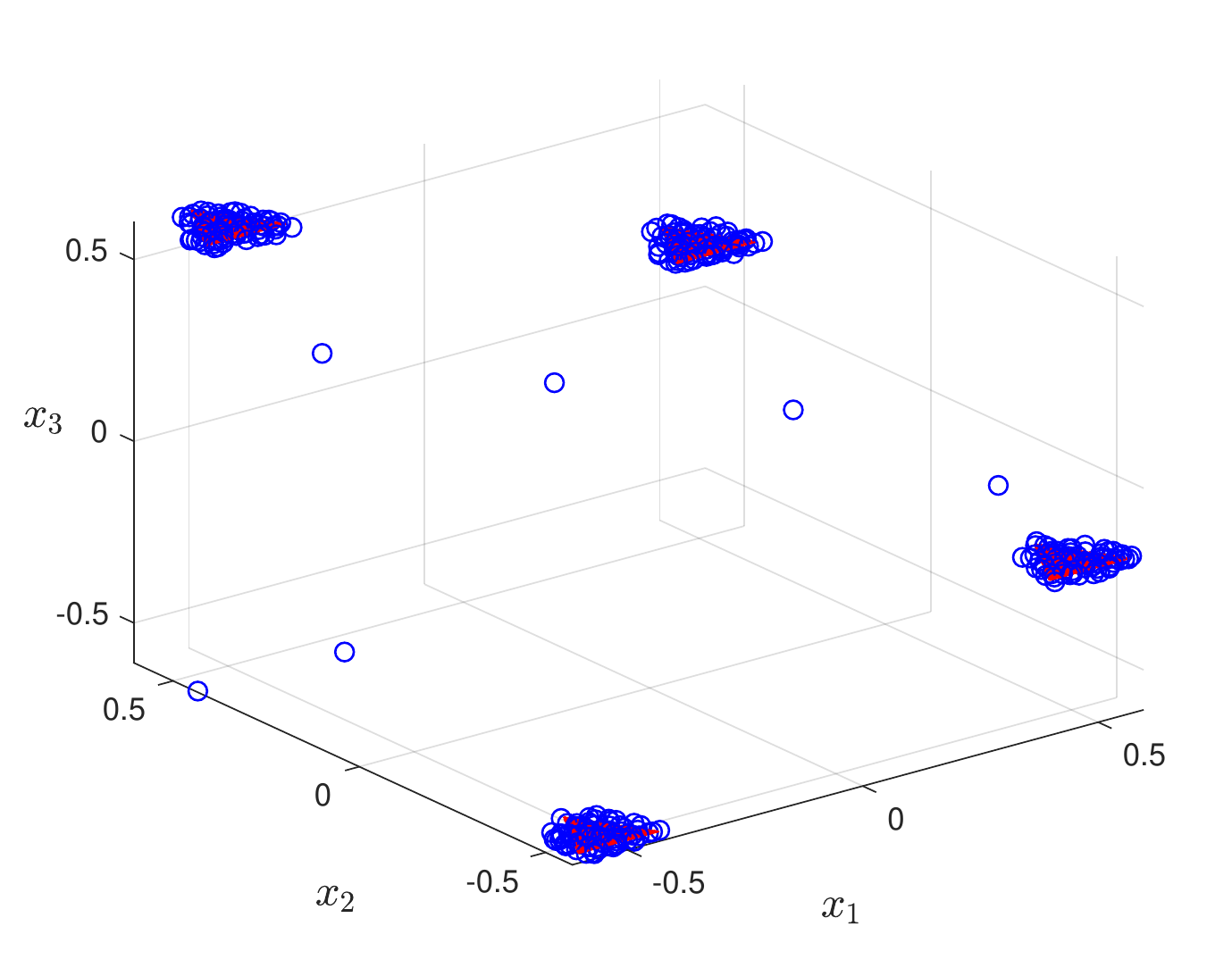}}
\subfigure[MMEA-WI]{\includegraphics[width=1.1in]{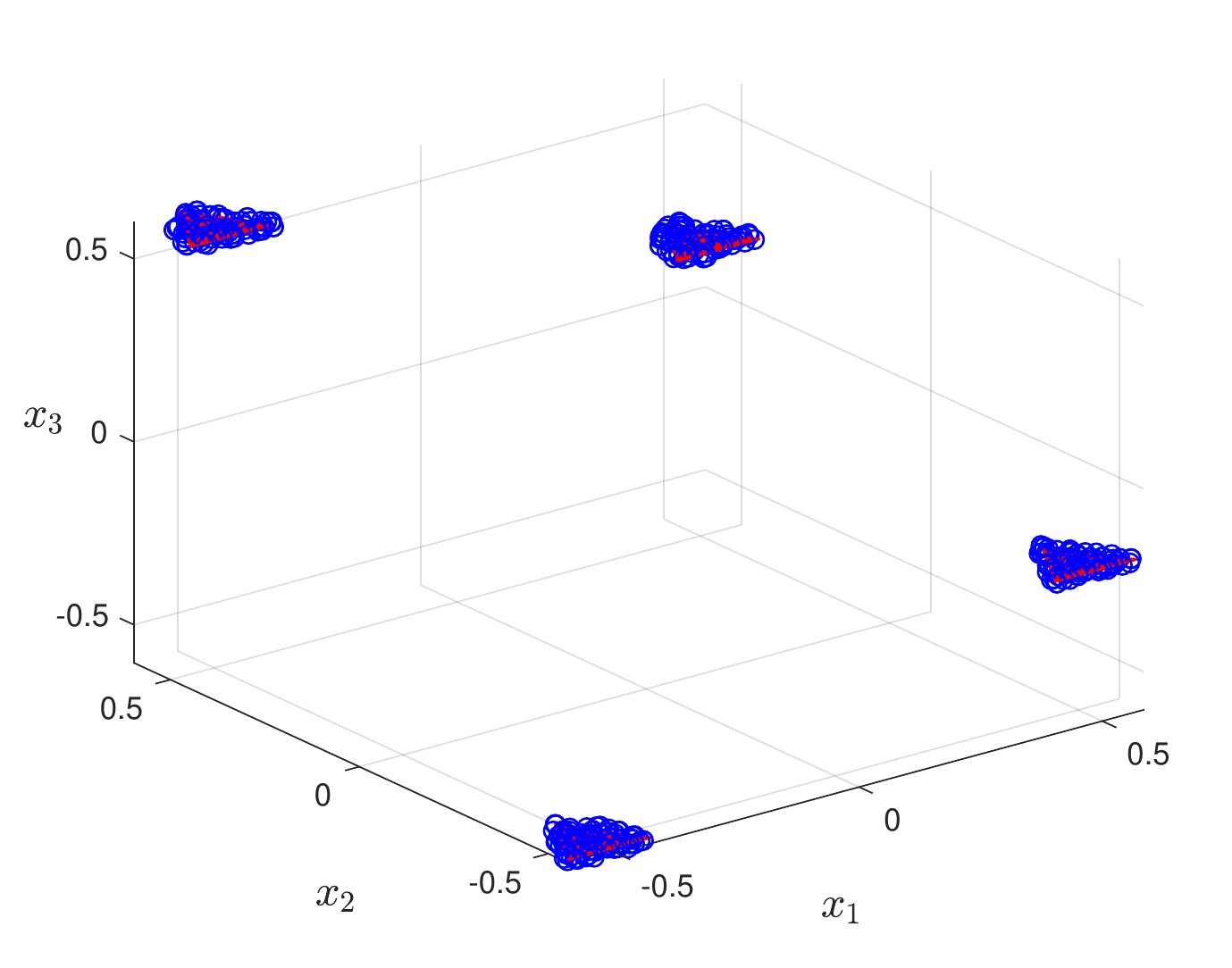}}
\subfigure[HREA]{\includegraphics[width=1.1in]{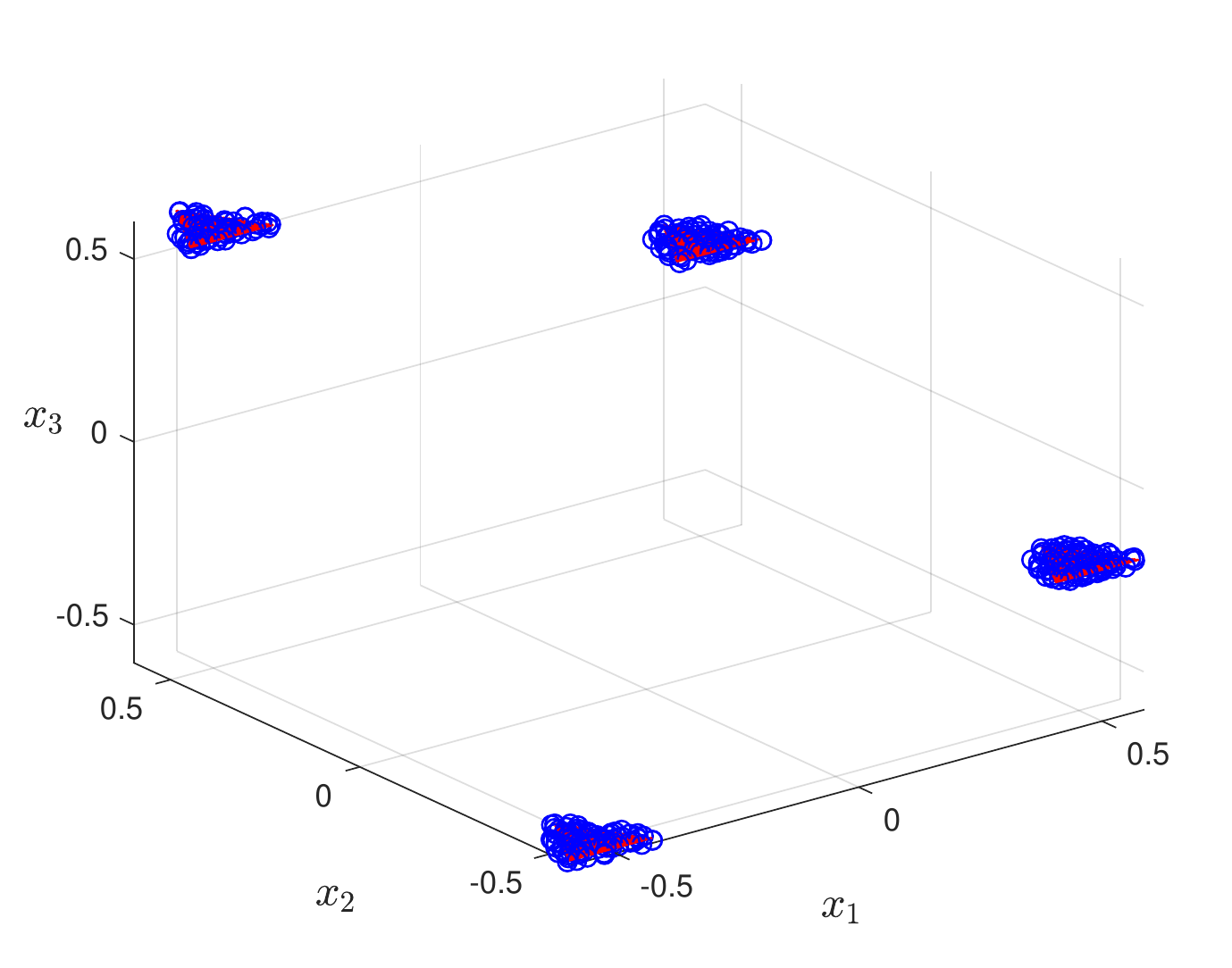}}
\caption{The distribution of solutions obtained by all algorithms (MO\_R\_PSO\_SCD is the short name for MO\_Ring\_PSO\_SCD) in the decision spaces on IDMPM3T4, where the red points and blue circles are true PS and obtained solutions respectively.}
\label{fig_idmpresult}
\end{figure*}

\subsection{Performance comparison on problems with local PFs}
\label{sec_mmoplresult}

In this part, the performance of MMEAs on MMOPLs is discussed. To be specific, IDMP\_e and some of the MMF test problems (MMF10-MMF13, MMF15) are chosen as the test problems. Since there is no performance indicator designed for MMOPLs, we regard both global and local PSs as the true PS for calculating $IGD$, $IGDX$ and $1/PSP$.

\begin{figure}[tbph]
	\begin{center}
		\includegraphics[width=3.5in]{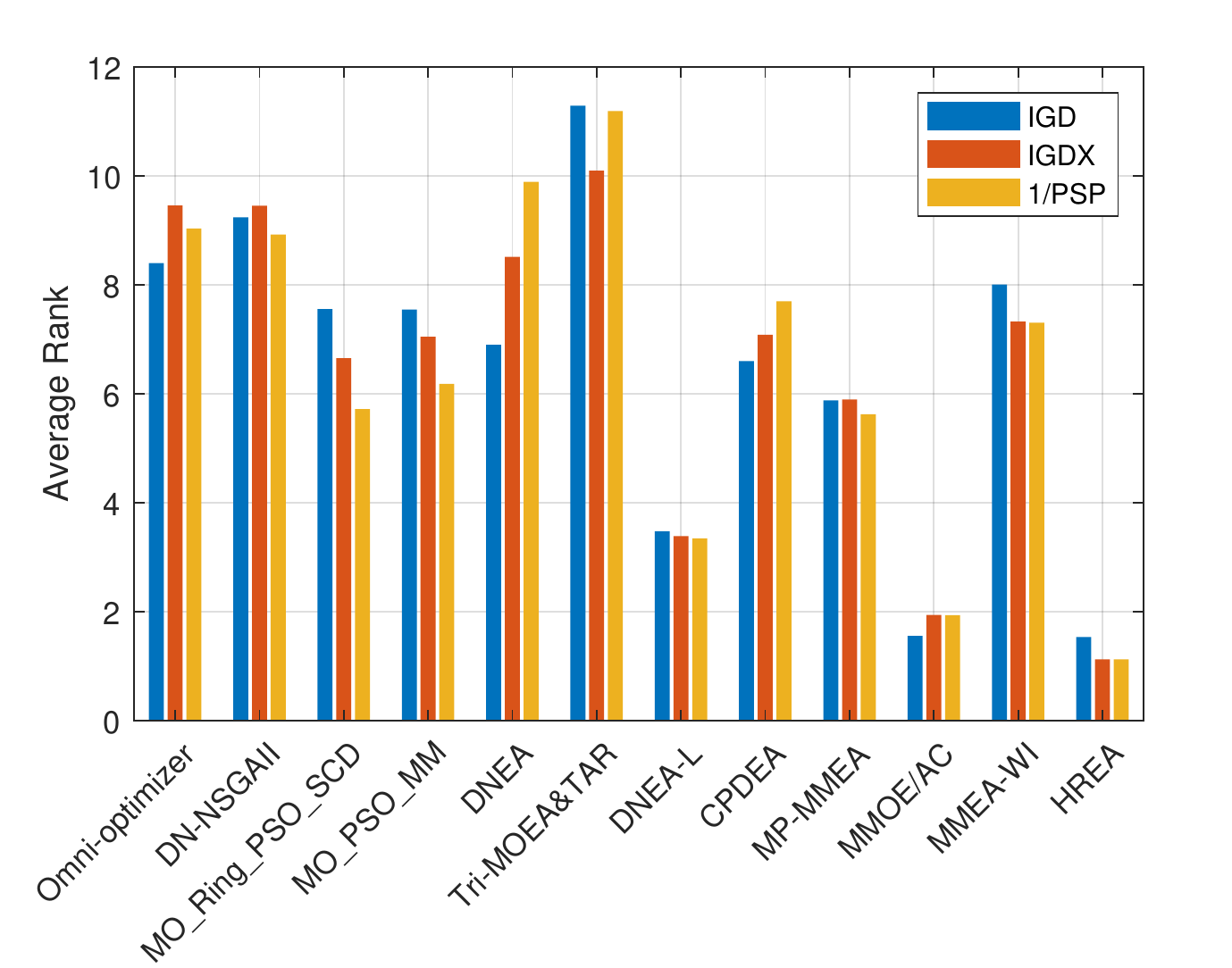}
		\caption{The average rank of all compared MMEAs on MMOPL test problems (IDMP\_e MMF10-MMF13 and MMF15) in terms of $IGD$, $IGDX$ and $1/PSP$.}
		\label{fig_avgrankidmpe}
	\end{center}
\end{figure}

\begin{figure*}[htbp]
\centering
\subfigure[Omni-optimizer]{\includegraphics[width=1.1in]{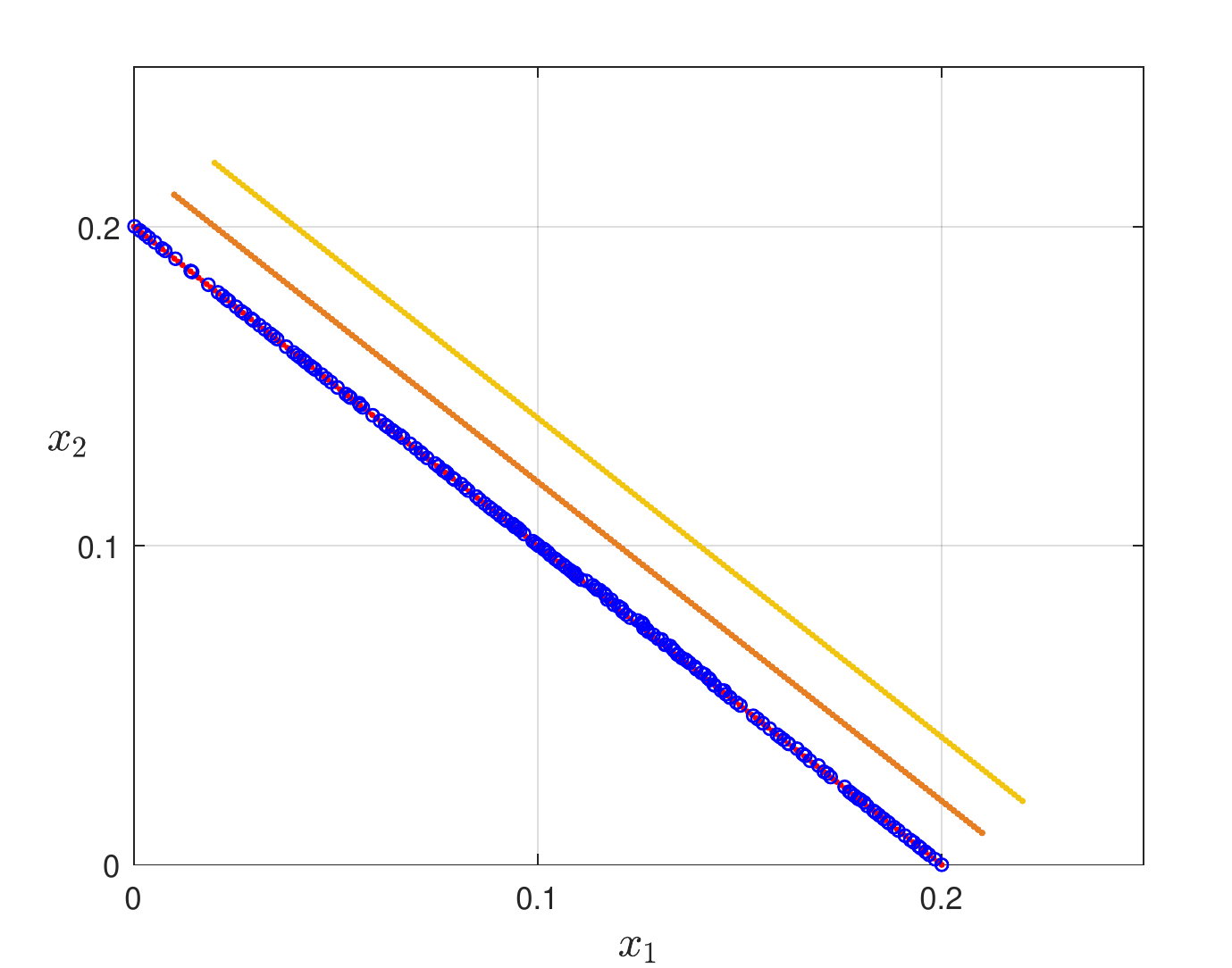}}
\subfigure[DN-NSGAII]{\includegraphics[width=1.1in]{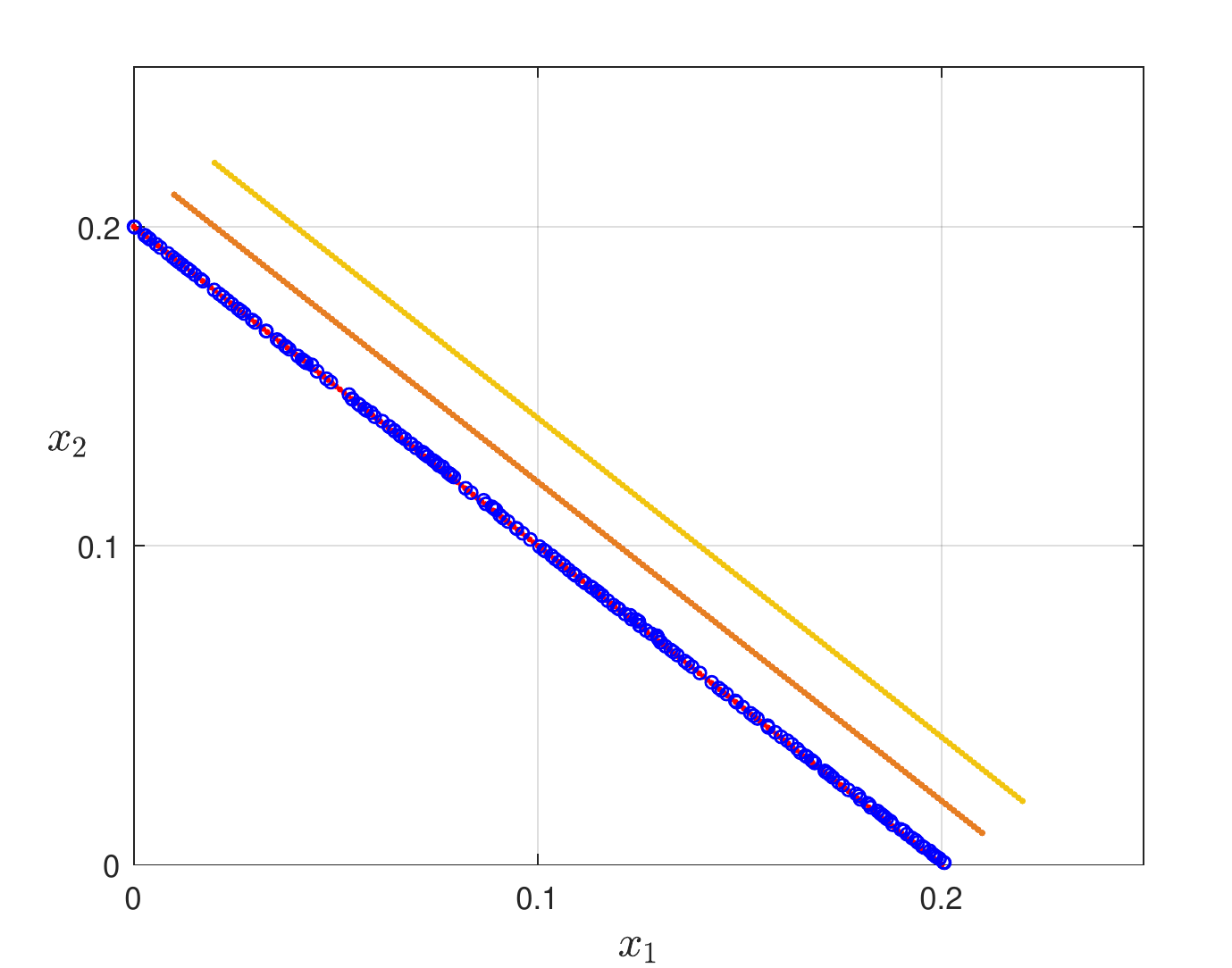}}
\subfigure[MO\_R\_PSO\_SCD]{\includegraphics[width=1.1in]{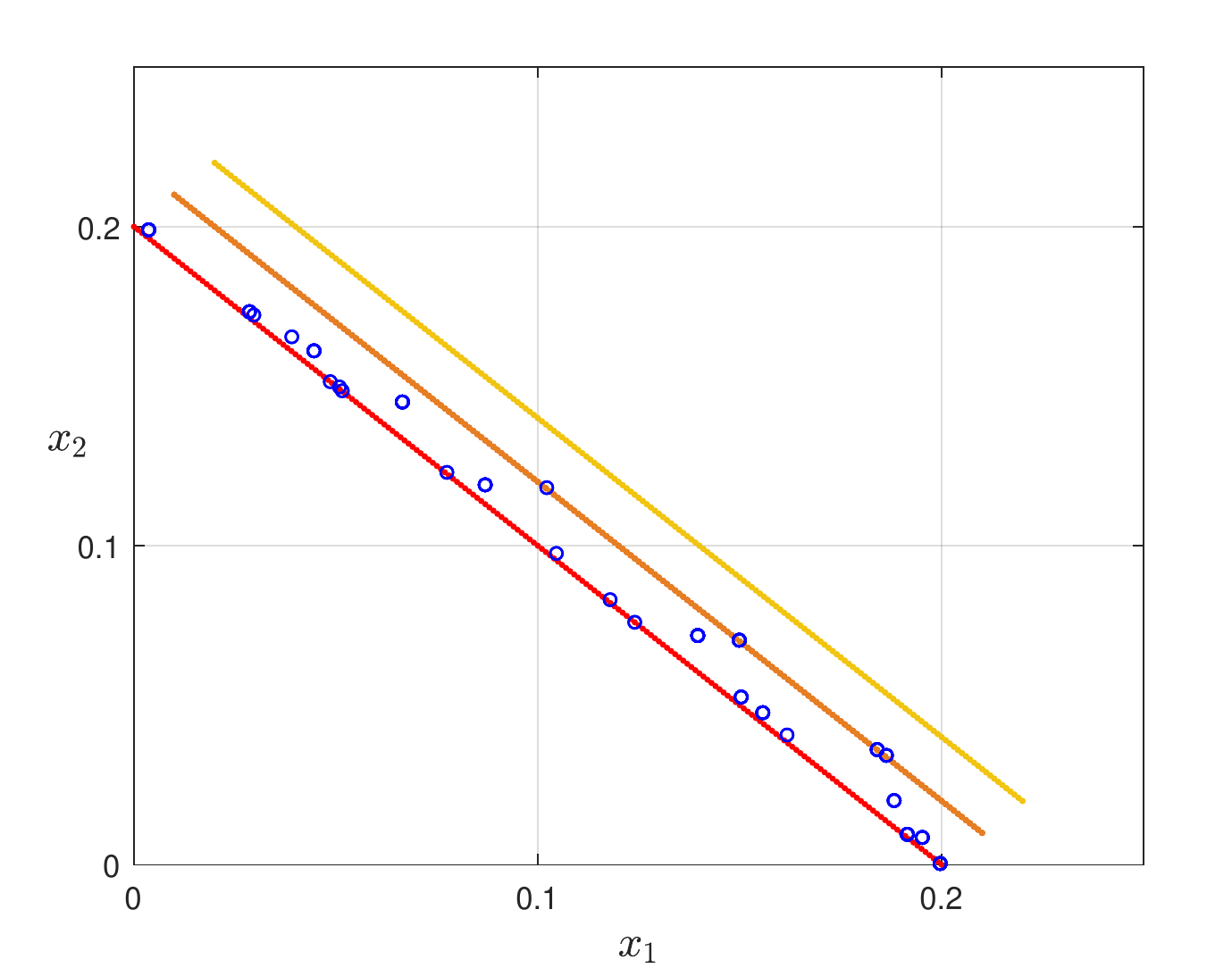}}
\subfigure[MO\_PSO\_MM]{\includegraphics[width=1.1in]{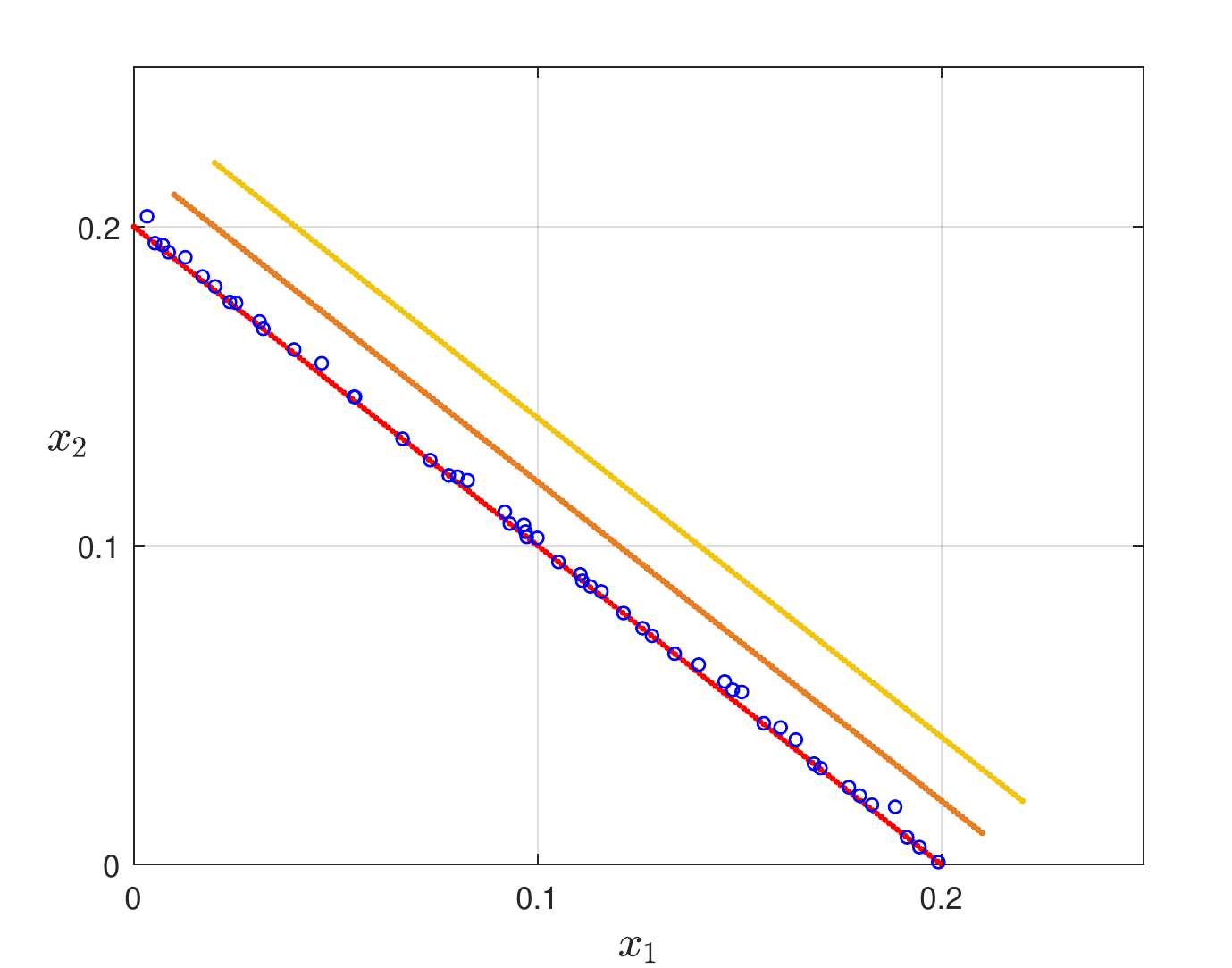}}
\subfigure[DNEA]{\includegraphics[width=1.1in]{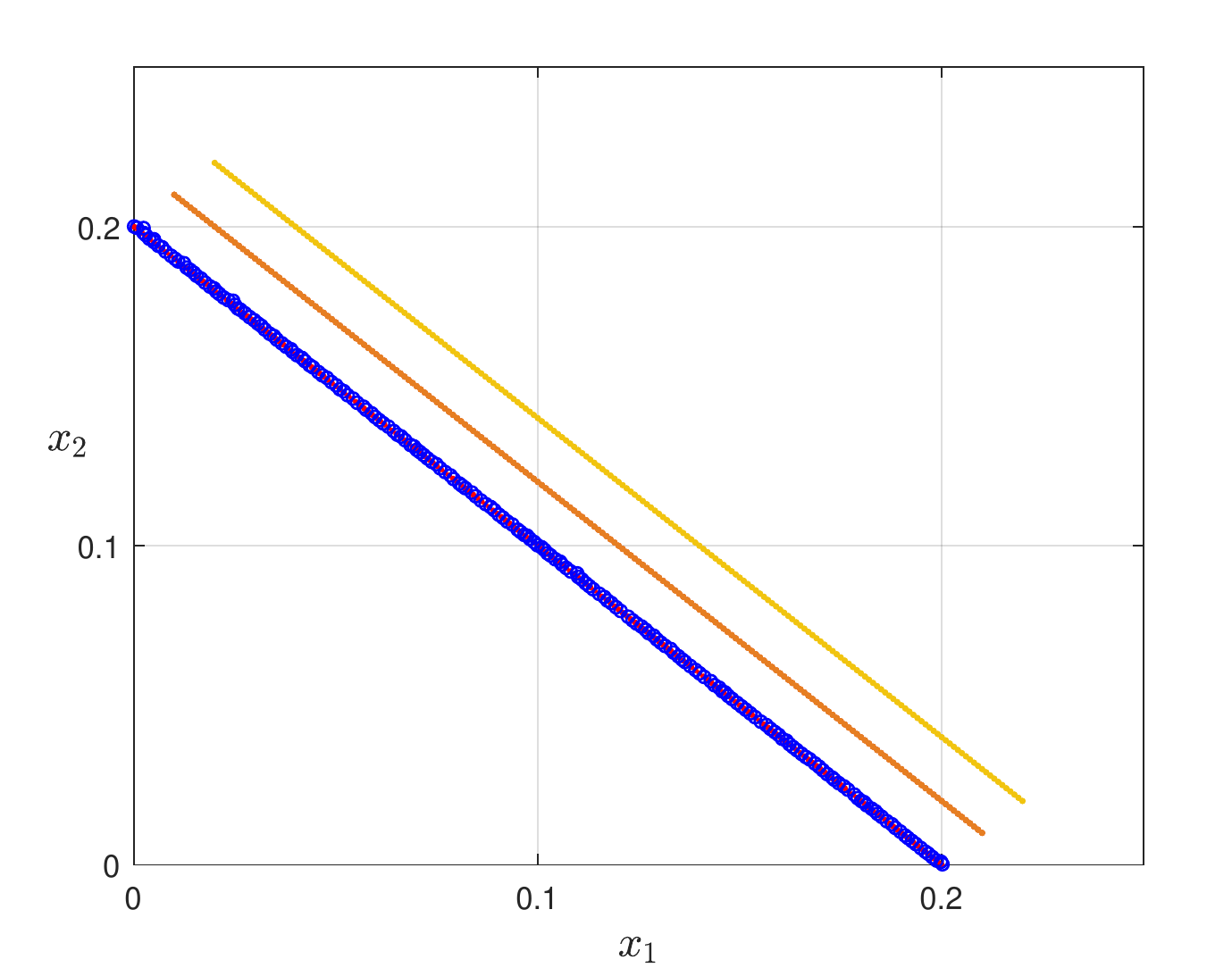}}
\subfigure[Tri-MOEA\&TAR]{\includegraphics[width=1.1in]{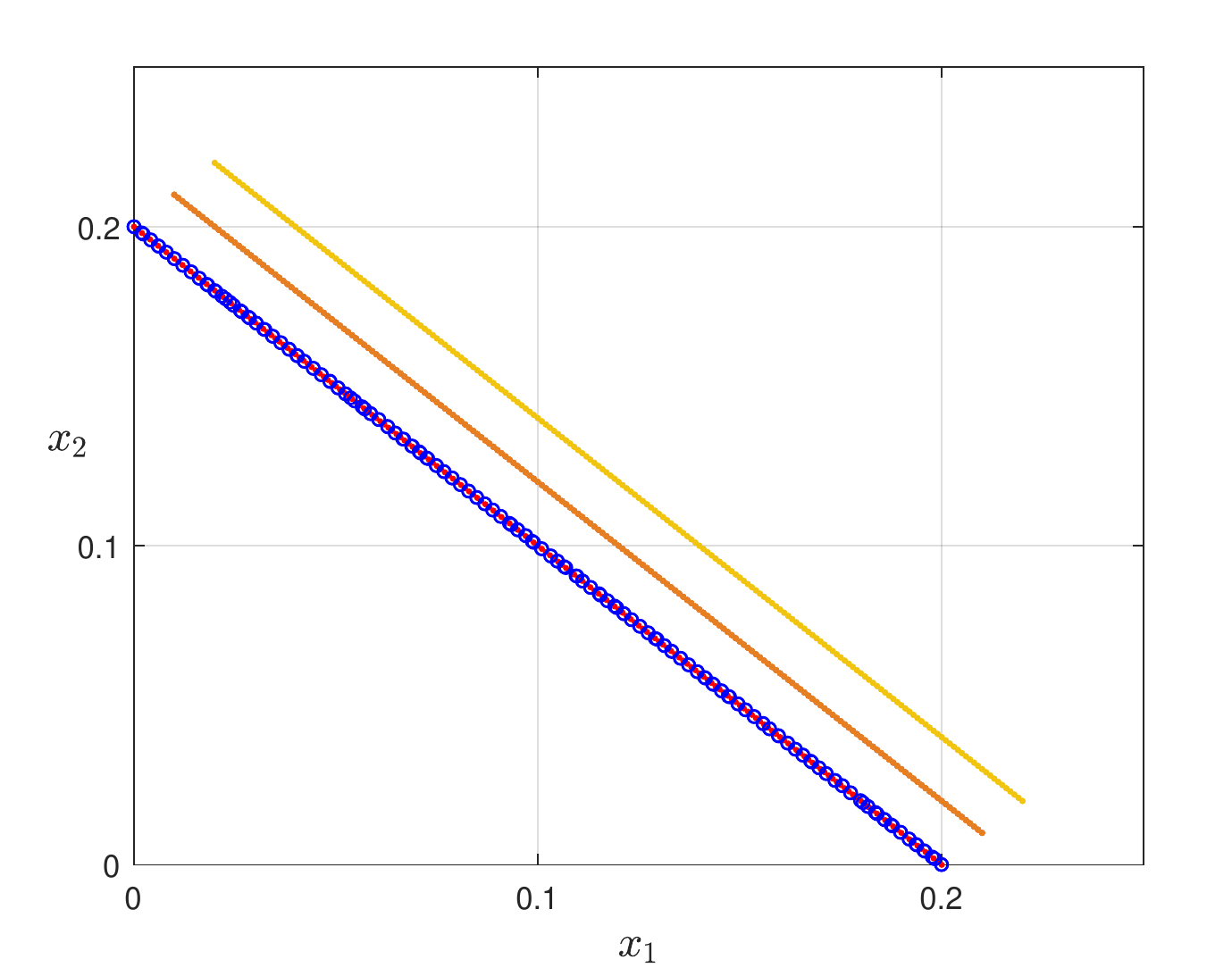}}

\subfigure[Omni-optimizer]{\includegraphics[width=1.1in]{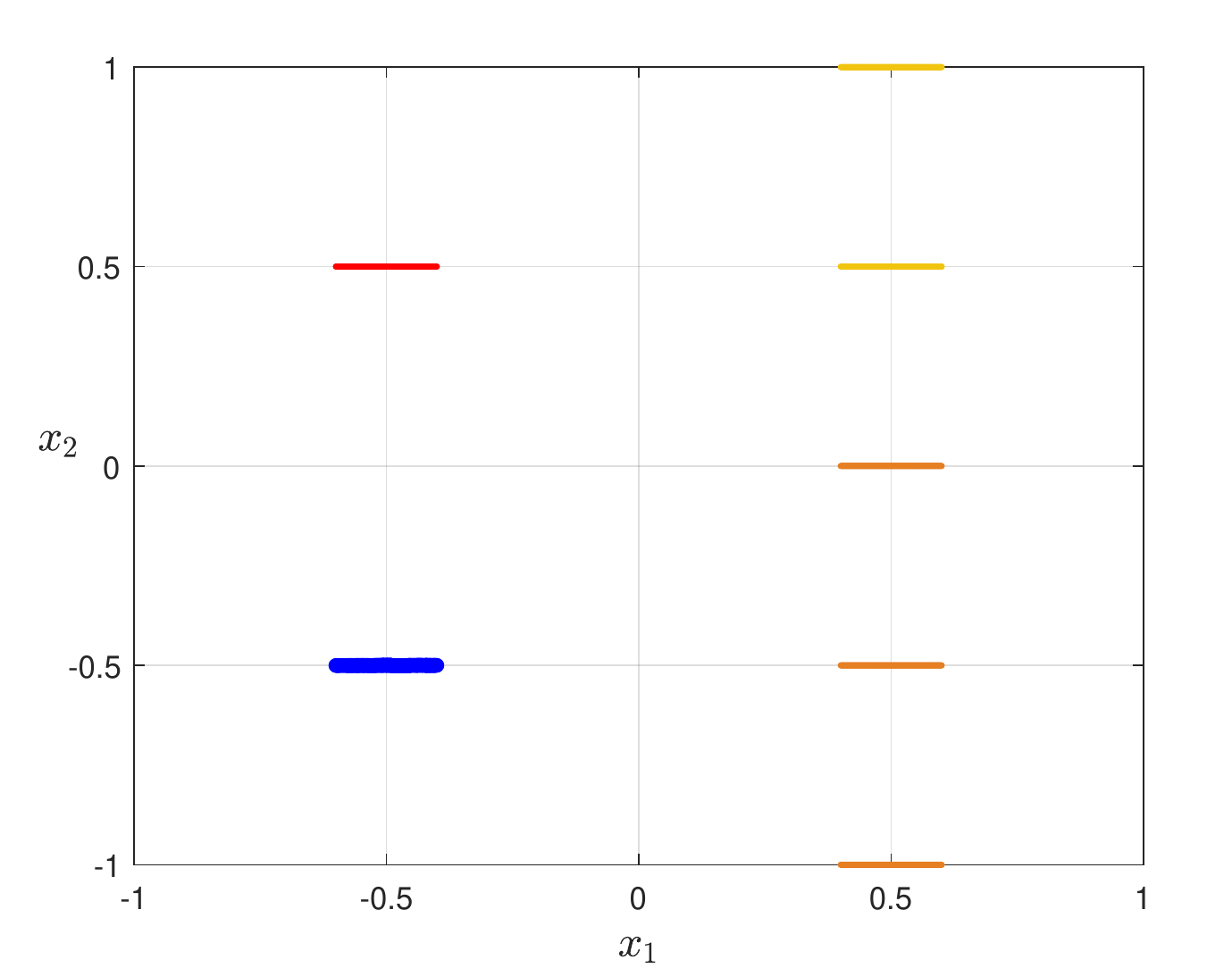}}
\subfigure[DN-NSGAII]{\includegraphics[width=1.1in]{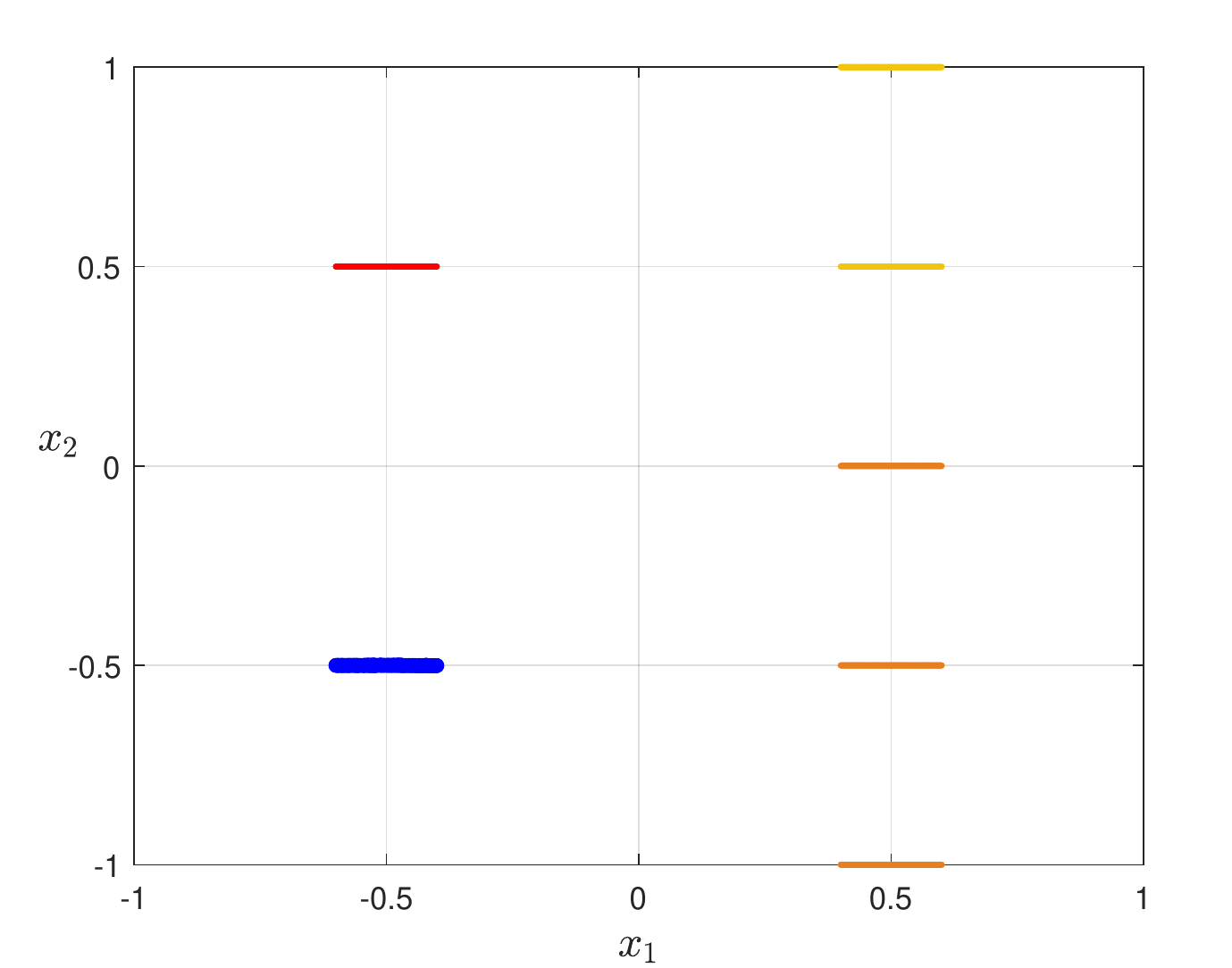}}
\subfigure[MO\_R\_PSO\_SCD]{\includegraphics[width=1.1in]{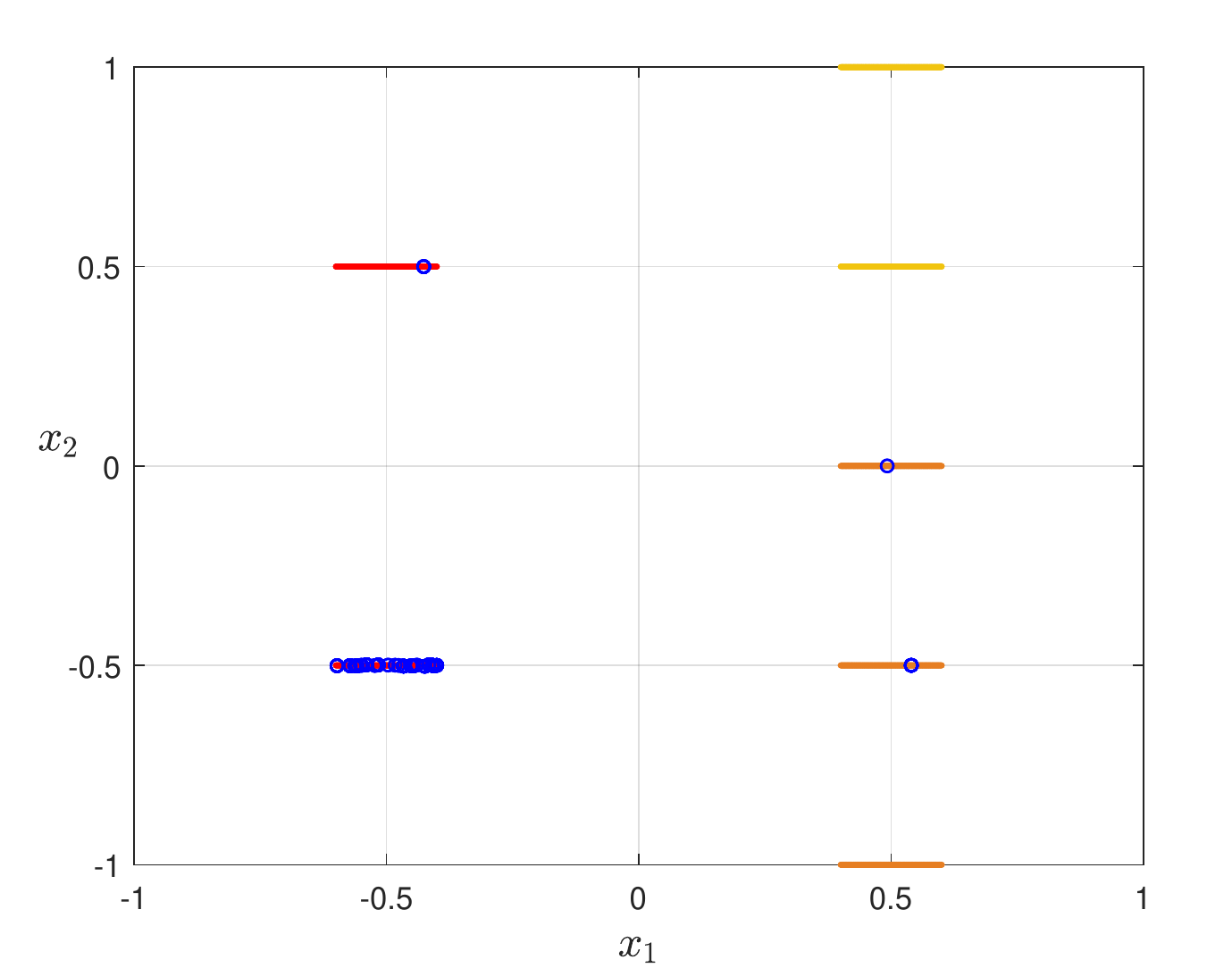}}
\subfigure[MO\_PSO\_MM]{\includegraphics[width=1.1in]{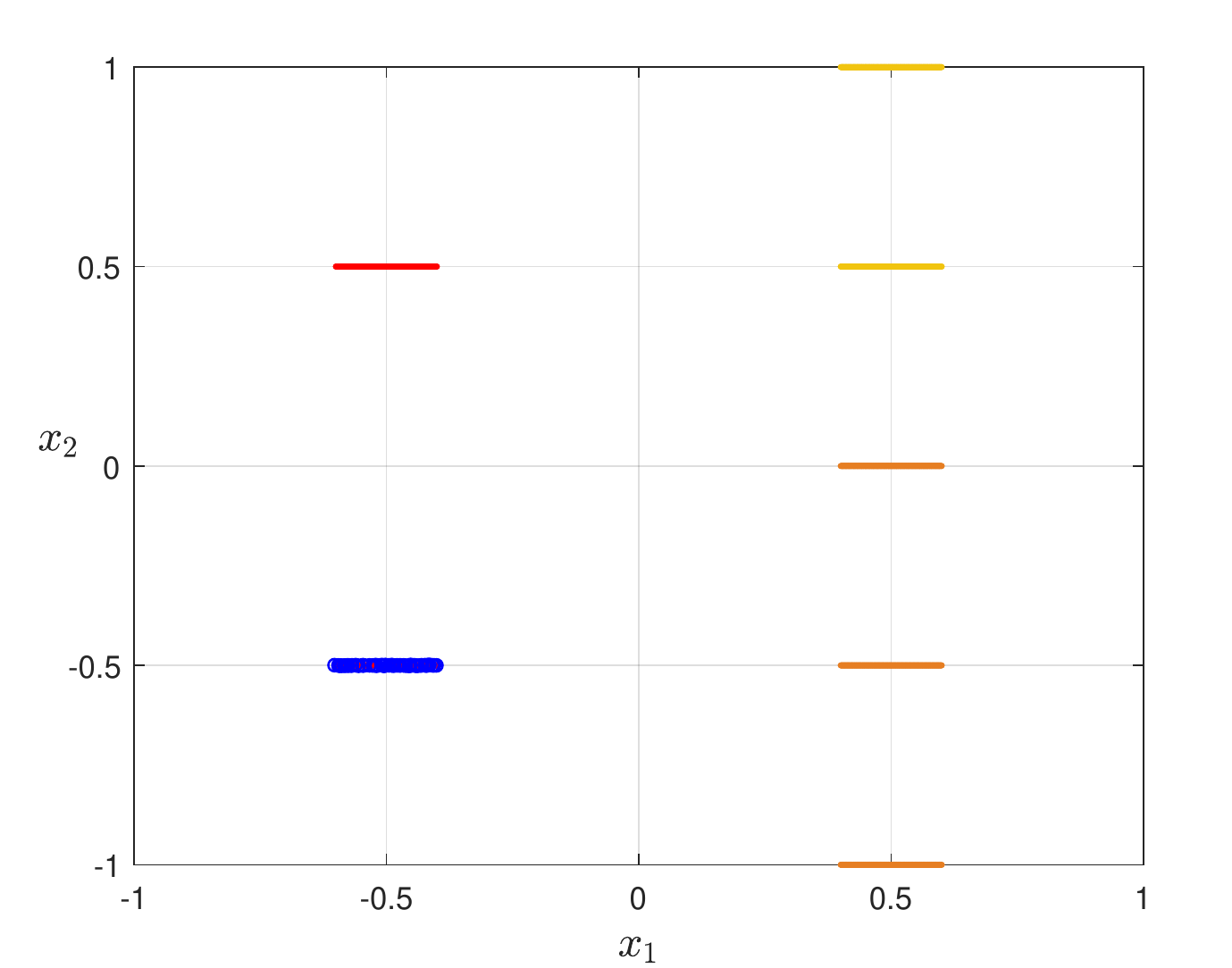}}
\subfigure[DNEA]{\includegraphics[width=1.1in]{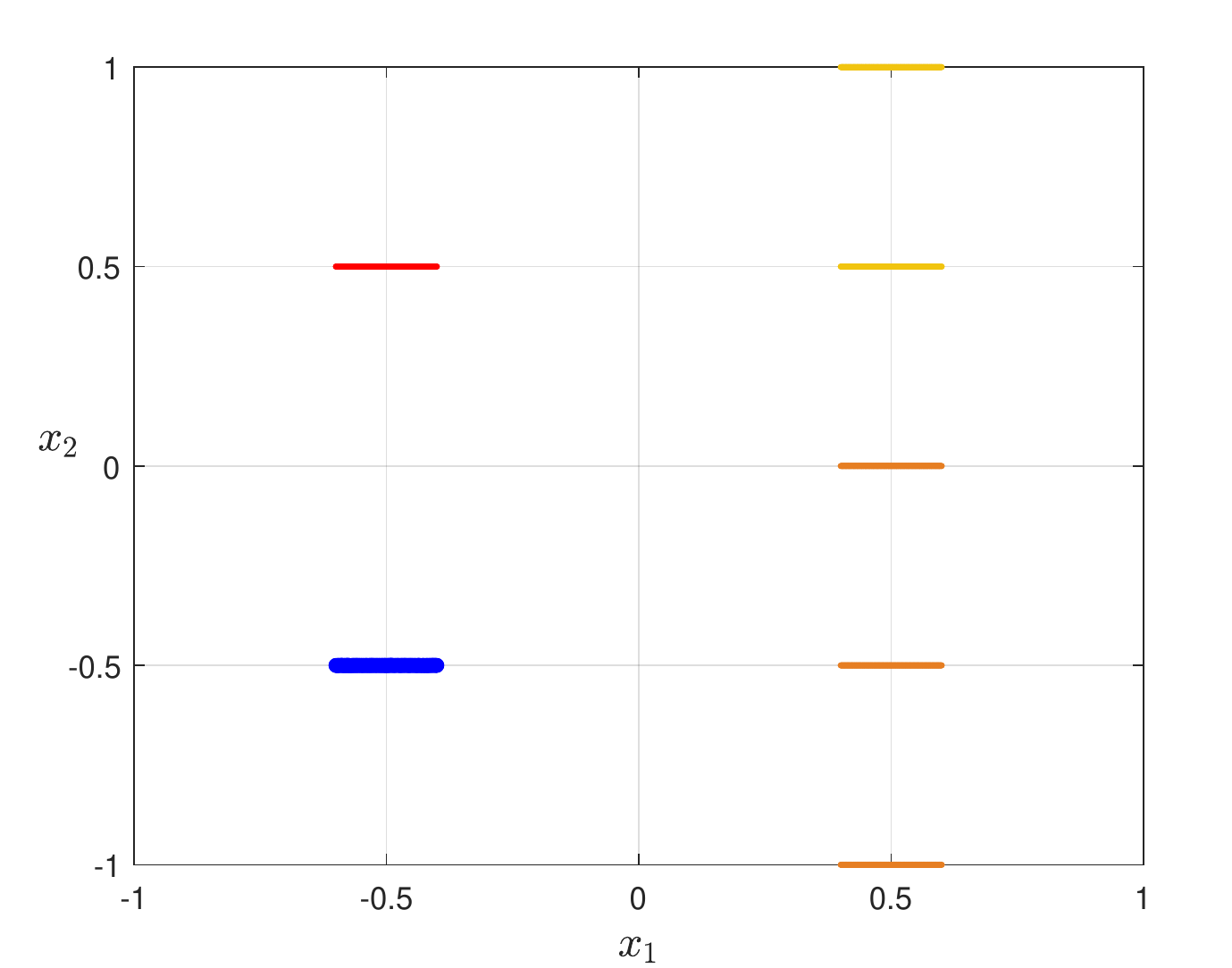}}
\subfigure[Tri-MOEA\&TAR]{\includegraphics[width=1.1in]{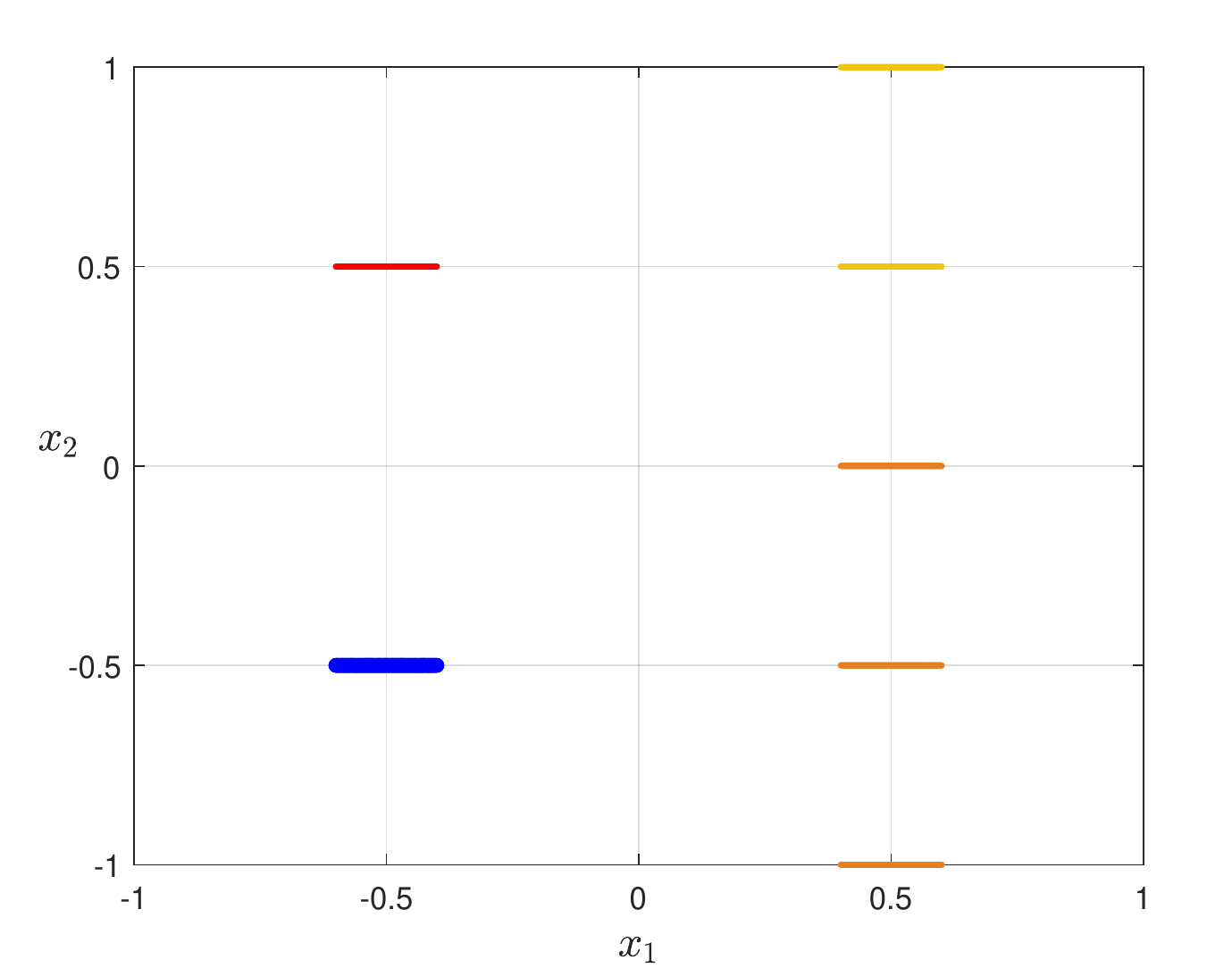}}

\subfigure[DNEA-L]{\includegraphics[width=1.1in]{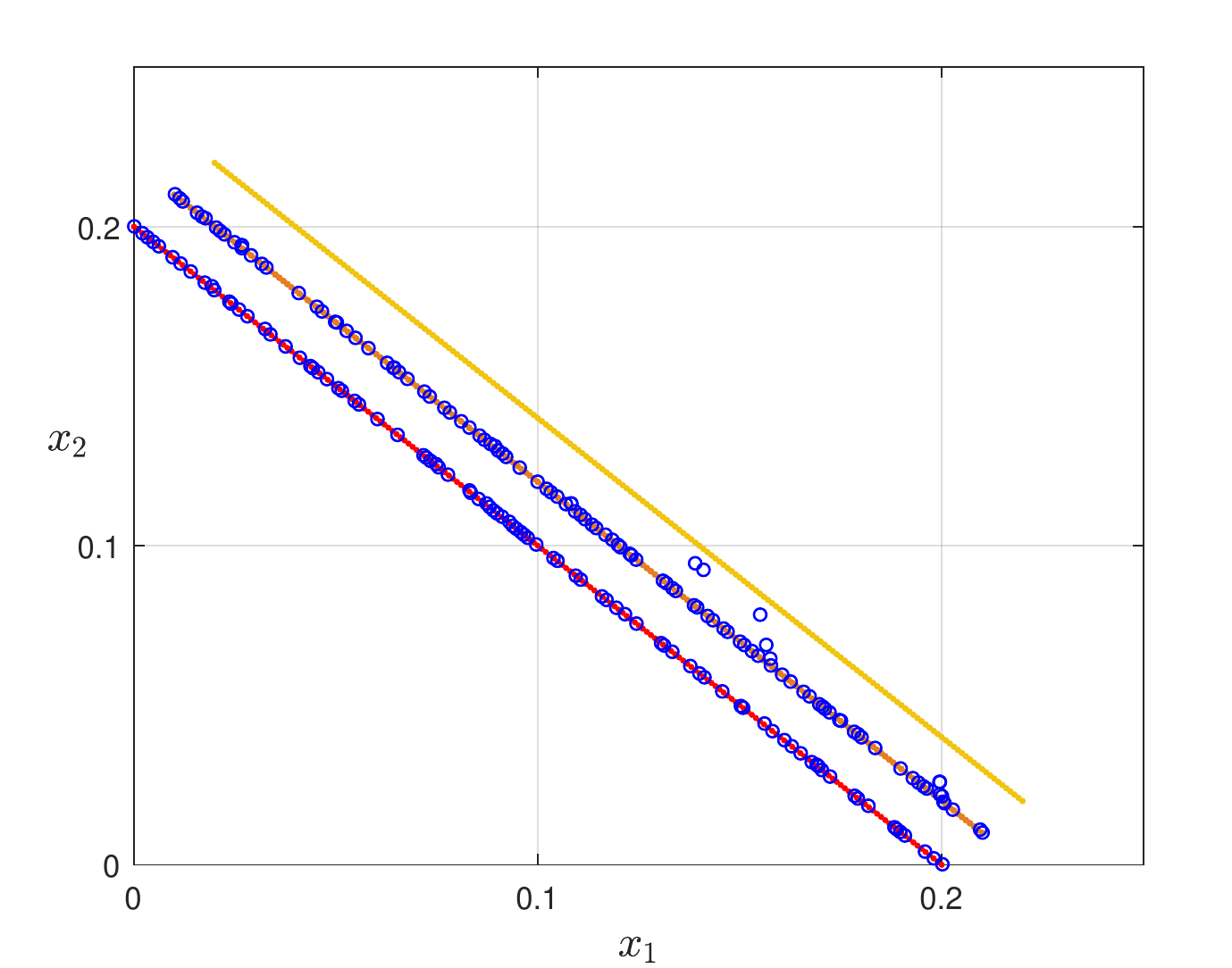}}
\subfigure[CPDEA]{\includegraphics[width=1.1in]{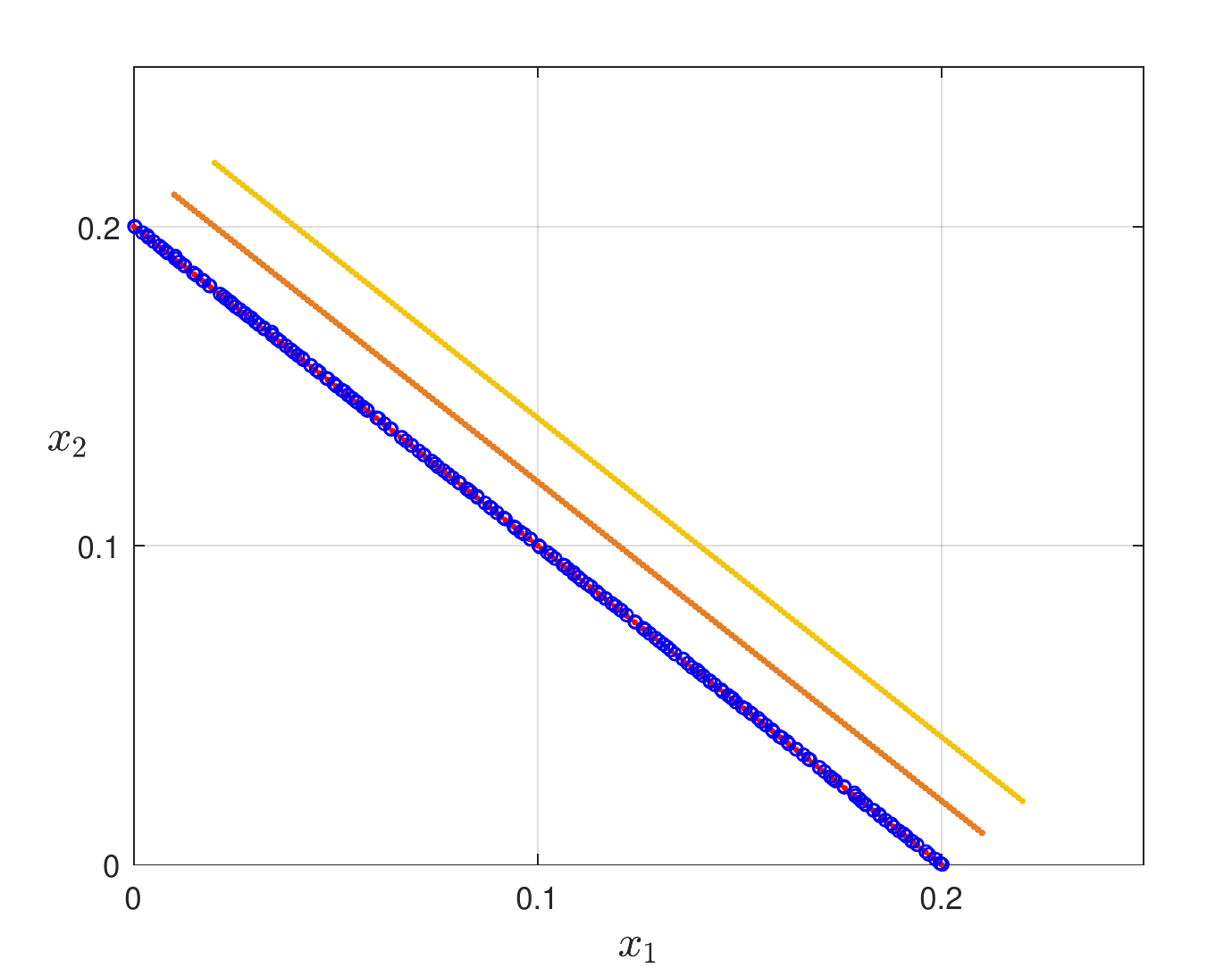}}
\subfigure[MP-MMEA]{\includegraphics[width=1.1in]{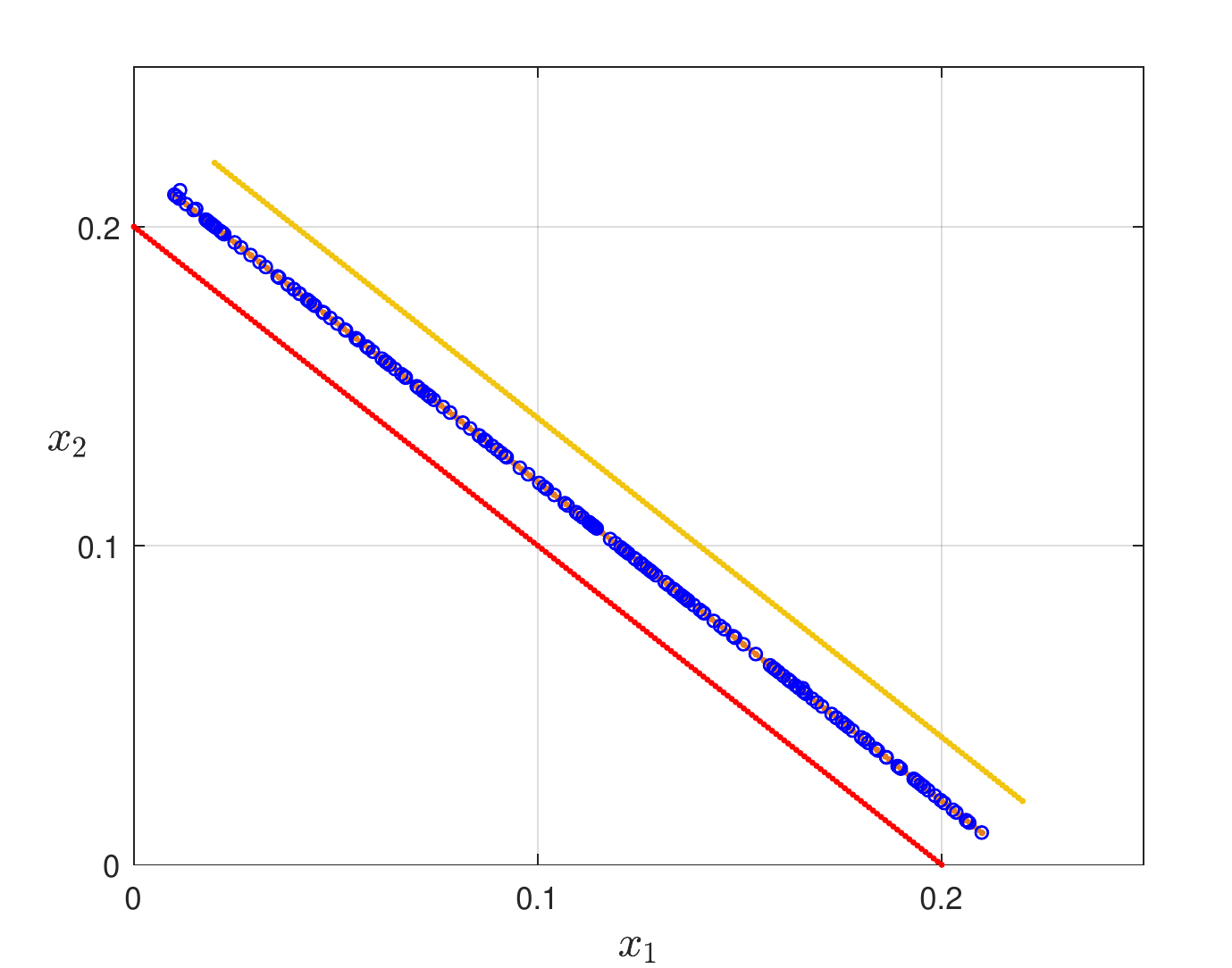}}
\subfigure[MMOEA/DC]{\includegraphics[width=1.1in]{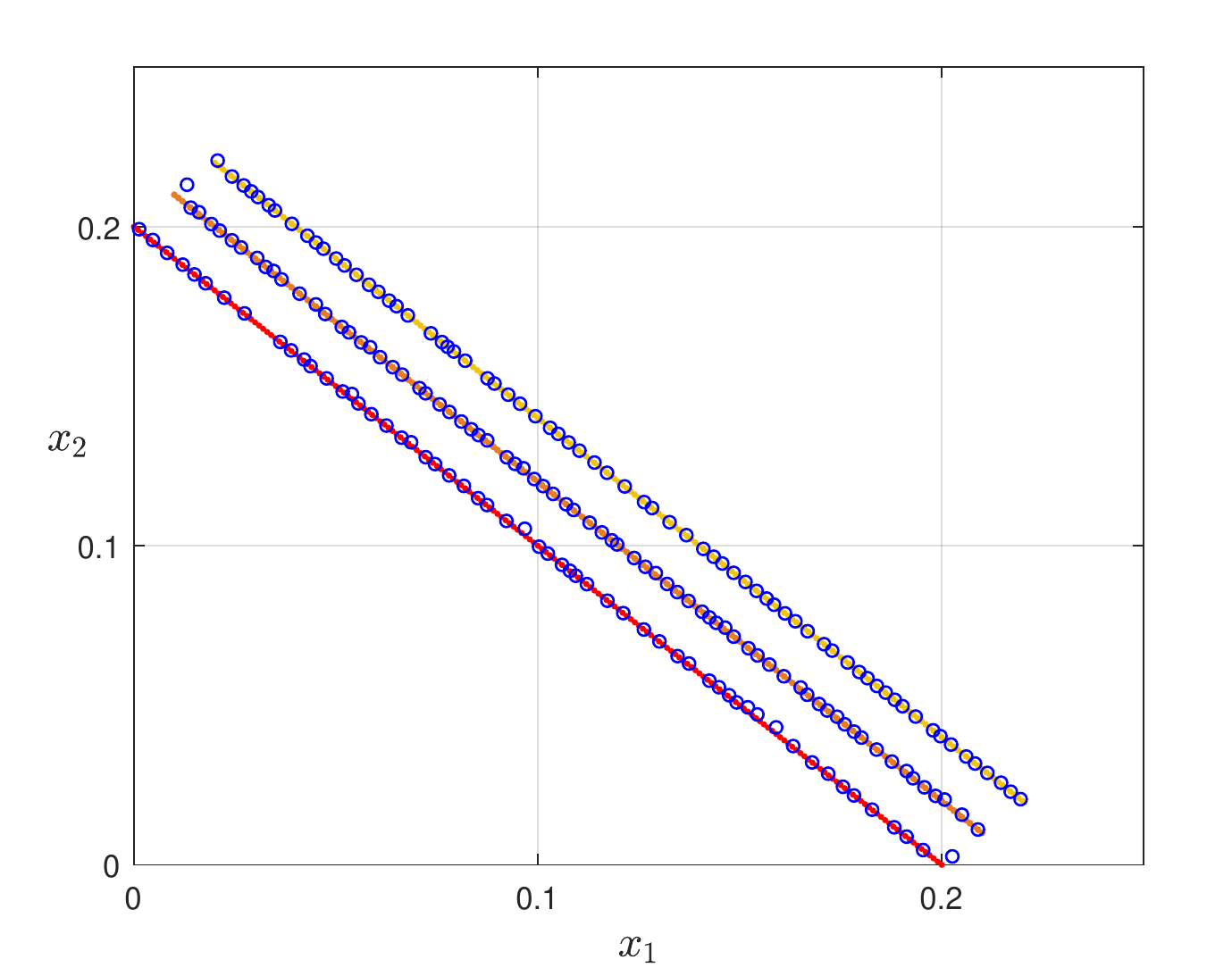}}
\subfigure[MMEA-WI]{\includegraphics[width=1.1in]{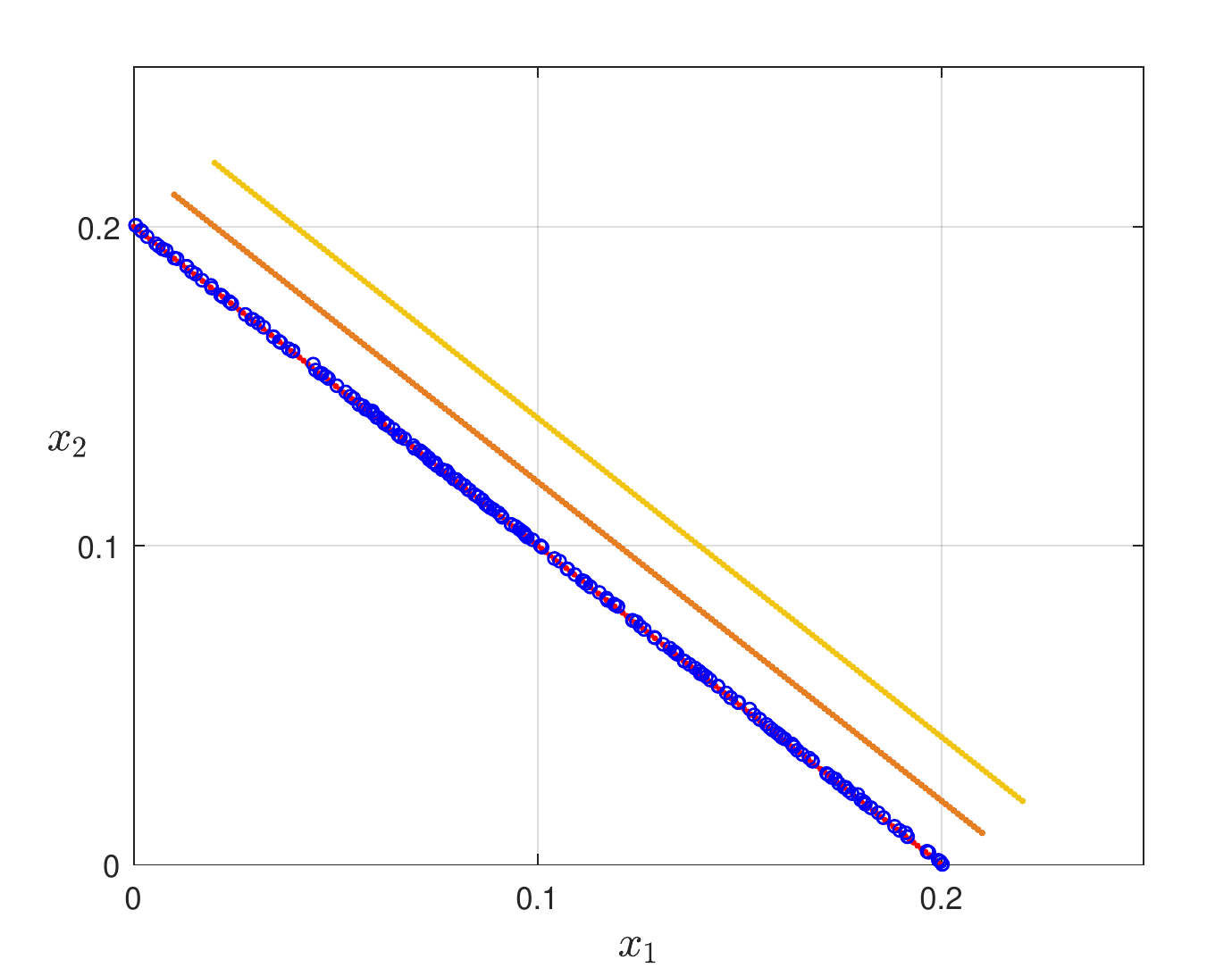}}
\subfigure[HREA]{\includegraphics[width=1.1in]{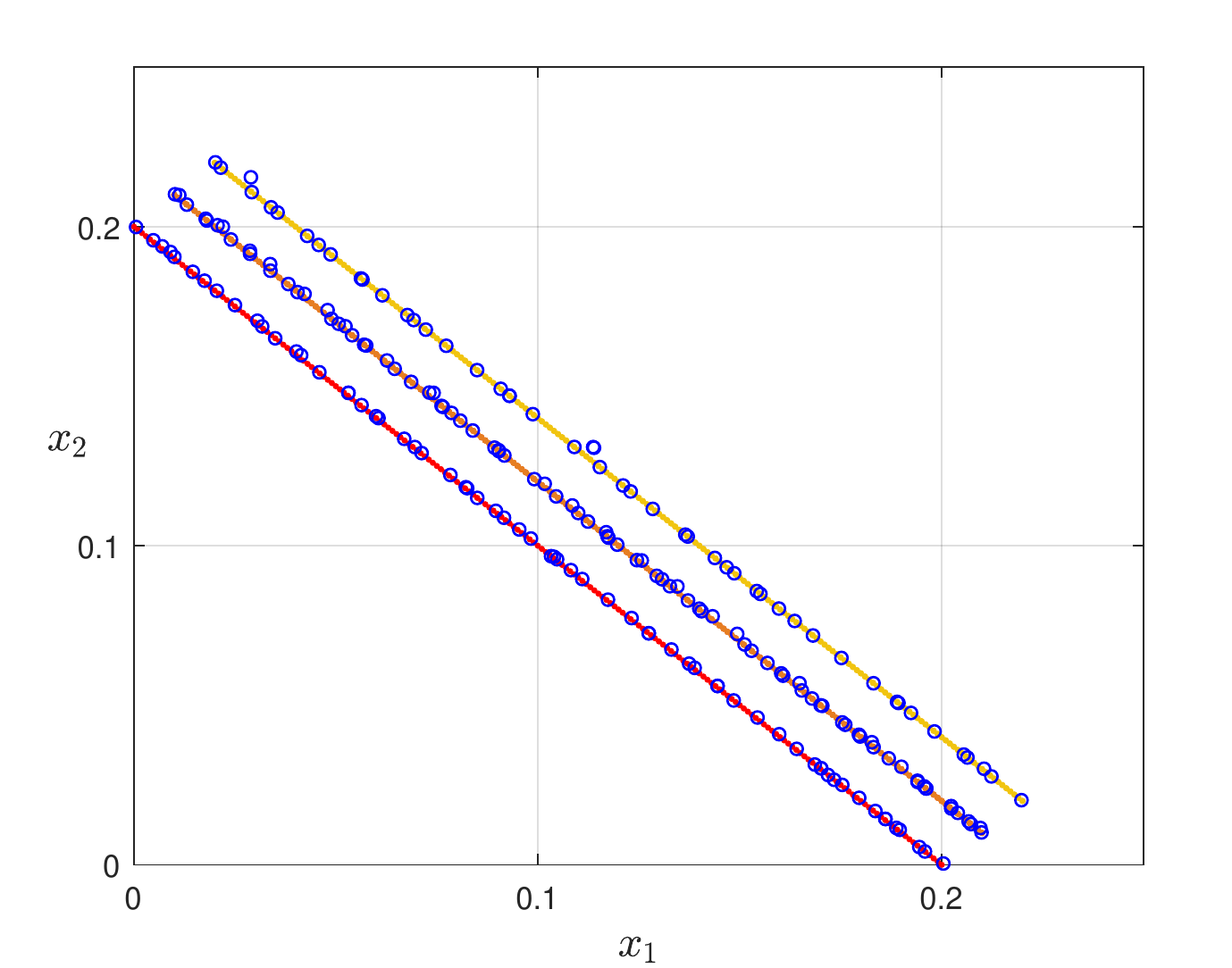}}

\subfigure[DNEA-L]{\includegraphics[width=1.1in]{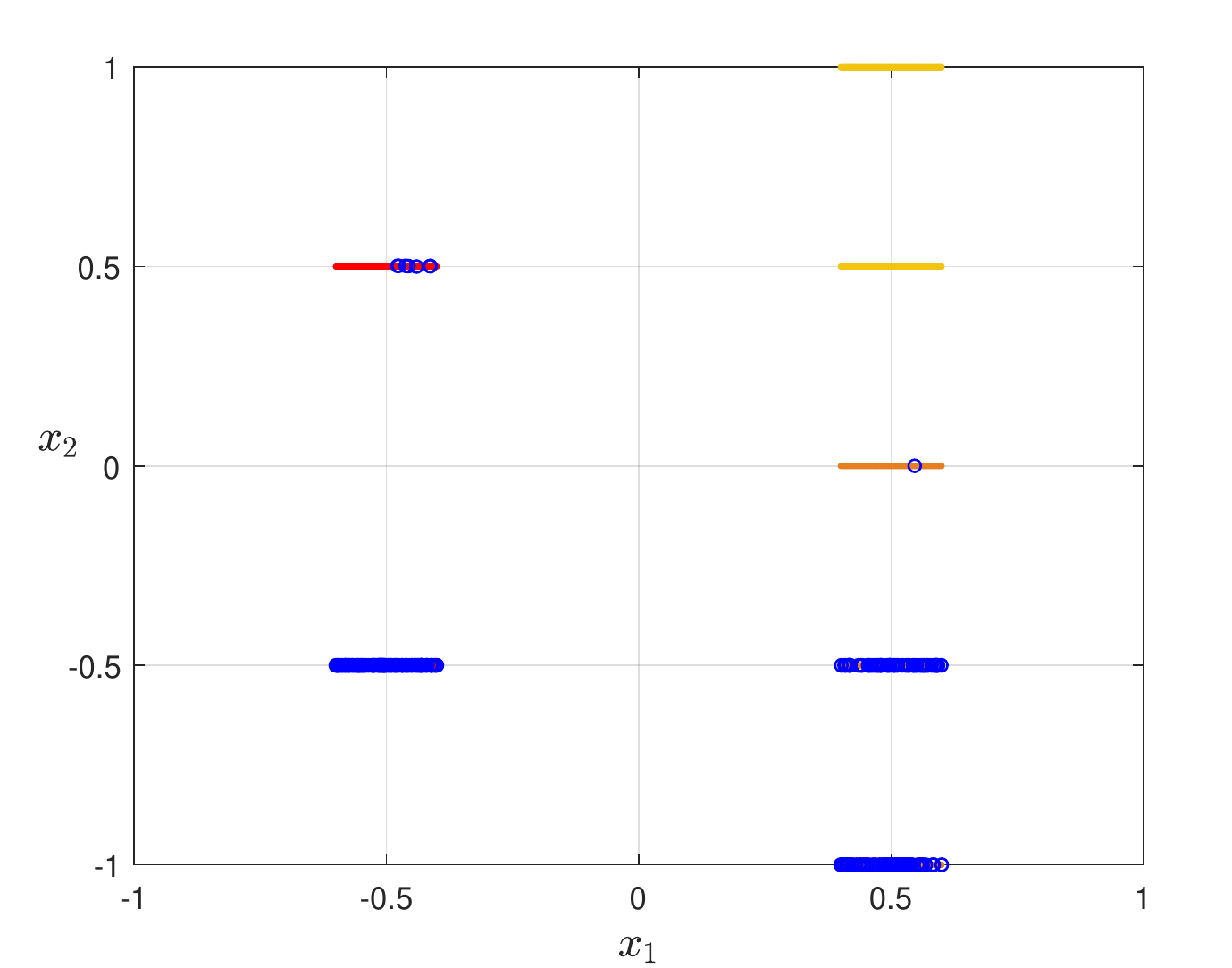}}
\subfigure[CPDEA]{\includegraphics[width=1.1in]{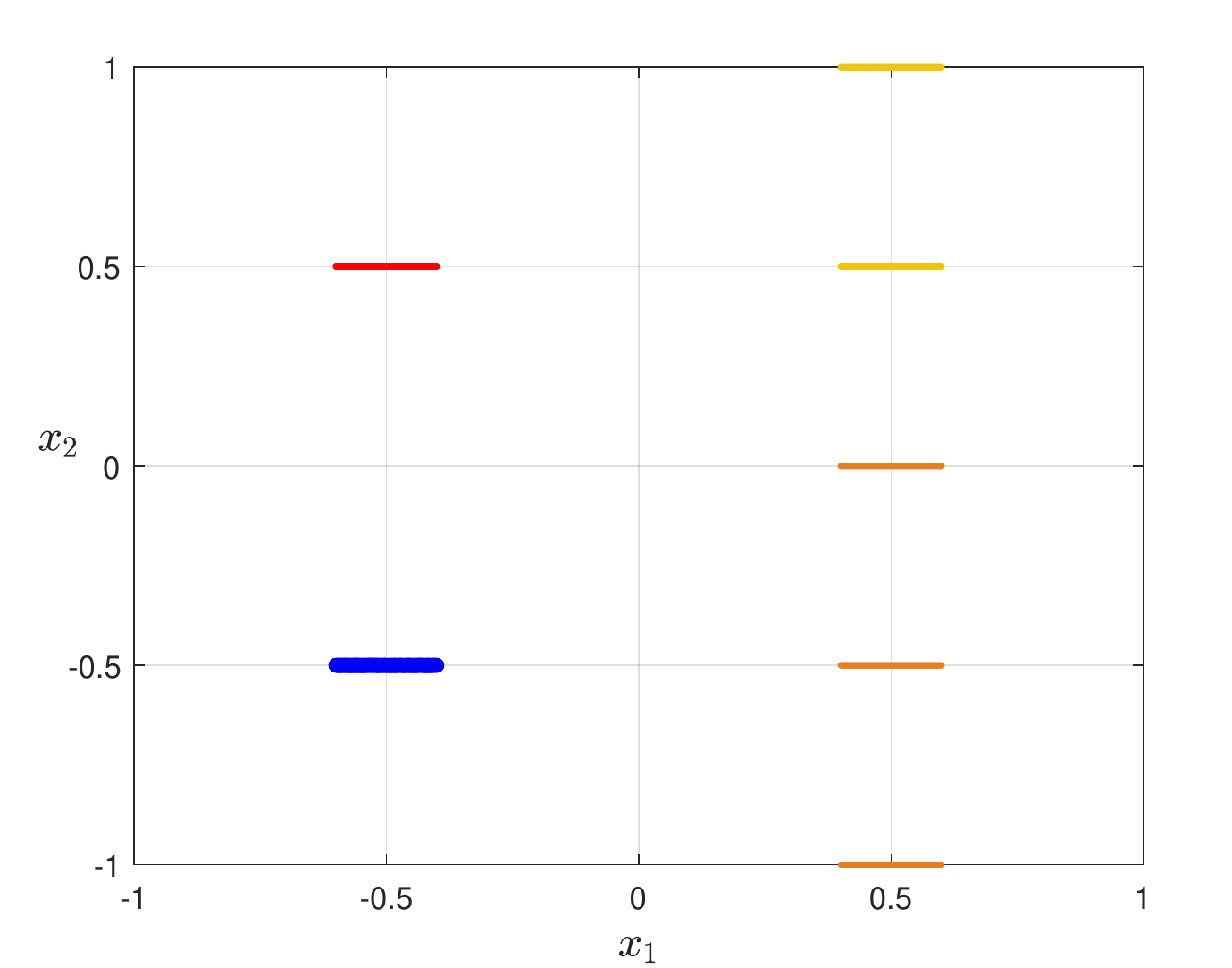}}
\subfigure[MP-MMEA]{\includegraphics[width=1.1in]{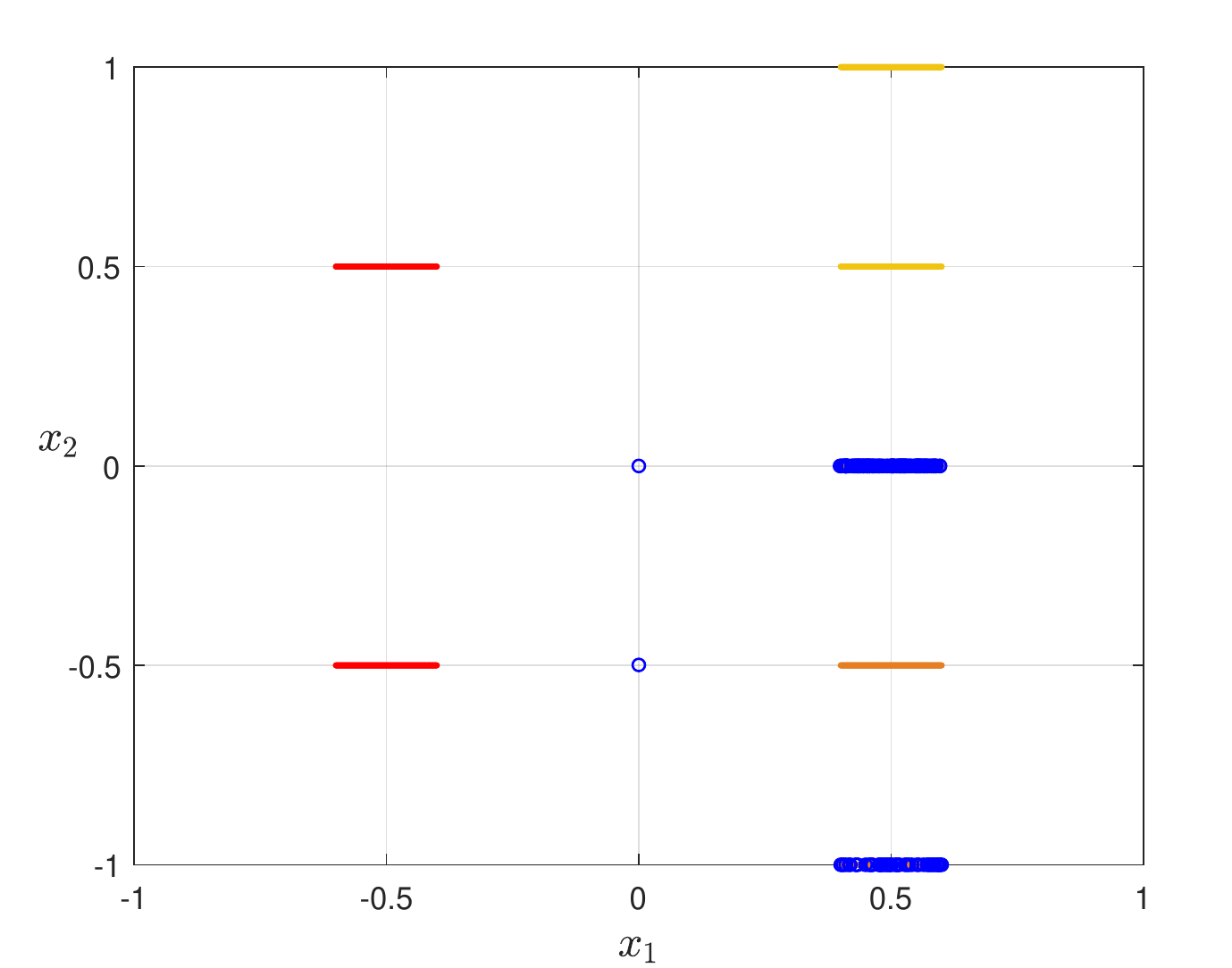}}
\subfigure[MMOEA/DC]{\includegraphics[width=1.1in]{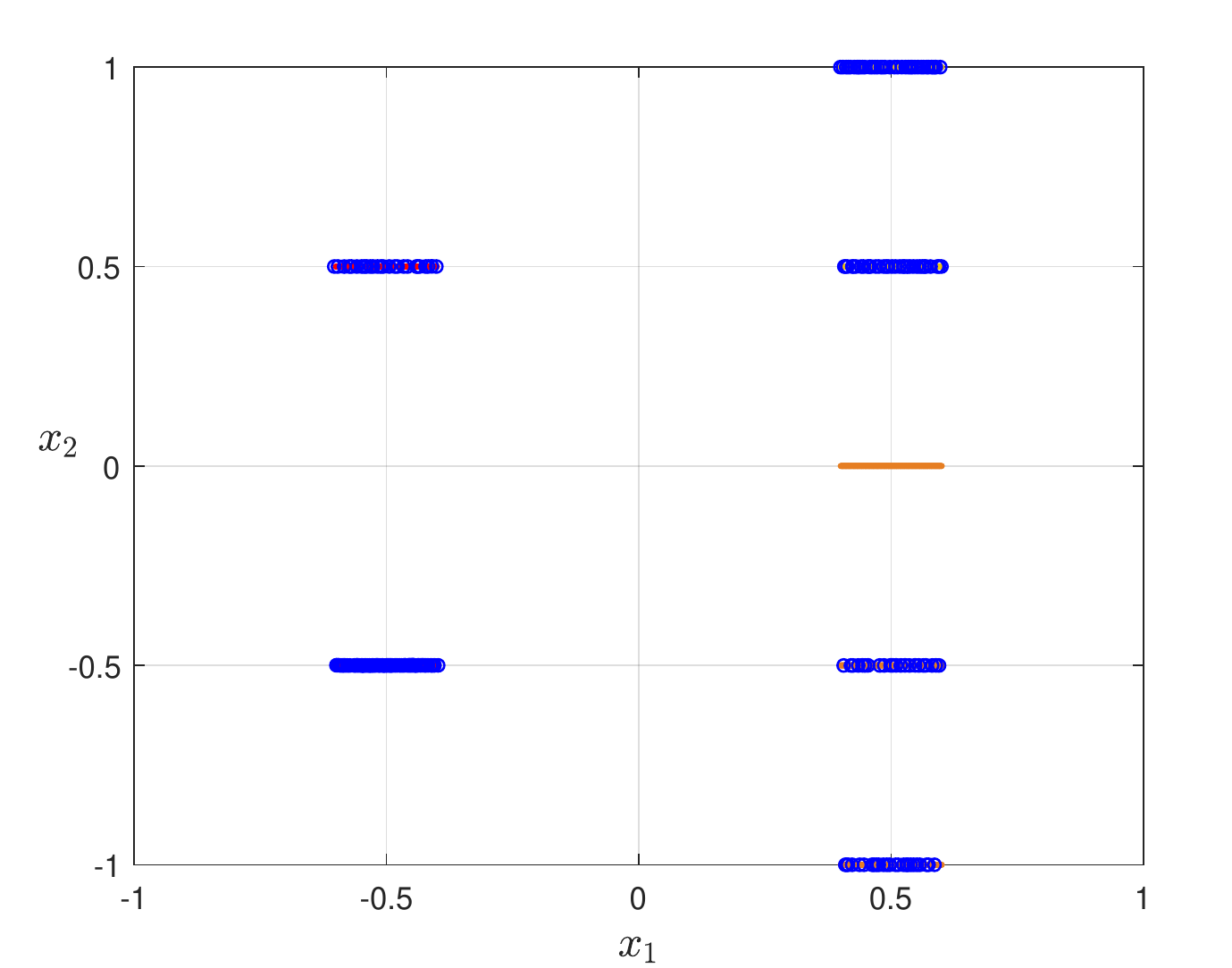}}
\subfigure[MMEA-WI]{\includegraphics[width=1.1in]{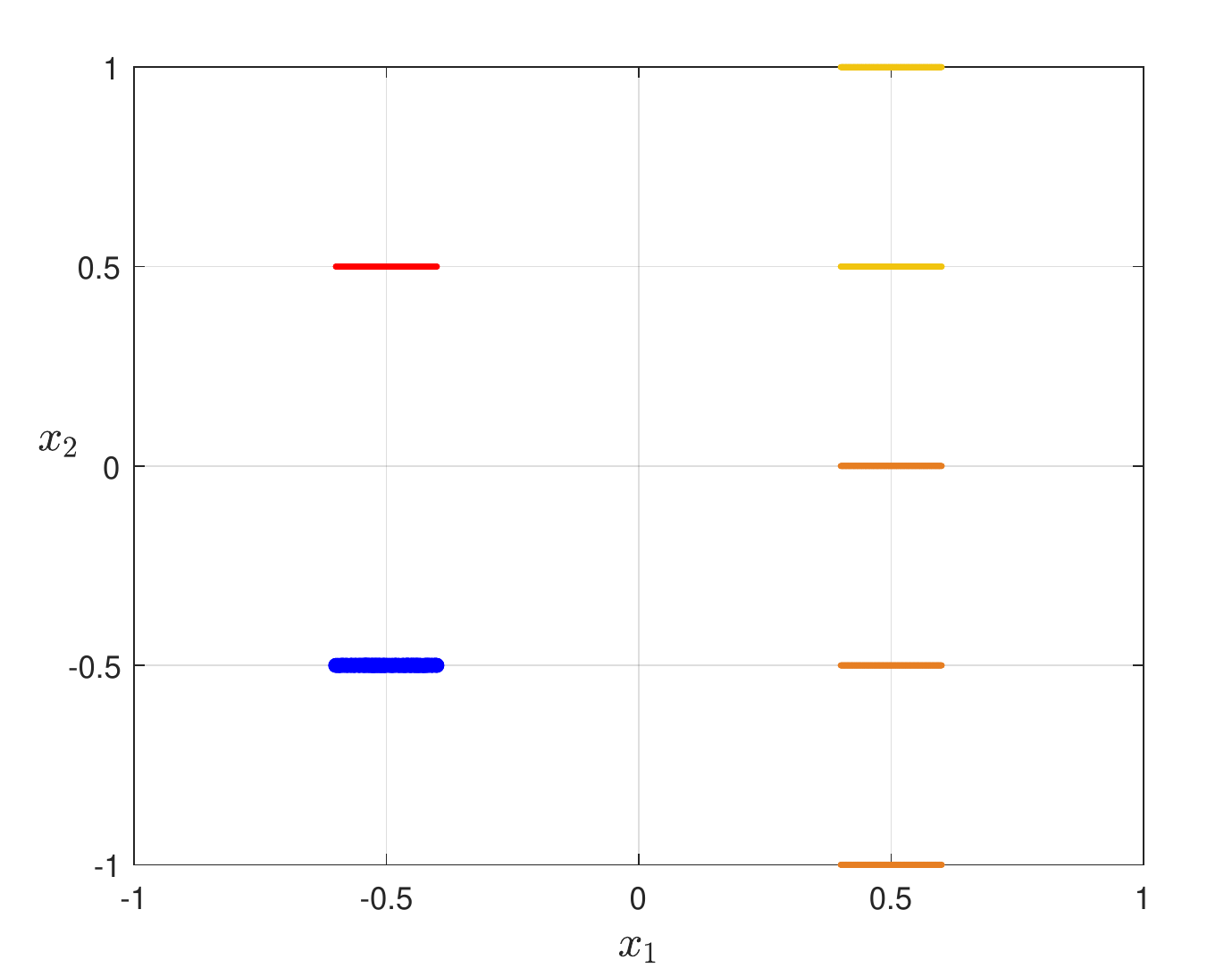}}
\subfigure[HREA]{\includegraphics[width=1.1in]{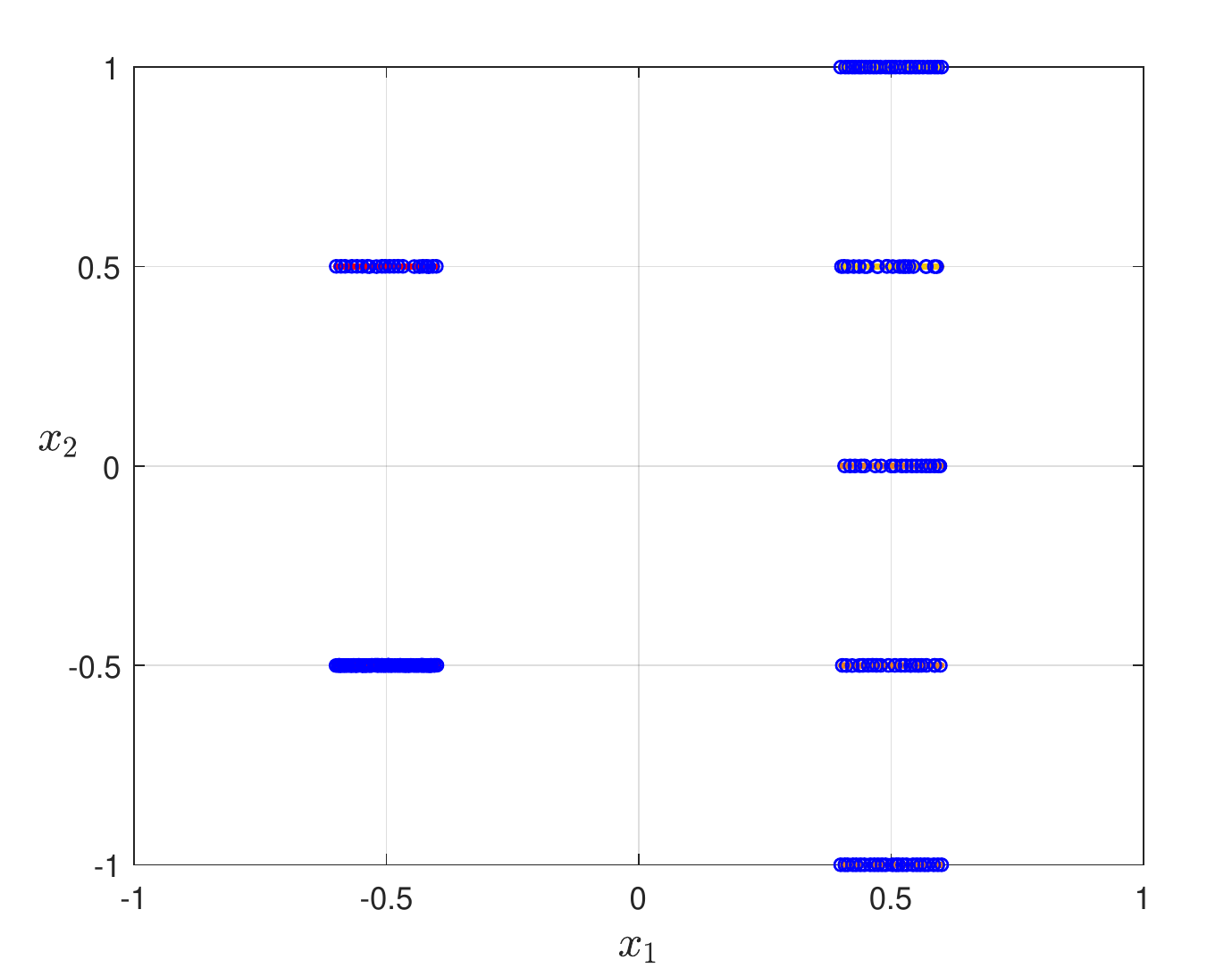}}
\caption{The distribution of solutions obtained by all algorithms (MO\_R\_PSO\_SCD is the short name for MO\_Ring\_PSO\_SCD) in the objective space (the first and third rows) and in the decision space (the second and fourth rows) on IDMPM2T4\_e, where there are two global PSs and 5 local PSs.}
\label{fig_idmperesult}
\end{figure*}

Since only DNEA-L, MMOEA/DC and HREA are designed for MMOPLs, in this part, we mainly focus on their performance. The average rank of all algorithms on MMOPLs is presented in \fref{fig_avgrankidmpe}. As indicated in \fref{fig_avgrankidmpe}, HREA obtains significant best results on the chosen benchmark problems, followed by MMOEA/DC and DNEA-L. The overall average ranks for these three algorithms in terms of IGDX are 1.12, 1.94 and 3.39 respectively. It's worth mentioning that, HREA wins 12 instances for all 13 problems in terms of $IGDX$ and $1/PSP$ as shown in Table S-X and Table S-XI. In addition, \fref{fig_idmperesult} presents the final distribution of solutions in the objective and decision spaces on IDMPM2T4\_e. HREA can stably obtain all two global PSs and 5 local PSs, while MMOEA/DC and DNEA-L can obtain some of the global and local PSs. For DNEA-L, a parameter $K$ (set to 3 in this study) is introduced to get the first $K$-layer PFs. However, for IDMPM2T4\_e, it can only obtain the first two PF layers. For MMOEA/DC, the double clustering method is proposed based on DBSCAN \cite{schubert2017dbscan} to form evenly distributed solutions in each PS area. Experimental results show this method performs well on many MMOPs. The main problem is that the clustering could be inaccurate and cause unstable convergence. For HREA, the local convergence quality is effective in finding and obtaining local PSs. A parameter $\epsilon$ is introduced to control the quality of the obtained local PSs. For DMs who are not familiar with MMOP, the setting of this parameter could be an obstacle, e.g., the authors suggested setting $\epsilon=0.3$ for most problems but $\epsilon=0.3$ should be set to 0.5 for IDMPM2T4\_e to obtain all PSs. Since IDMP\_e is derived from IDMP, it's still hard for primitive MMEAs to find all global PSs.

\subsection{Performance comparison on Multi-polygon problems}
\label{sec_polygonresult}
To study the performance of MMEAs on problems with many objectives and many decision variables, Multi-polygon problems are chosen as the benchmark, where $M$ is set to 3, 4, 8, 10 and $D$ is set to 2, 4, 10, 20 ,50 respectively.

\begin{figure}[tbph]
	\begin{center}
		\includegraphics[width=3.5in]{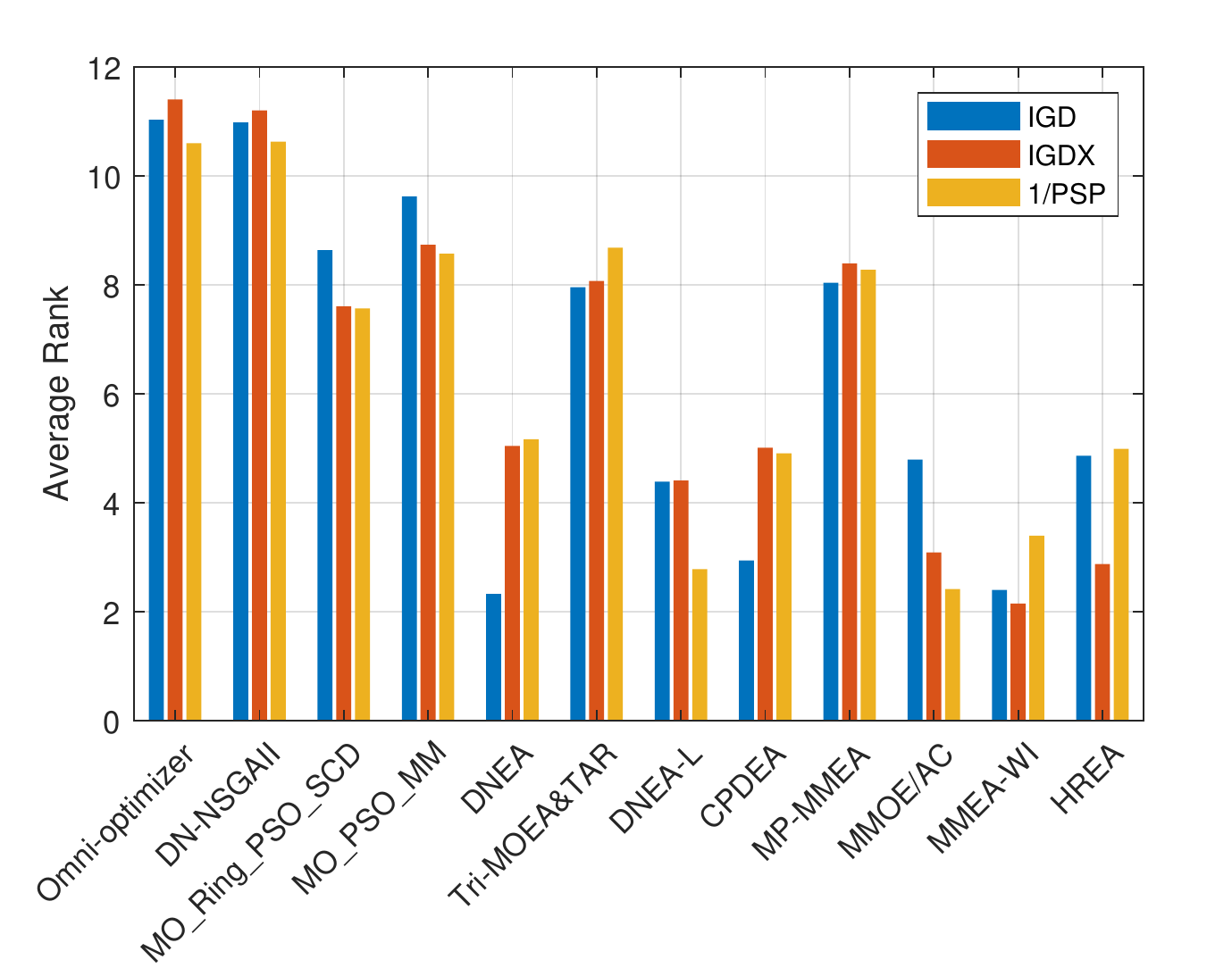}
		\caption{The average rank of all compared MMEAs on Multi-polygon test problems in terms of $IGD$, $IGDX$ and $1/PSP$.}
		\label{fig_polygon}
	\end{center}
\end{figure}

\fref{fig_polygon} shows the overall performance ranking of all compared MMEAs, which indicates that MMEA-WI, MMOEA/DC, DNEA-L and HREA are competitive. No algorithm shows overwhelming better performance on Multi-polygon problems. In addition, different from other MMOP test suites, the average ranks in terms of $IGDX$ and $1/PSP$ vary for different algorithms. As shown in Table S-XIV, there are many $Inf$ values, which is the reason for the above issue. As we can see from \eref{equ_psp}, when the maximum value of the $i$-th decision variable in the obtained solutions is less than the minimum value of the $i$-th decision variable in the PS, then $PSP=0$ and $1/PSP=Inf$, which means that it fails in obtaining the true PS. From Table S-XIV, only DNEA, DNEA-L, CPDEA, MMOEA/DC, MMEA-WI and HREA can find the true PS when $D$ is larger than 4. As for $IGDX$, MMEA-WI, DNEA-L, HREA and MMOEA/DC win 9, 5, 4 and 2 instances respectively. To be specific, HREA wins all instances when $D=2$ and MMEA-WI wins all instances when $D=20, 50$. HREA shows the best performance on low-dimension problems but very poor ability in dealing with high-dimension problems. The use of local convergence quality becomes an obstacle in converging to the true PSs.

\begin{figure*}[htbp]
\centering
\subfigure[Omni-optimizer]{\includegraphics[width=1.1in]{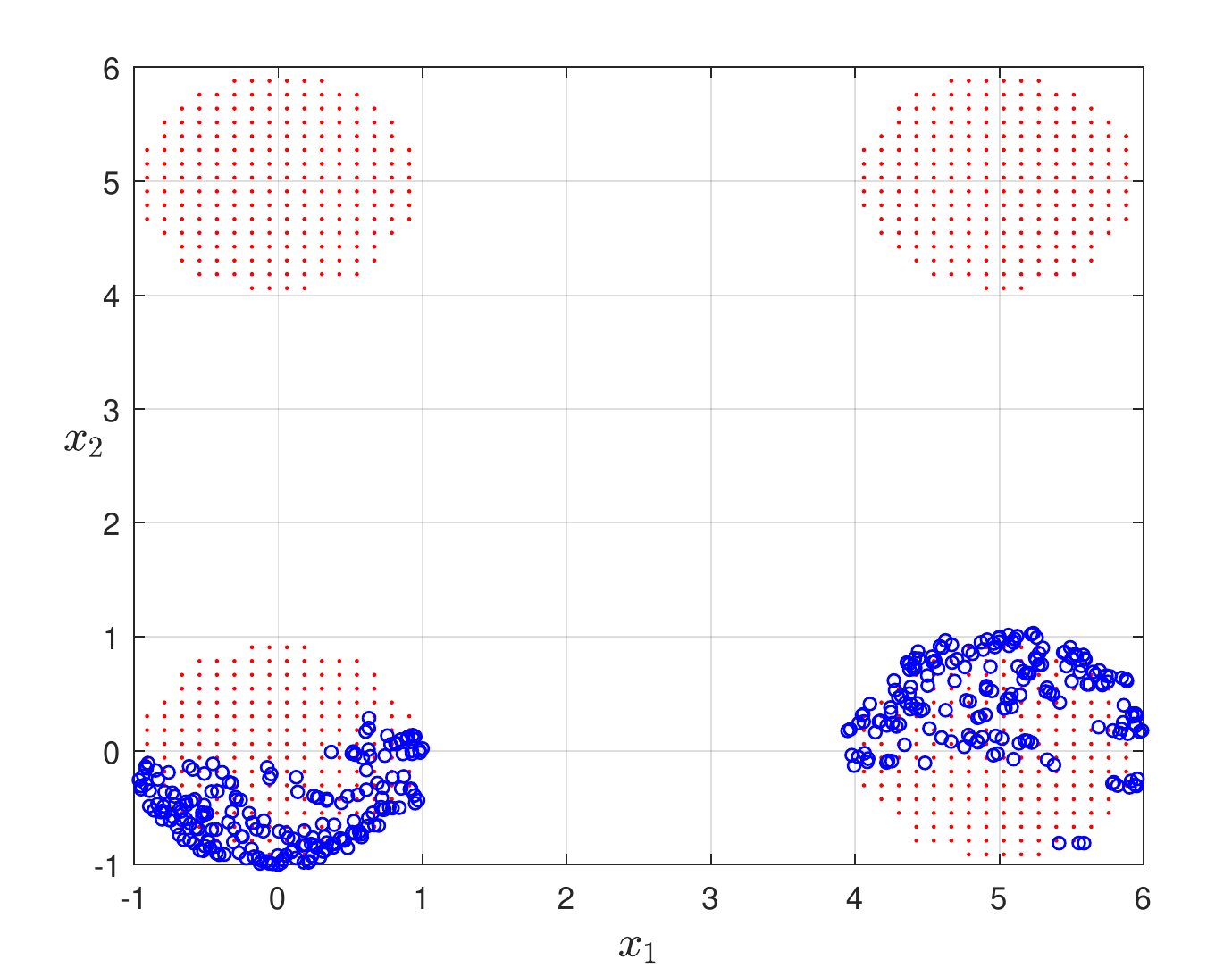}}
\subfigure[DN-NSGAII]{\includegraphics[width=1.1in]{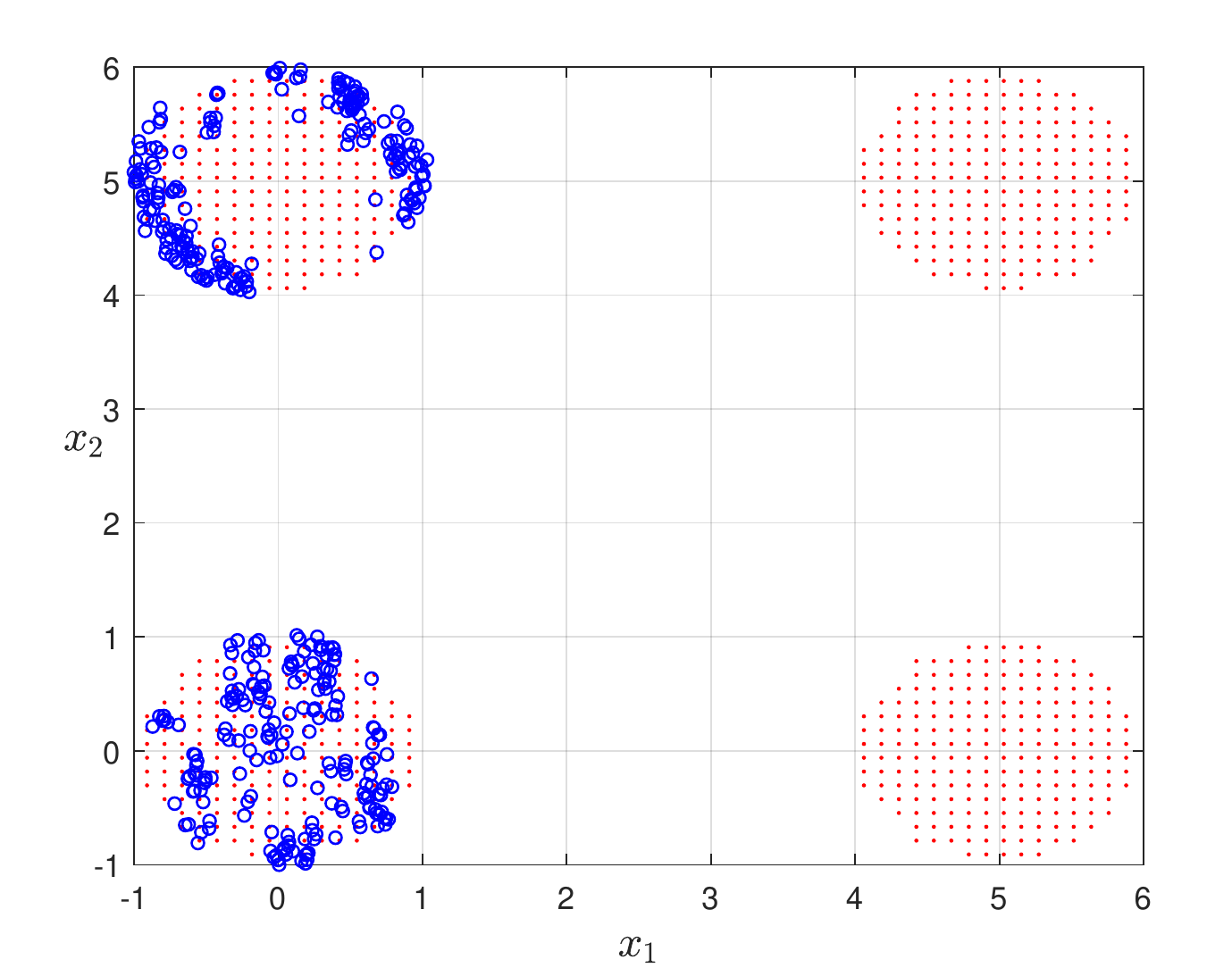}}
\subfigure[MO\_R\_PSO\_SCD]{\includegraphics[width=1.1in]{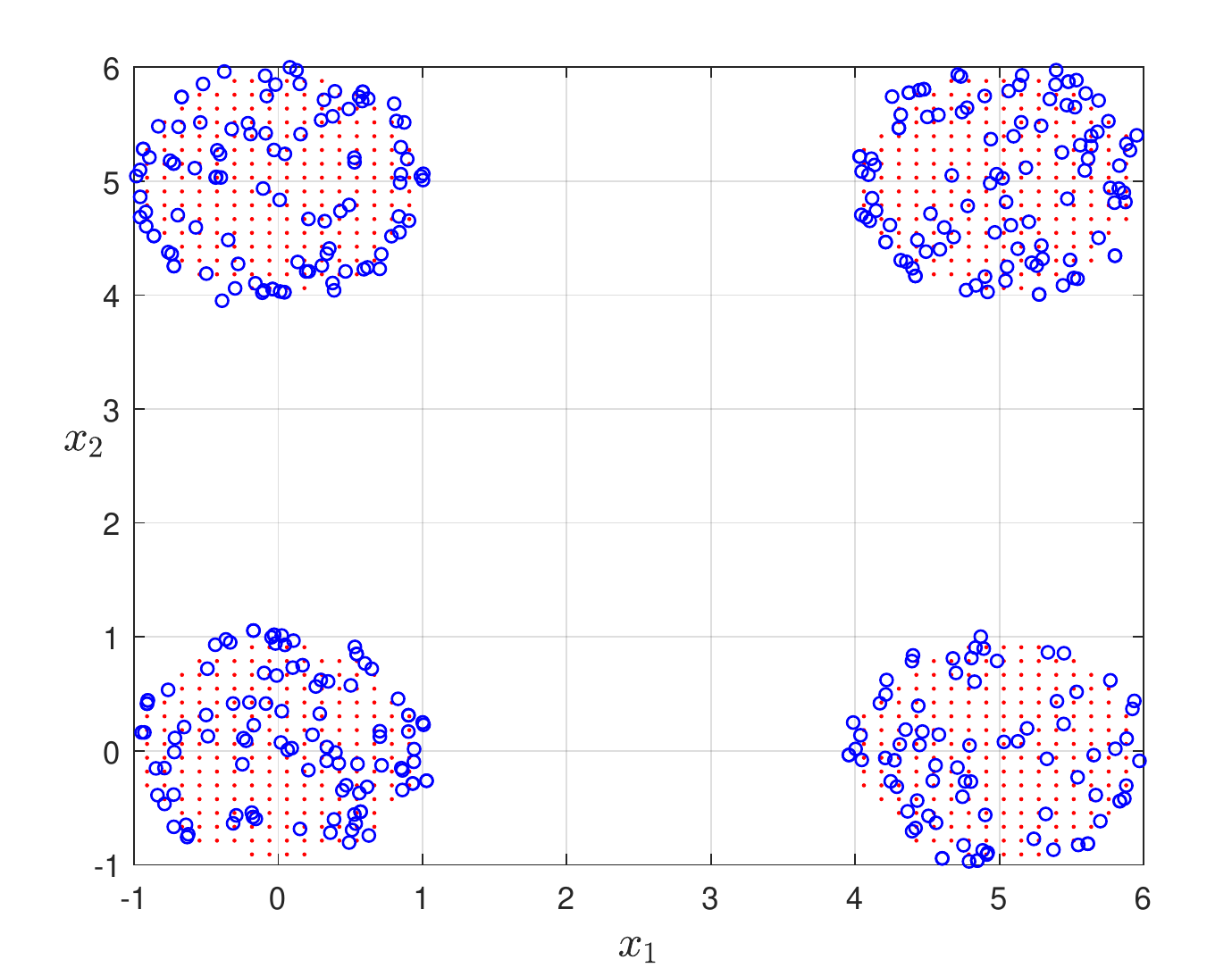}}
\subfigure[MO\_PSO\_MM]{\includegraphics[width=1.1in]{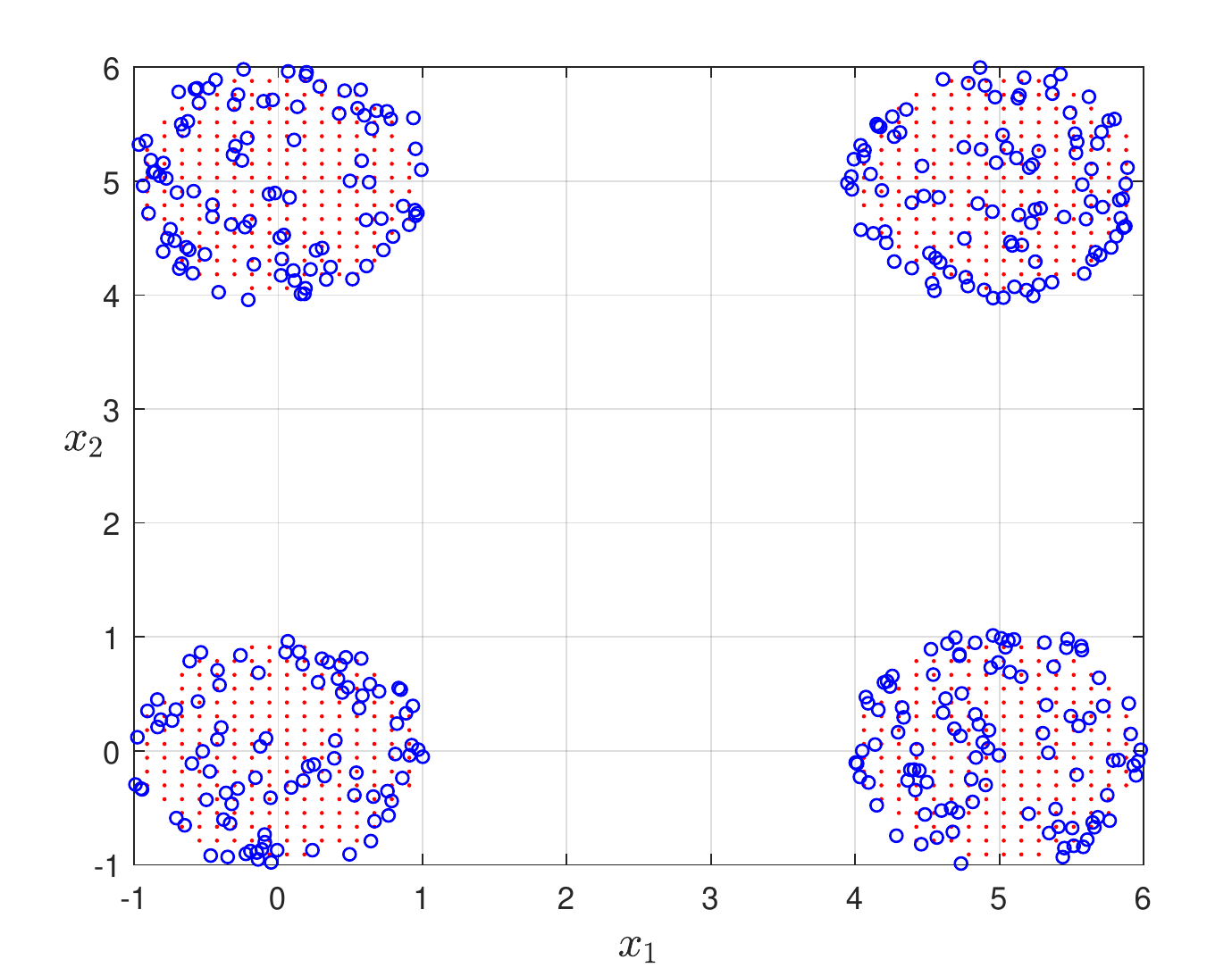}}
\subfigure[DNEA]{\includegraphics[width=1.1in]{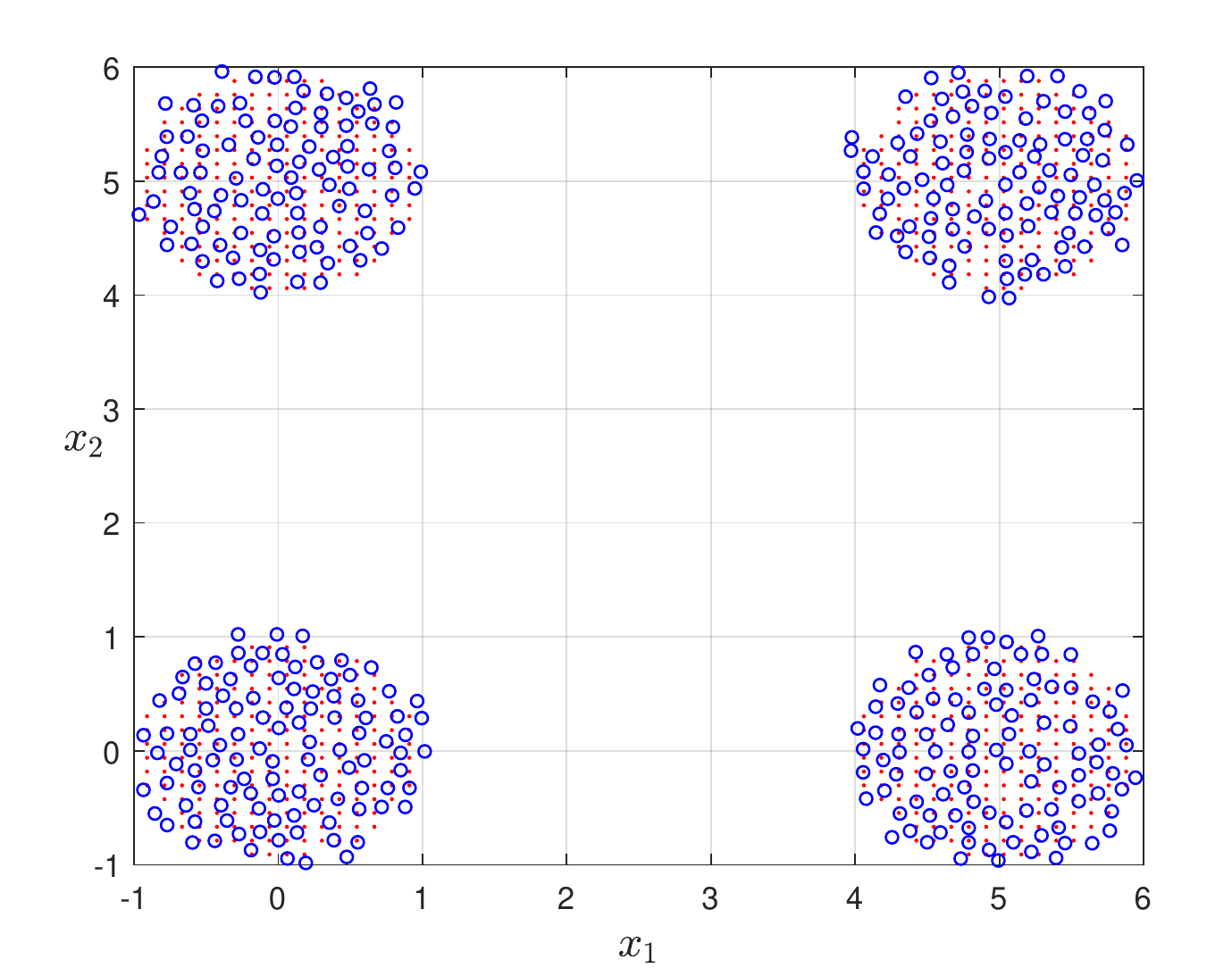}}
\subfigure[Tri-MOEA\&TAR]{\includegraphics[width=1.1in]{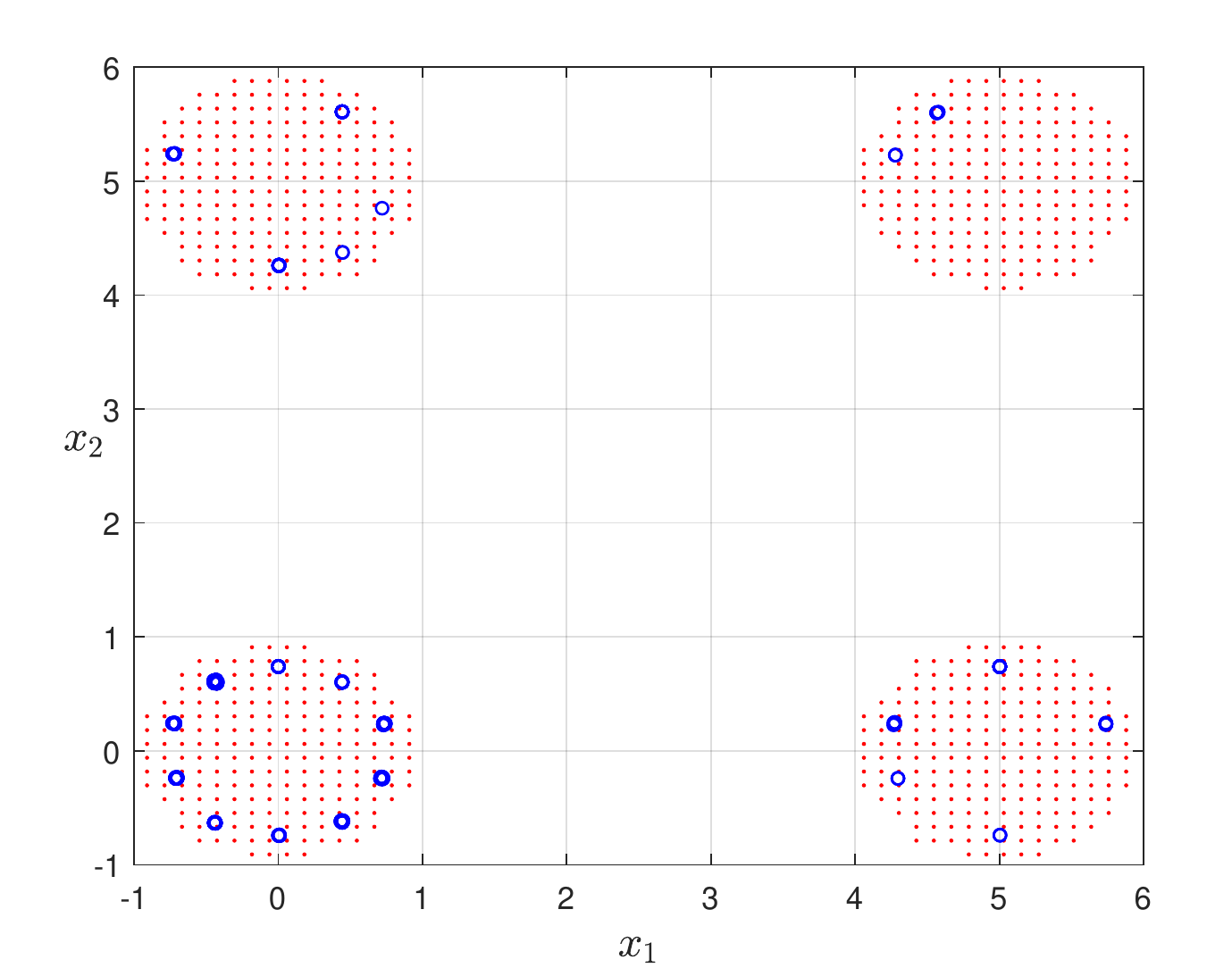}}
\subfigure[DNEA-L]{\includegraphics[width=1.1in]{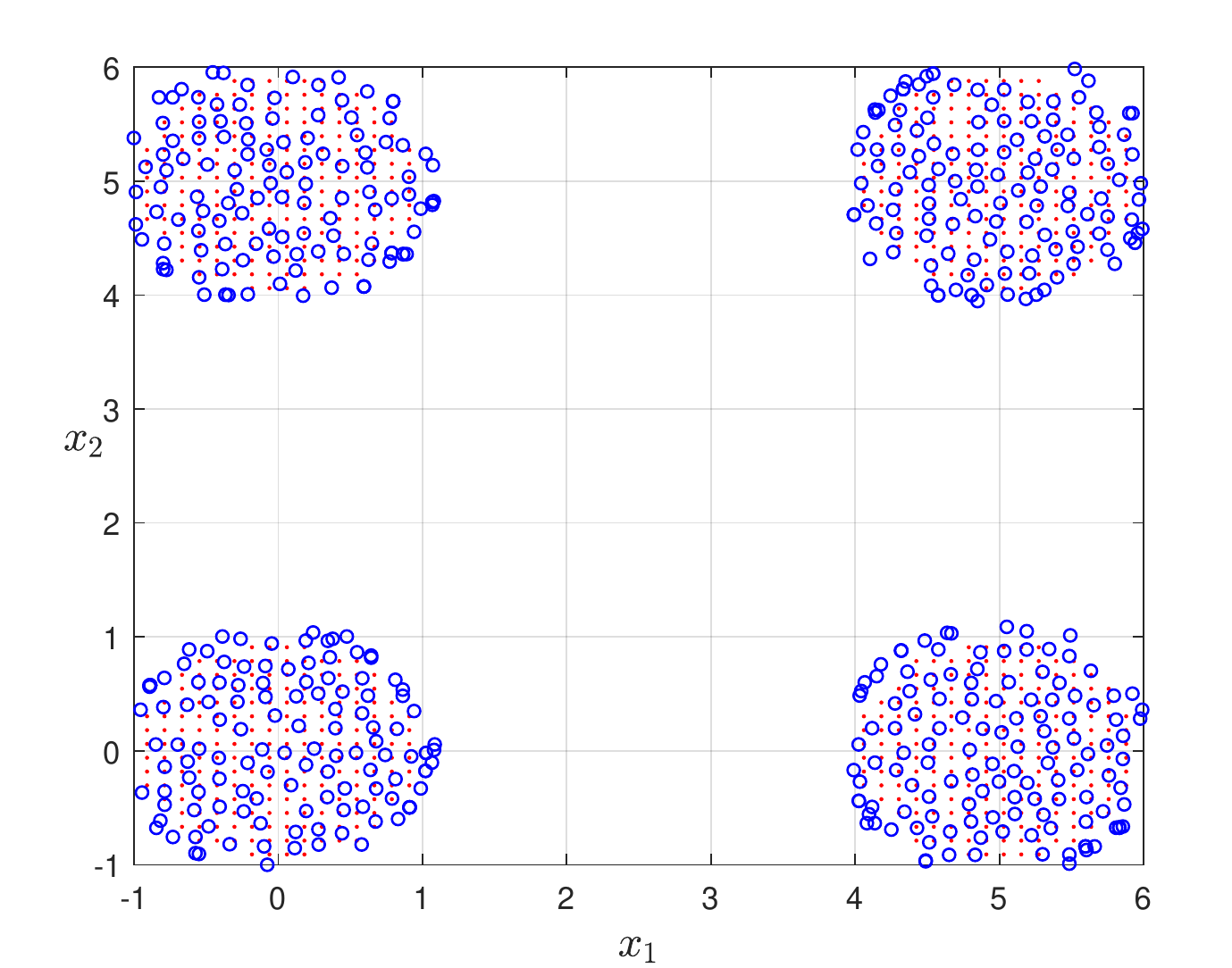}}
\subfigure[CPDEA]{\includegraphics[width=1.1in]{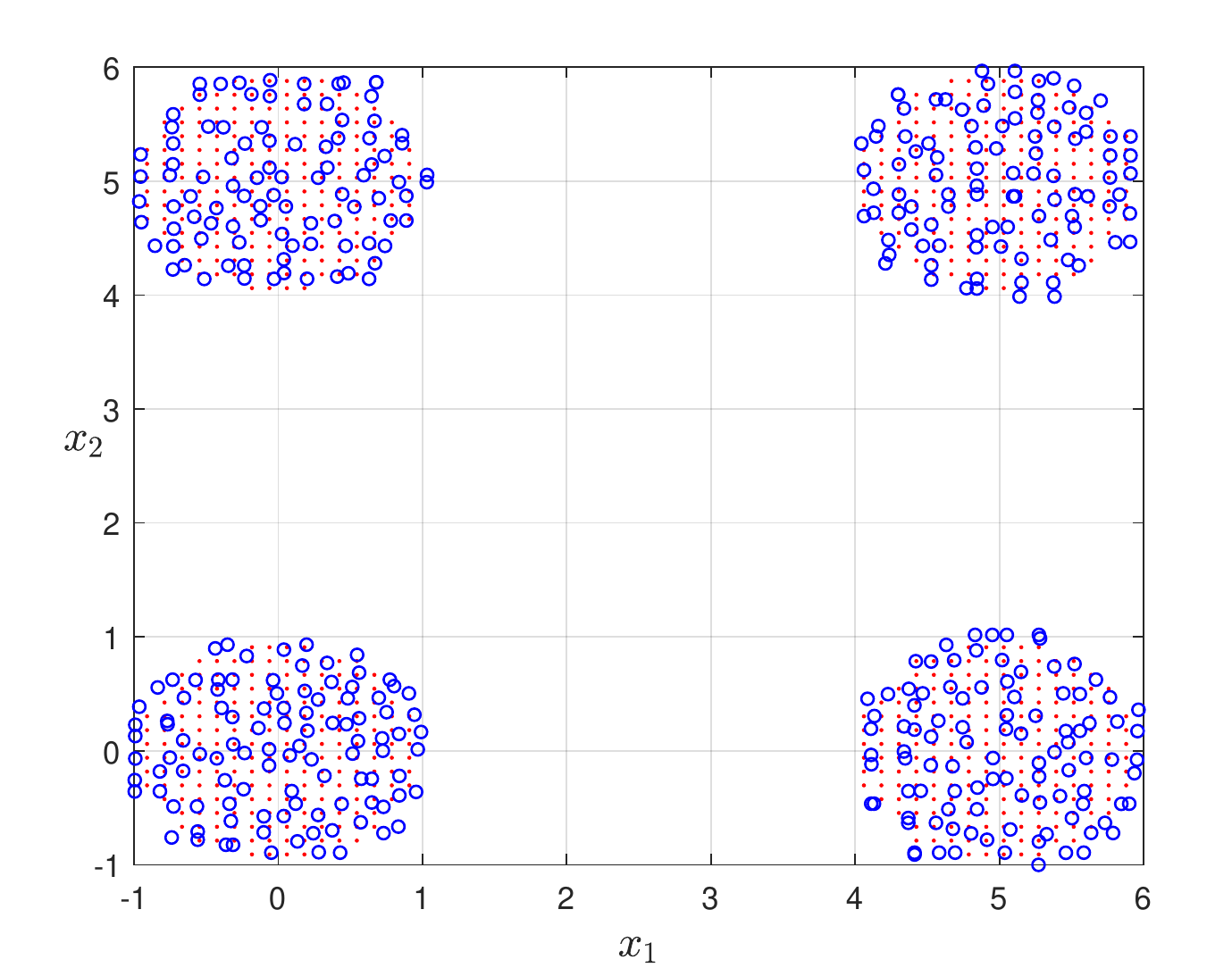}}
\subfigure[MP-MMEA]{\includegraphics[width=1.1in]{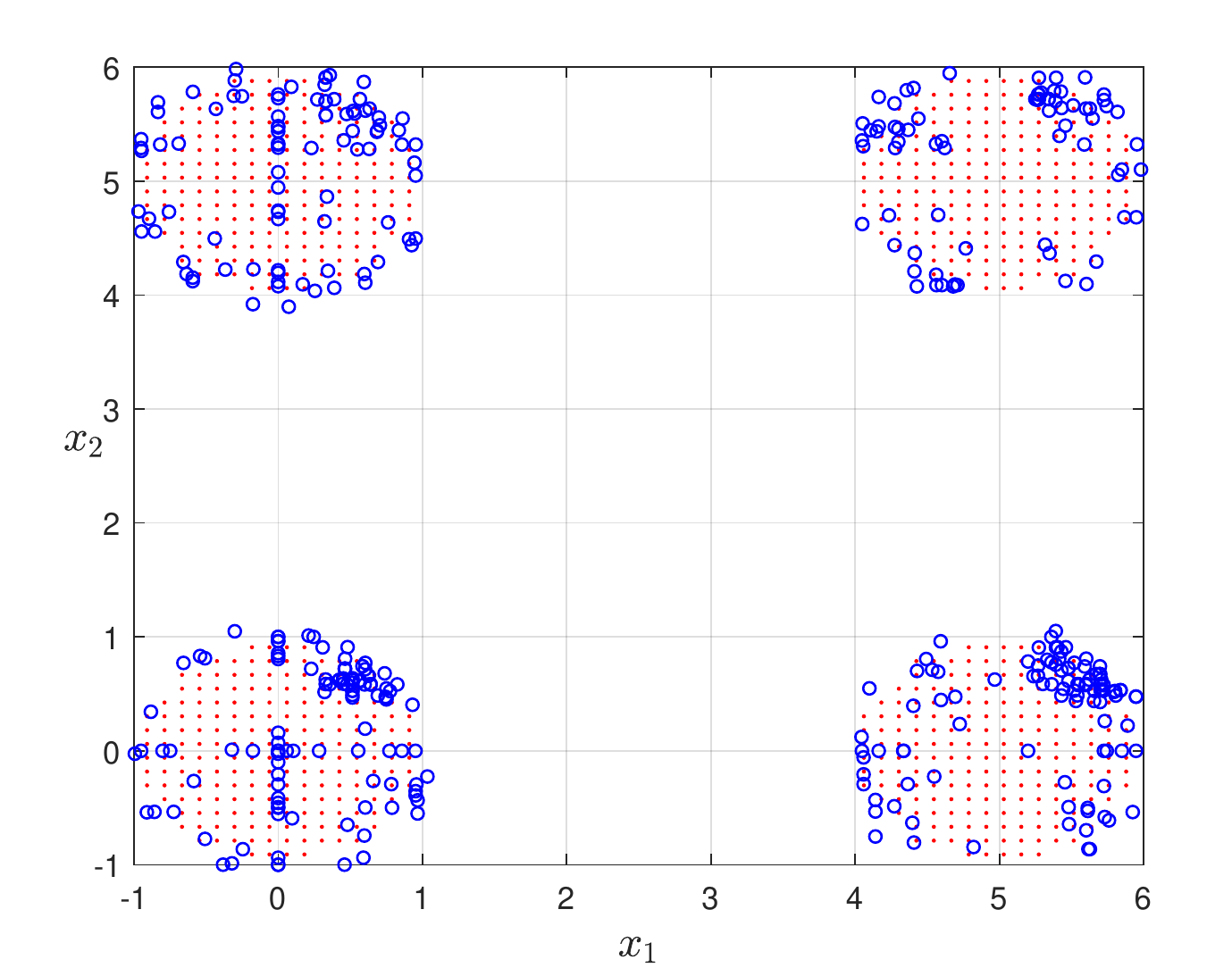}}
\subfigure[MMOEA/DC]{\includegraphics[width=1.1in]{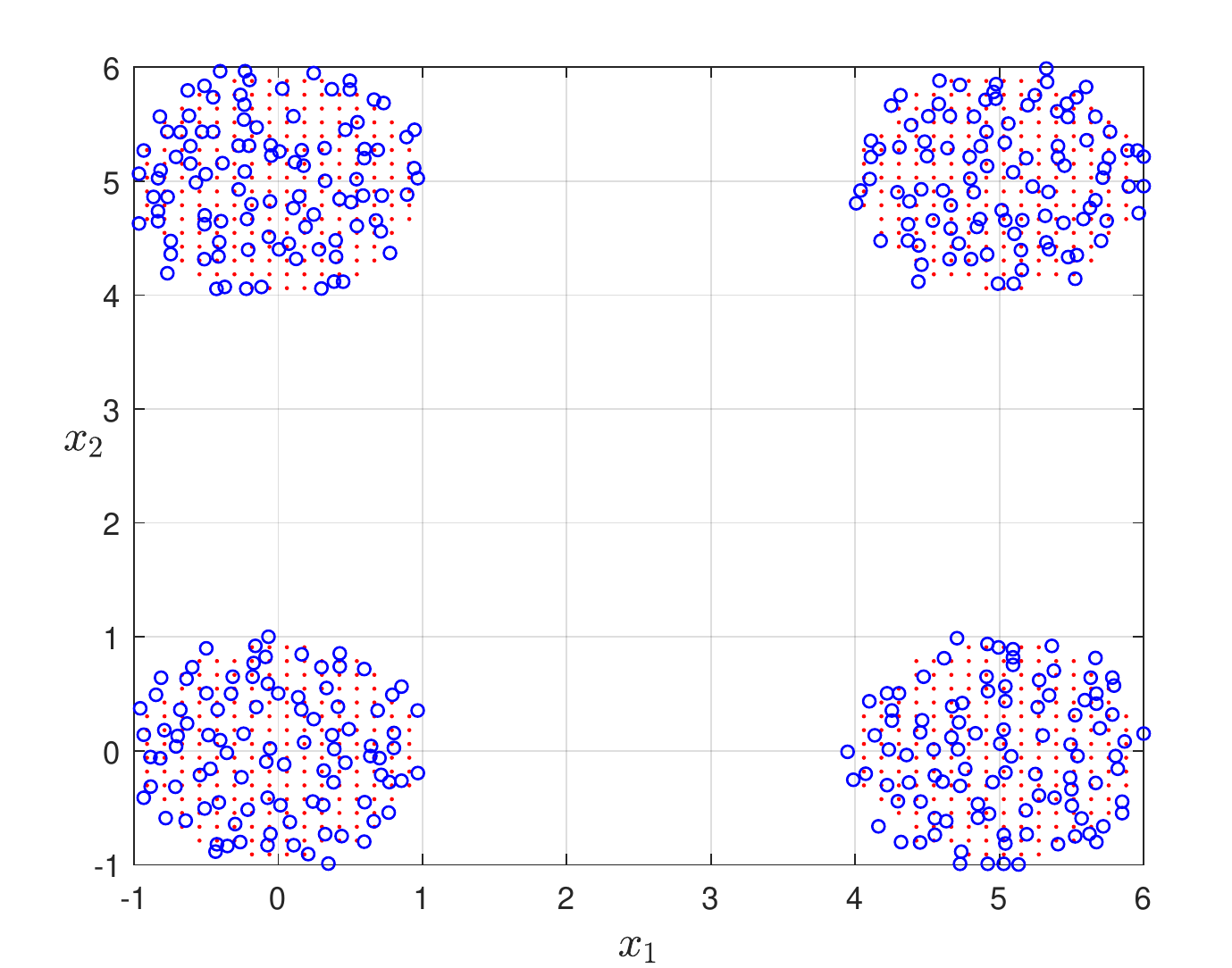}}
\subfigure[MMEA-WI]{\includegraphics[width=1.1in]{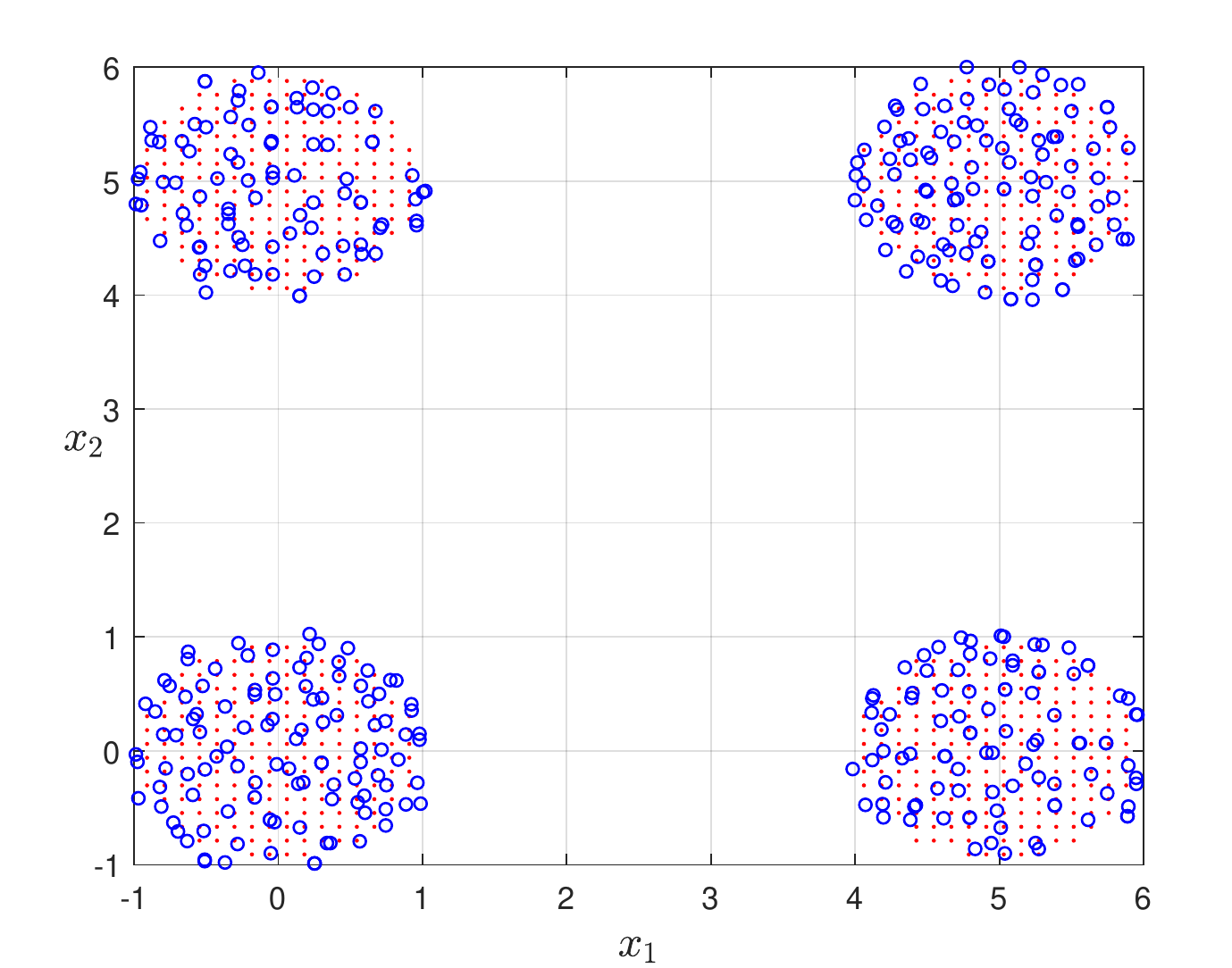}}
\subfigure[HREA]{\includegraphics[width=1.1in]{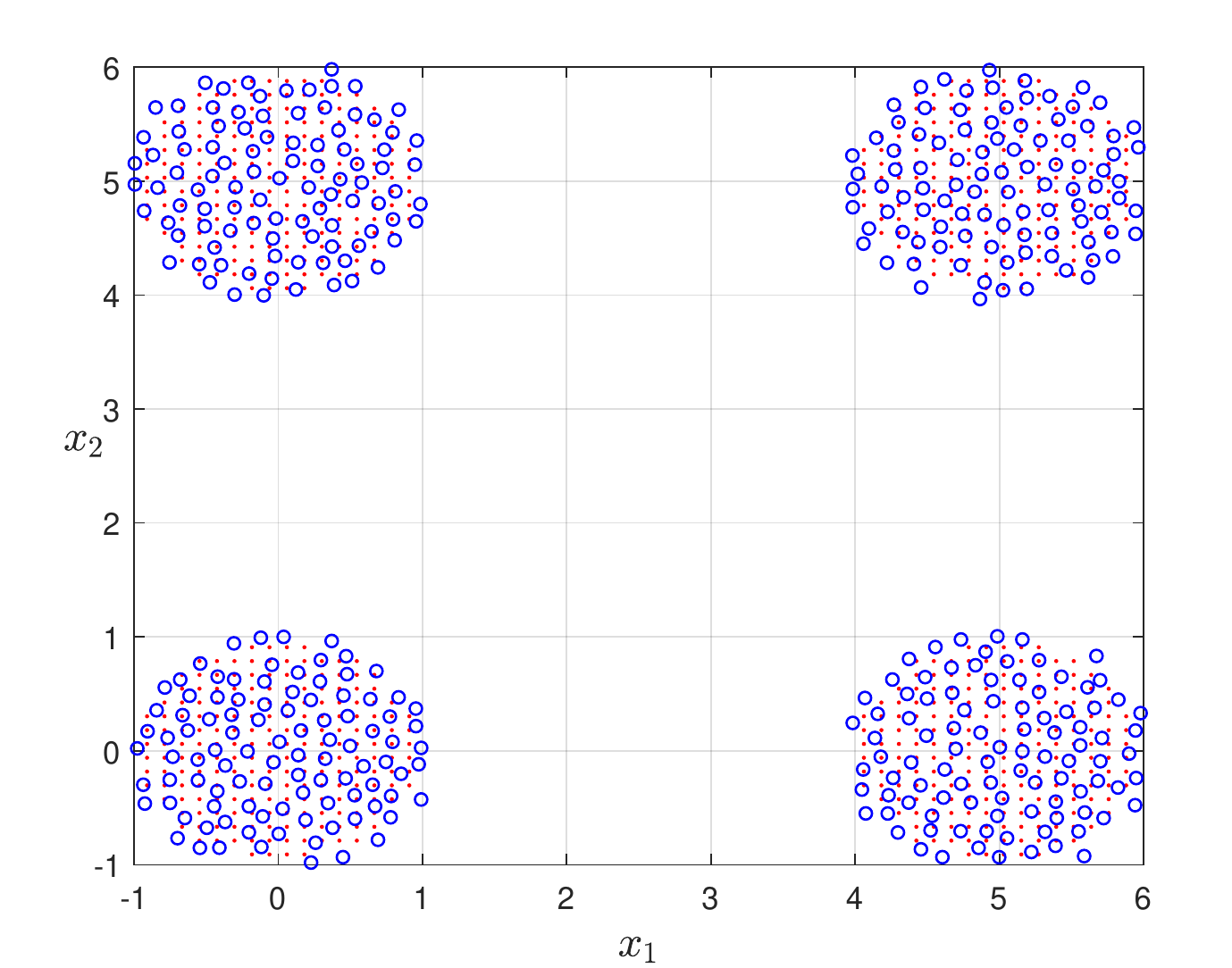}}

\subfigure[Omni-optimizer]{\includegraphics[width=1.1in]{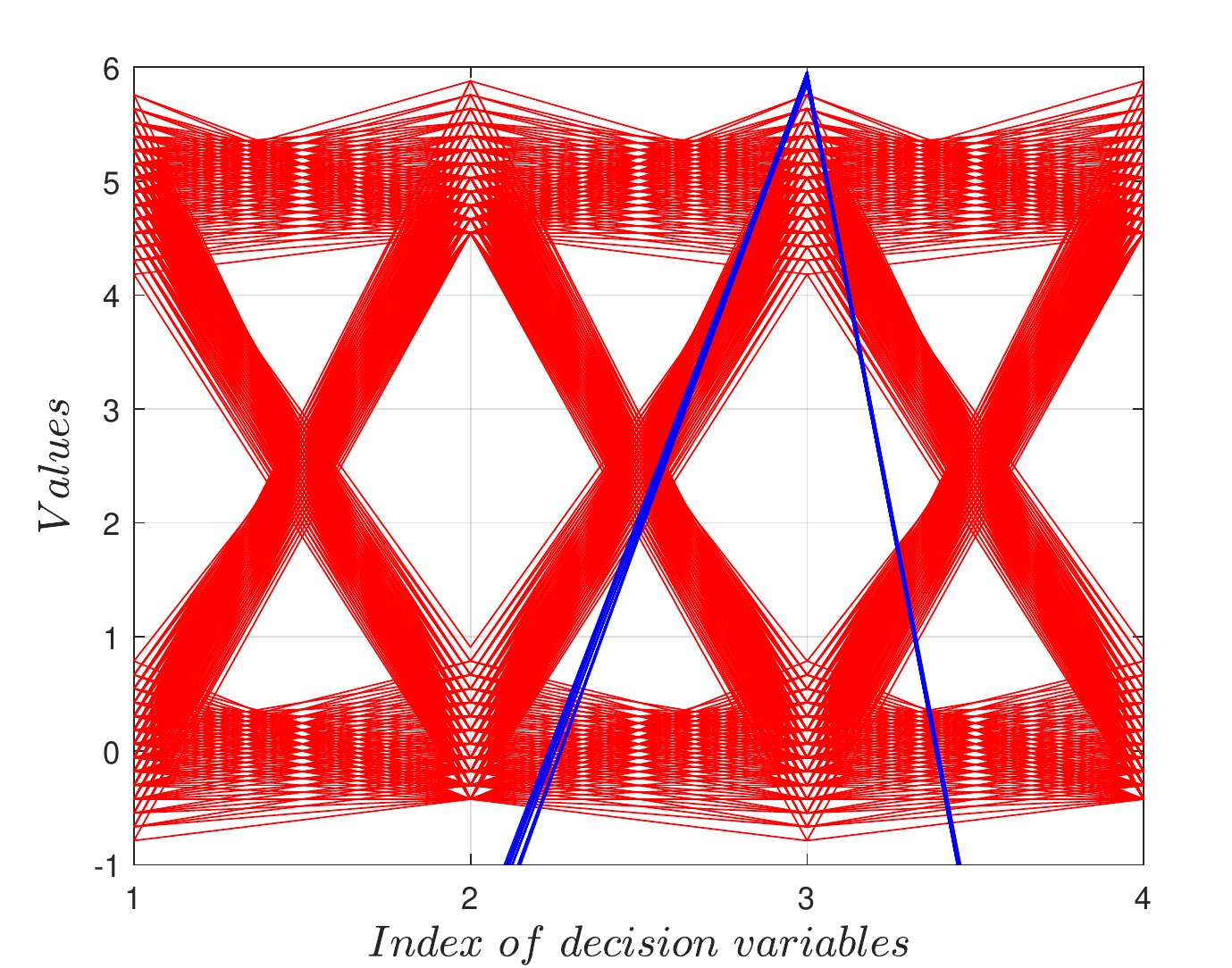}}
\subfigure[DN-NSGAII]{\includegraphics[width=1.1in]{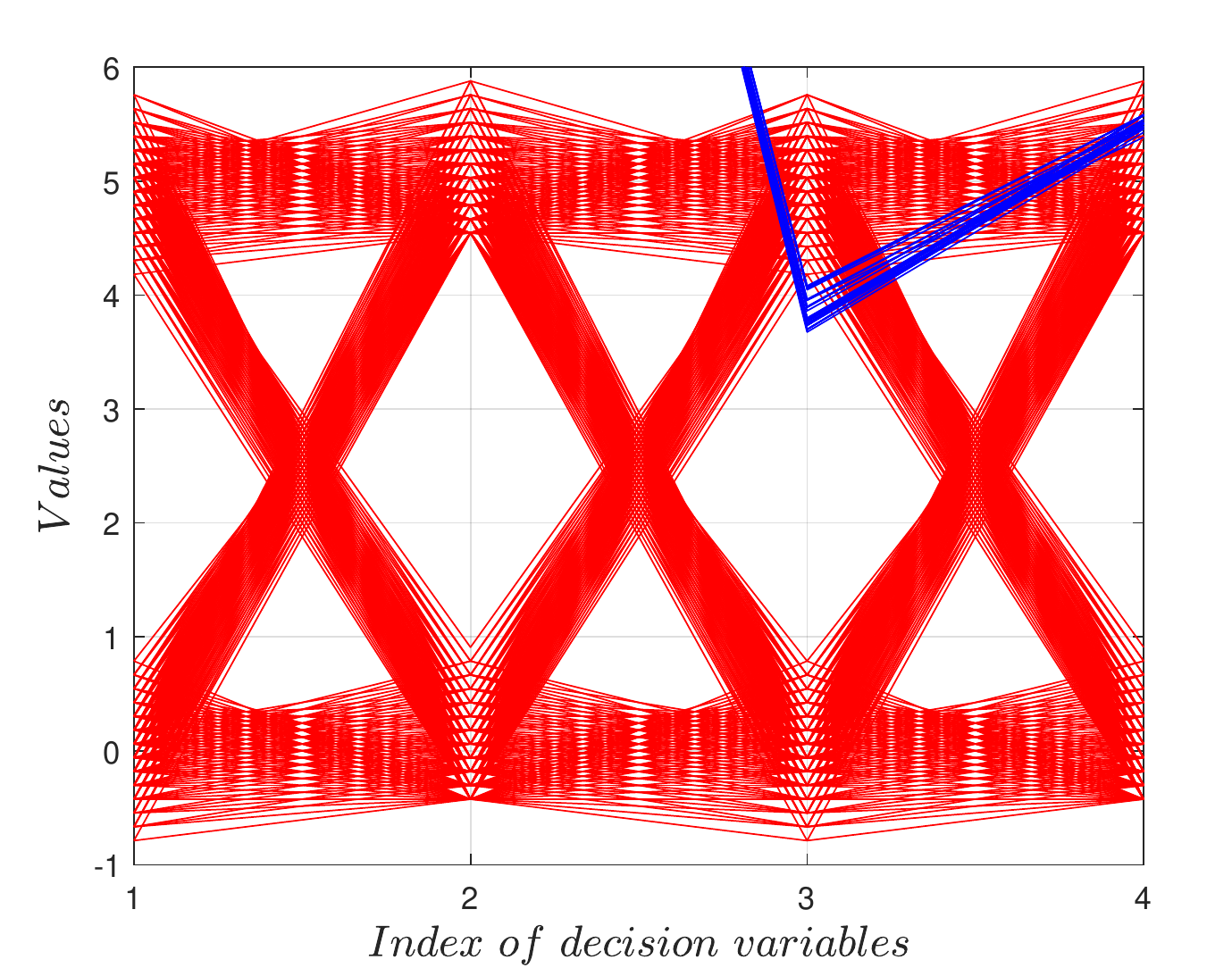}}
\subfigure[MO\_R\_PSO\_SCD]{\includegraphics[width=1.1in]{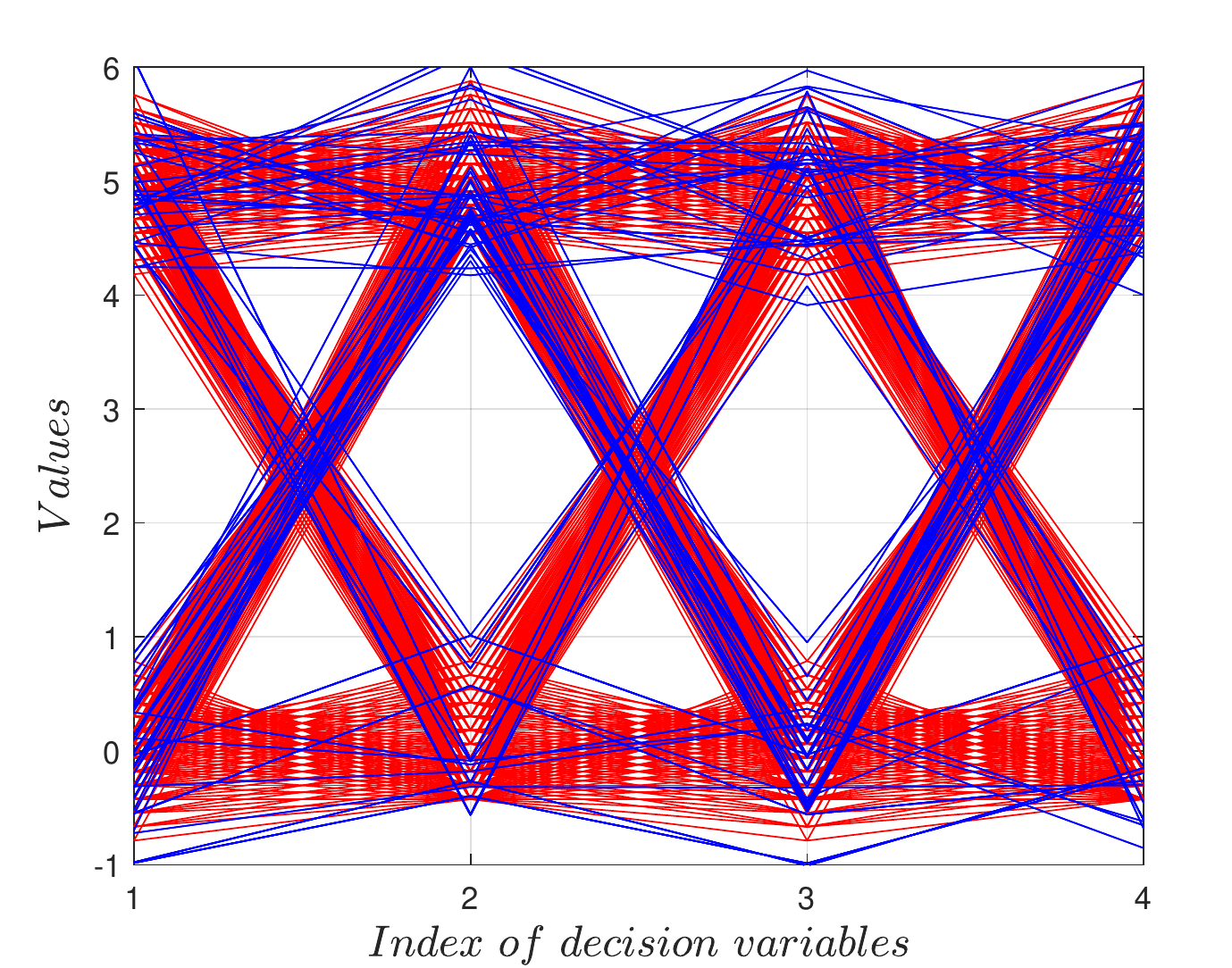}}
\subfigure[MO\_PSO\_MM]{\includegraphics[width=1.1in]{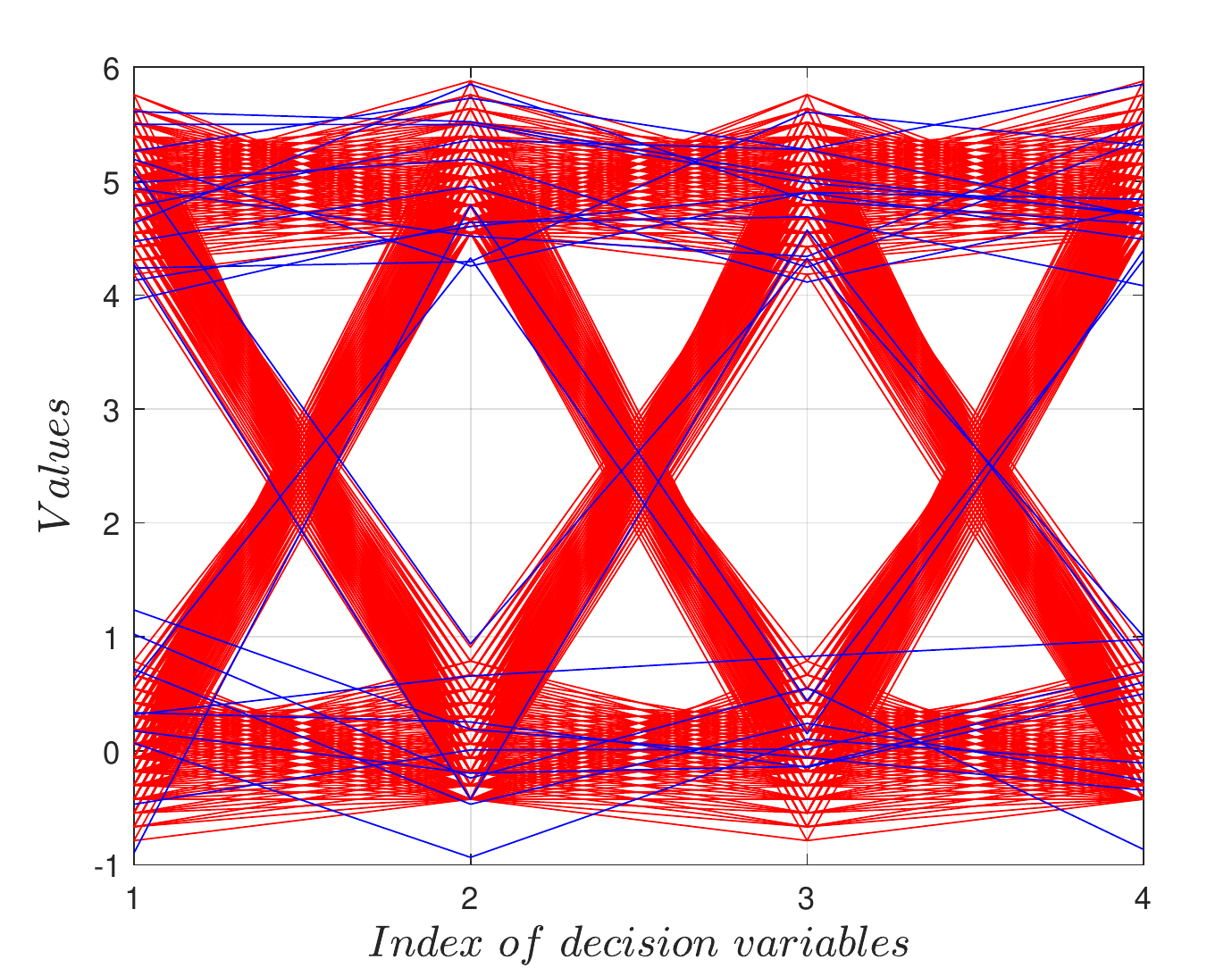}}
\subfigure[DNEA]{\includegraphics[width=1.1in]{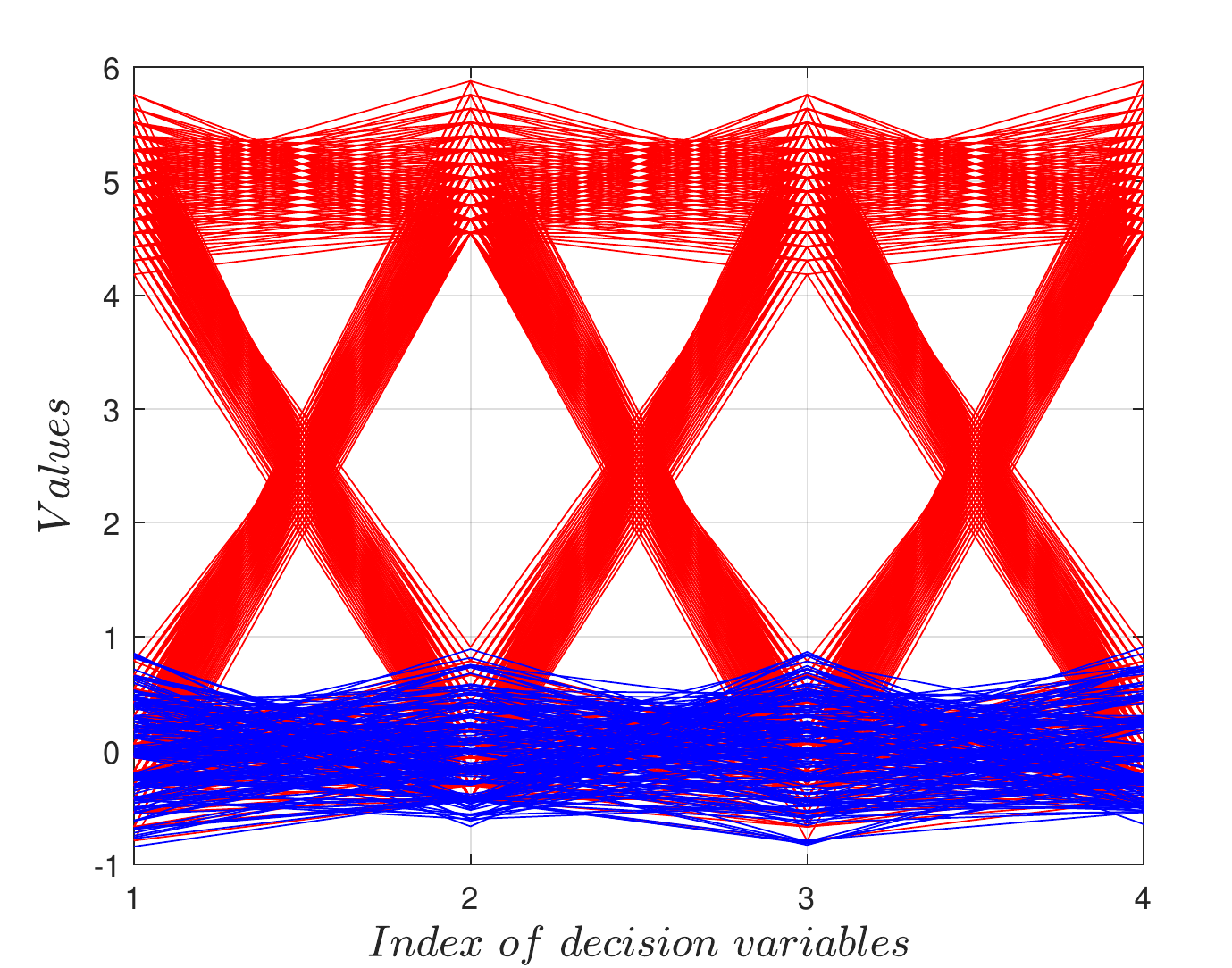}}
\subfigure[Tri-MOEA\&TAR]{\includegraphics[width=1.1in]{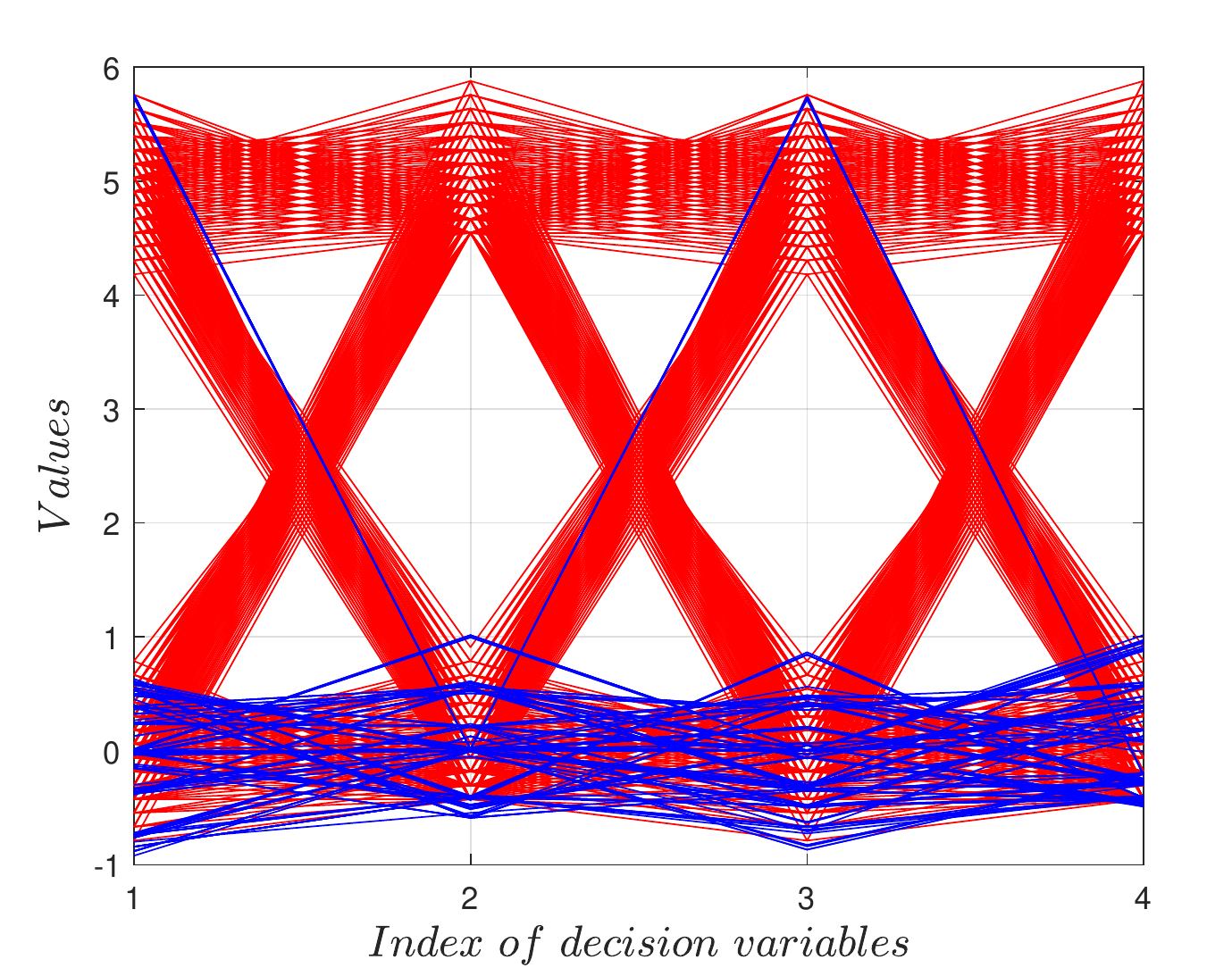}}
\subfigure[DNEA-L]{\includegraphics[width=1.1in]{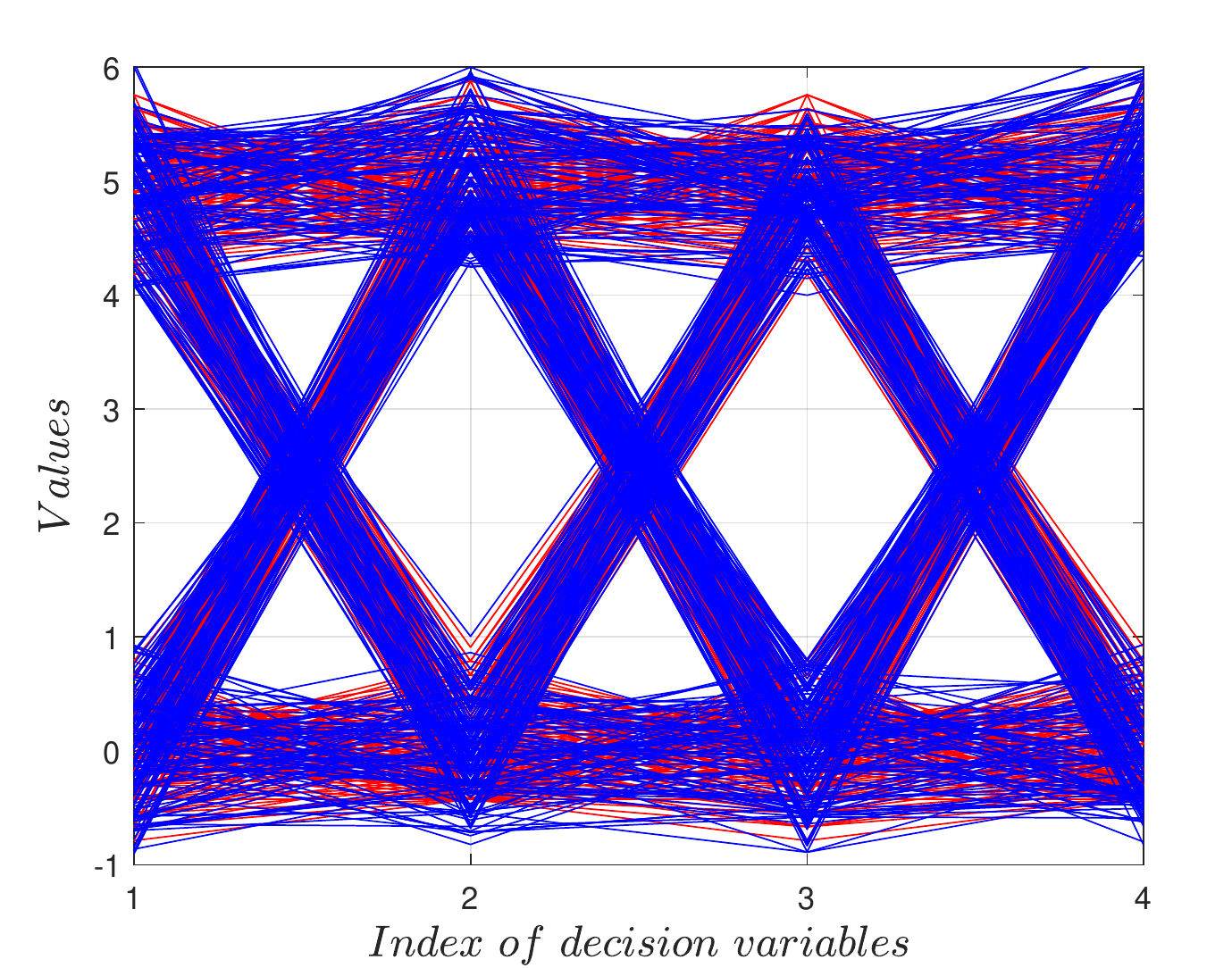}}
\subfigure[CPDEA]{\includegraphics[width=1.1in]{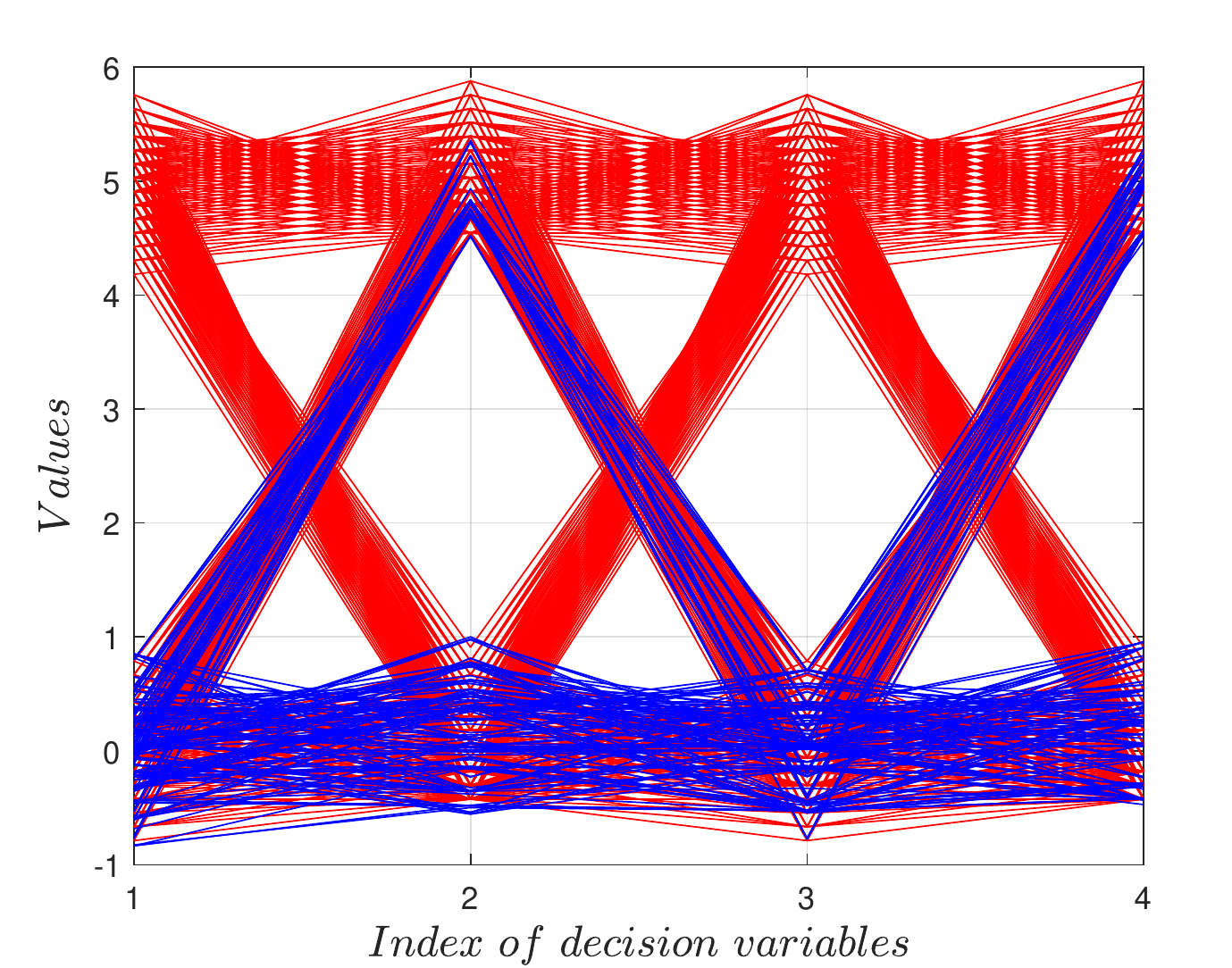}}
\subfigure[MP-MMEA]{\includegraphics[width=1.1in]{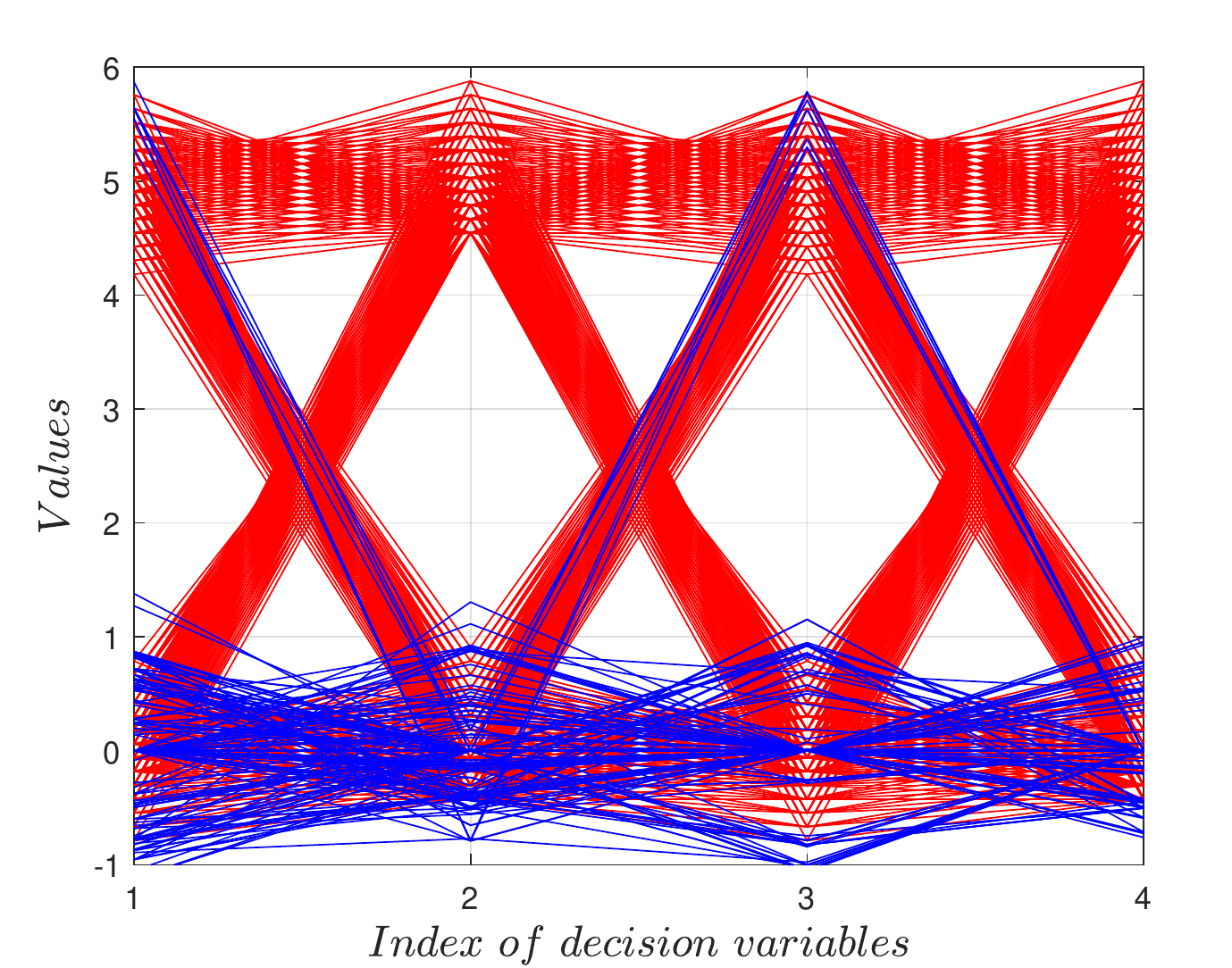}}
\subfigure[MMOEA/DC]{\includegraphics[width=1.1in]{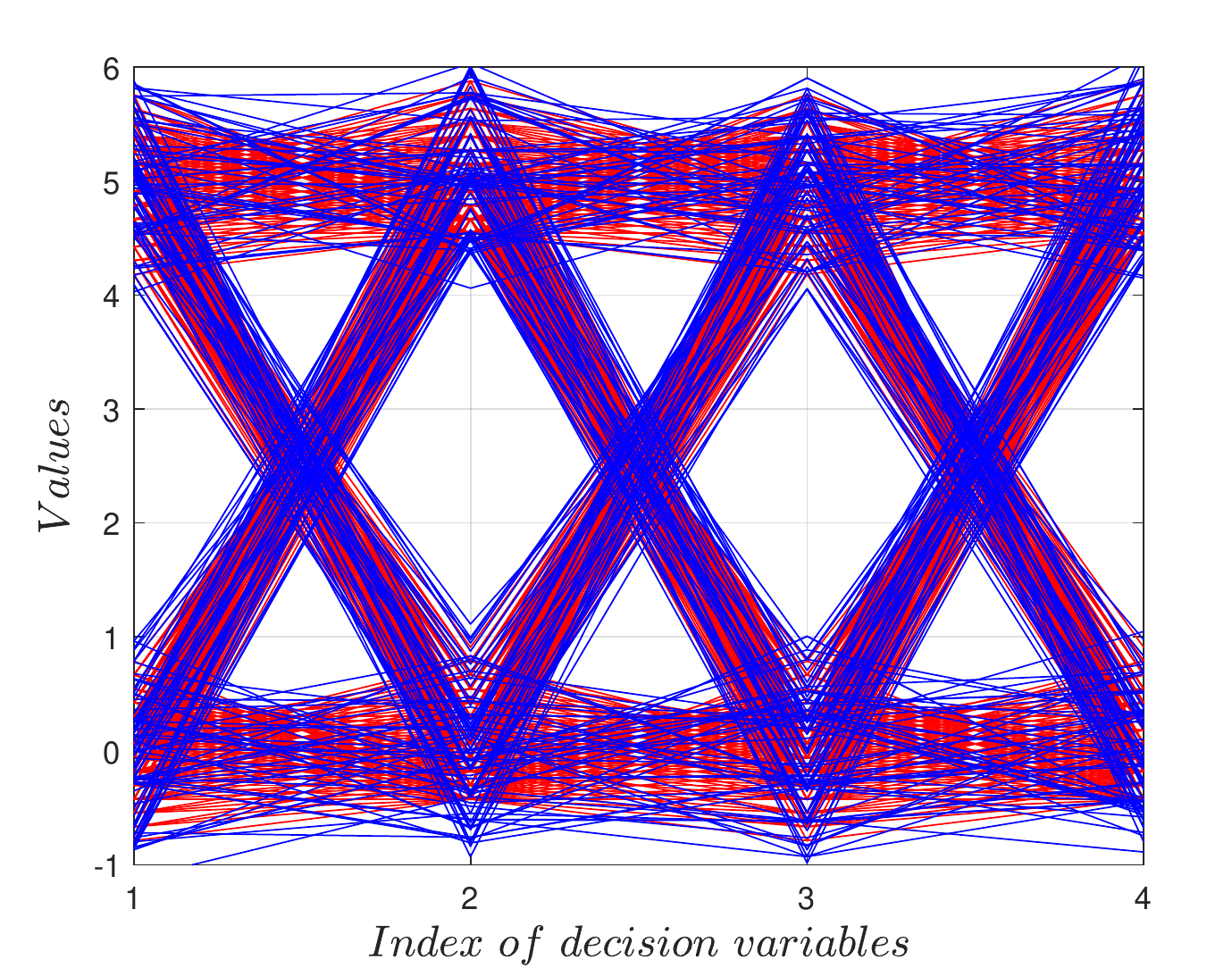}}
\subfigure[MMEA-WI]{\includegraphics[width=1.1in]{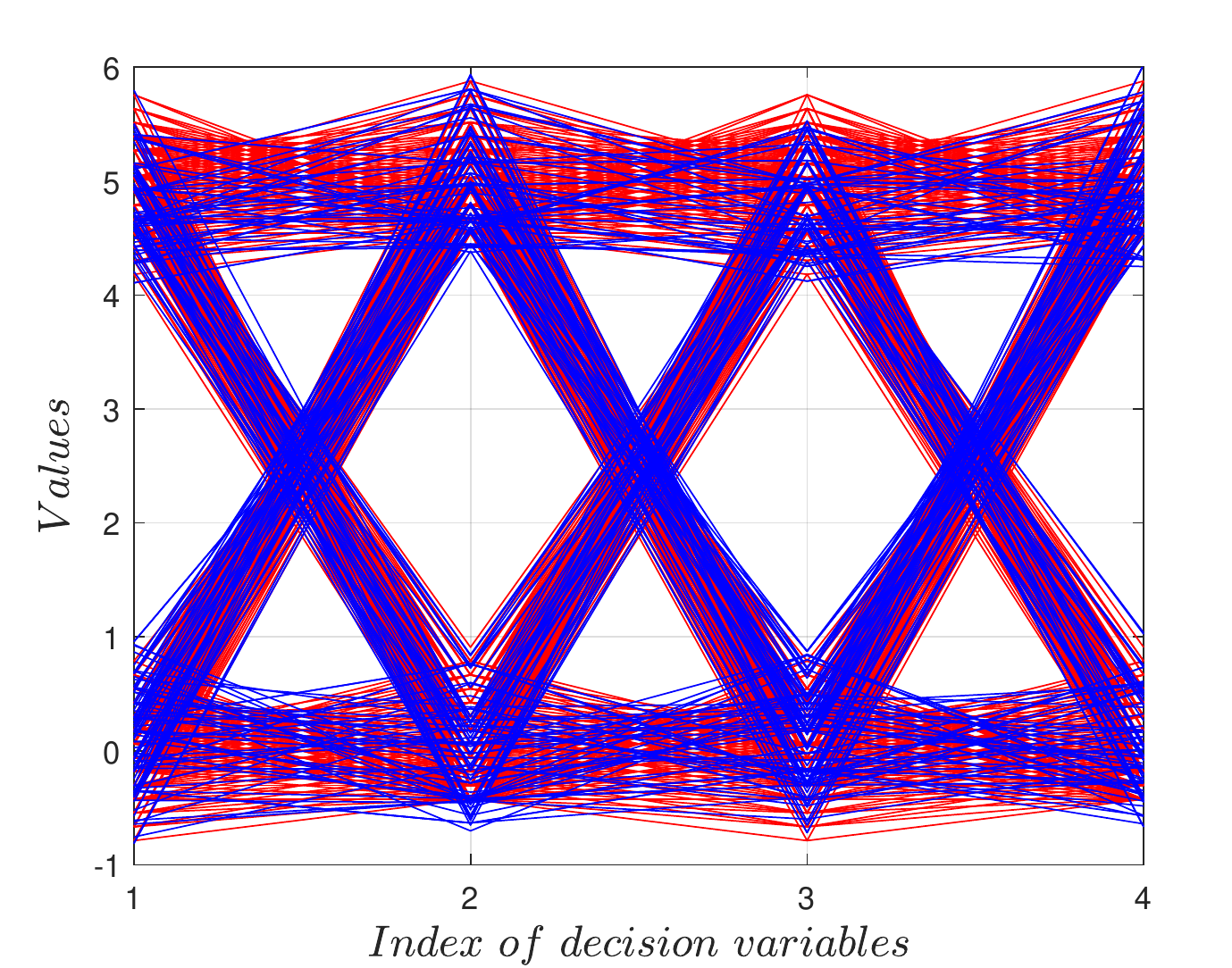}}
\subfigure[HREA]{\includegraphics[width=1.1in]{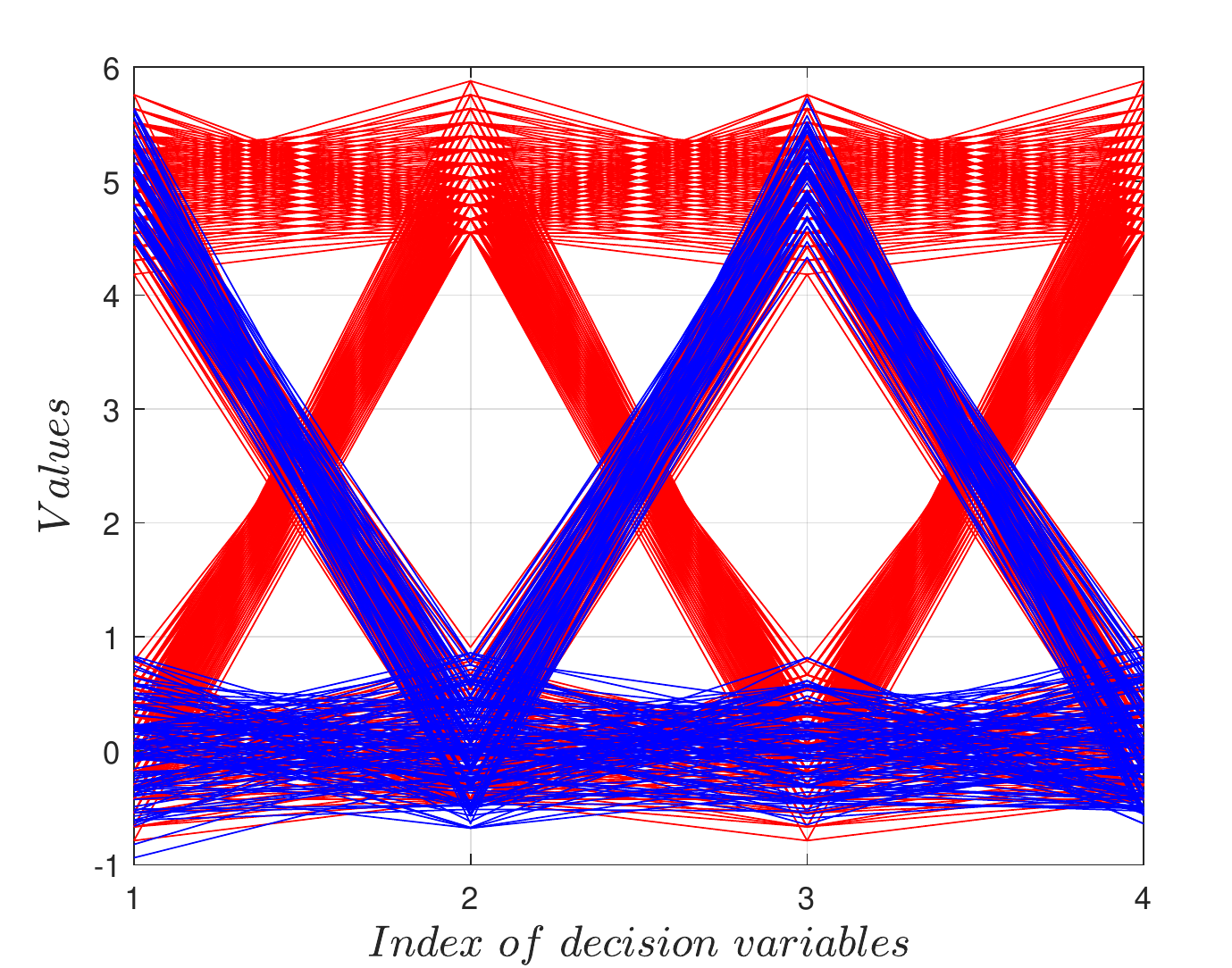}}

\caption{The distribution of solutions obtained by all algorithms (MO\_R\_PSO\_SCD is the short name for MO\_Ring\_PSO\_SCD) in the decision space on Multi-polygon problems (the first two rows are the problem with 10 objectives and 2 decision variables, the last two rows are the problem with 3 objectives and 4 decision variables).}
\label{fig_polygonresult}
\end{figure*}

The first two rows and the last two rows of \fref{fig_polygonresult} present the obtained solutions for Multi-polygon with $M=10,D=2$ and $M=3,D=4$, respectively. For the first two rows, almost all MMEAs can obtain well-distributed solutions except Omni-optimizer, DN-NSGAII and Tri-MOEA\&TAR. That is, the existing MMEAs can well handle low-dimension and many-objective problems. For the last two rows, we can find that MO\_Ring\_PSO\_SCD, MO\_PSO\_MM, DNEA-L, MMOEA/DC and MMEA-WI can obtain all PSs. However, for Multi-polygon with more than 10 decision variables, all MMEAs fail to obtain all PSs. It means that the existing MMEAs show poor performance when the number of decision variables is large. Although MMEA-WI performs best on high-dimension problems, it can not obtain all PSs. In the primitive studies of MMOP, the test suites and MMEAs needed to consider intuitively presenting the effect of obtaining several different PSs. Therefore, almost all test suites are designed with 2 or 3 decision variables. The experimental results tell us that existing MMEAs may face great challenges in dealing with high-dimension problems. Thus, an MMOP test suite with intuitively different PSs for many decision variables is needed.

\section{Further discussions}
\label{sec_discuss}
\subsection{Overall performance of all algorithms}

\begin{figure}[tbph]
	\begin{center}
		\includegraphics[width=3.5in]{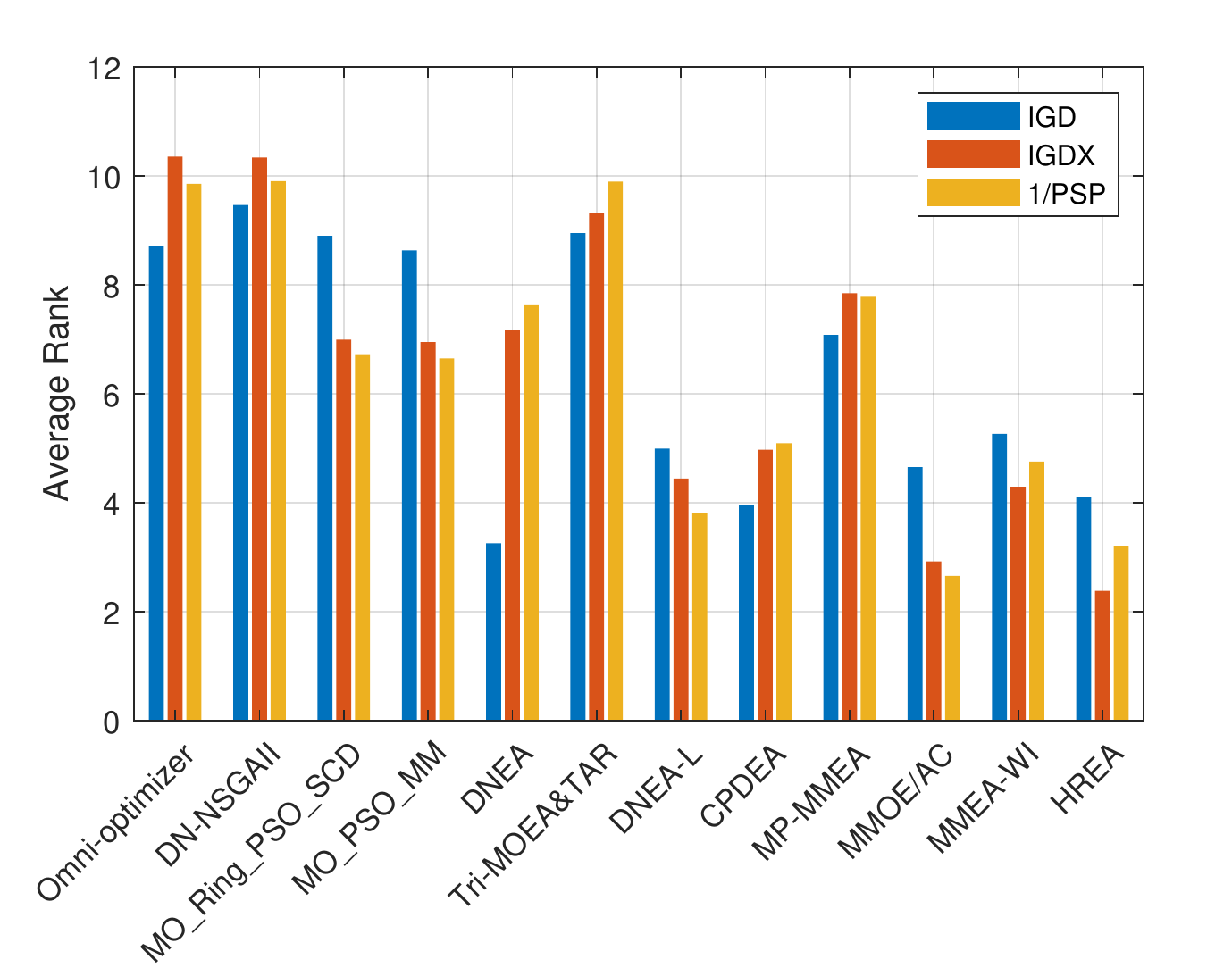}
		\caption{The average rank of all compared MMEAs on all test problems (53 problems in total) in terms of $IGD$, $IGDX$ and $1/PSP$.}
		\label{fig_rankoverall}
	\end{center}
\end{figure}

In the previous parts, we have discussed the performance of all compared MMEAs in solving different types of MMOPs in detail. \fref{fig_rankoverall} presents the overall performance comparison of all selected MMEAs on all test problems.
 
As indicated in \fref{fig_rankoverall}, HREA and MMOEA/DC are two competitive MMEAs for now, which rank 2.38 and 2.66 in terms of $IGDX$ and $1/PSP$ respectively. In the second echelon, DNEA-L, CPDEA and MMEA-WI shine on some test problems, e.g., CPDEA performs best on MMF and MMEA-WI performs best on Multi-polygon. In the third echelon, MO\_Ring\_PSO\_SCD, MO\_PSO\_MM and DNEA can obtain all PSs for primitive proposed test suites like MMF but show poor performance on complex test suites like IDMP. Omni-optimizer, DN-NSGAII, Tri-MOEA\&TAR and MP-MMEA are four algorithms that receive the worst results. To be specific, Omni-optimizer and DN-NSGAII are representative MMEAs in the early stage. They made a positive exploration of the MMOP community and motivated many later studies although their diversity-maintaining strategies are simple. Tri-MOEA\&TAR and MP-MMEA are proposed for special problems, e.g., Tri-MOEA\&TAR used the decision variable analysis method to better solve the MMMOP test suite and MP-MMEA is designed for large-scale sparse problems. Therefore, although they show poor performance on the selected test problems, they can well solve certain MMOPs. It's worth mentioning that, there is no algorithm that performs the best on all test suites. Most of them are verified through some of the proposed benchmarks. Thus, designing a robust algorithm is still the aim of the MMOP community.

In addition, many MMEAs have discussed their computational complexity theoretically and empirically. However, the overall comparison has not been made. In this part, we make an experimental computational complexity comparison. To be specific, the average run time over 30 times independent runs is collected. To better analyze the run time over the different number of objectives and the different number of decision variables, Multi-polygon problems are selected as the benchmark. It's worth mentioning that, we set $N=200$ and $FEs=20000$ for all experiments in this part.

\begin{figure*}[tbph]
	\begin{center}
		\includegraphics[width=7in]{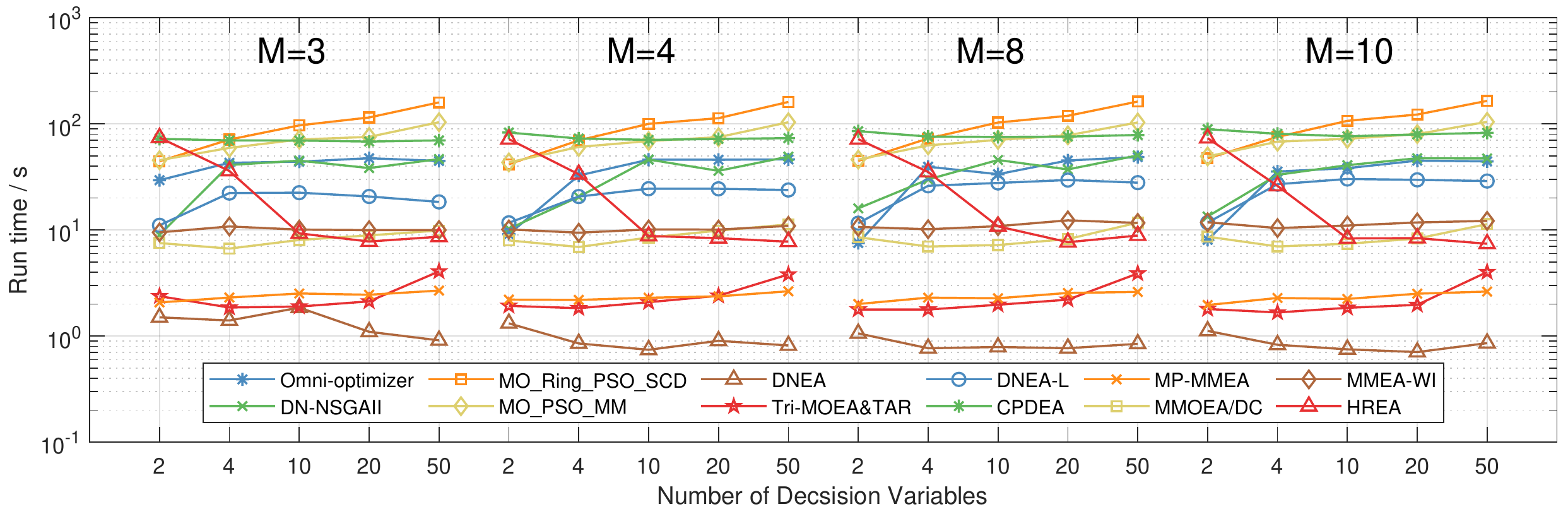}
		\caption{The average running time of all algorithms on Multi-polygon problems with different number of objectives and decision variables.}
		\label{fig_runtime}
	\end{center}
\end{figure*}

\fref{fig_runtime} and Table S-XV show the average running time of all algorithms on Multi-polygon problems with the different number of decision variables and objectives. As we can observe, in terms of computational complexity, there is no significant difference between the different number of objectives. That is, almost all of the MMEAs are not sensitive to the number of objectives. As for the effect of decision variable number, CPDEA, MP-MMEA, MMOEA/DC and MMEA-WI are not sensitive. As the decision variable number increases, the running time changes slightly. For CPDEA, the utilization of a one-by-one strategy needs to calculate the mating-selection and environmental-selection criteria for every single choice. Therefore, its running time is large. As a comparison, an apparent increase occurs in the running time of Omni-optimizer, DNEA-L, MO\_Ring\_PSO\_SCD and MO\_PSO\_MM as the decision variable number increases. An interesting thing is that, for HREA, the running time decreases drastically when the decision variable number increases. For HREA, when the number of decision variables is larger than 4, the size of the convergence archive remains small in the beginning stage since there are few non-dominated solutions. As a result, the running time is relatively small for problems with many decision variables. The same situation can be observed in DNEA. 

To sum up, the running time of all MMEAs is not sensitive to the number of objectives while most of the MMEAs show intuitively time increases when the number of decision variables increases. DNEA, Tri-MOEA\&TAR and MP-MMEA are time-saving in dealing with MMOPs, while MO\_Ring\_PSO\_SCD, CPDEA and MO\_PSO\_MM are the top three time-consuming algorithms. There is a 30 to 100 times difference in terms of running time for the compared MMEAs.

\subsection{Open issues and future study}
1) Through the analysis of the experimental results, we can see that most of the existing MMEAs can obtain a good result on normal MMOPs except for IDMP and MMOPLs. Considering local PSs can significantly enhance the algorithms' ability on maintaining the diversity, e.g, DNEA-L, MMOEA/DC and HREA. However, there are few works considering obtaining local PSs for now. The IEEE CEC 2019 multimodal optimization competition made some efforts on raising the attention of MMOPLs. After that, MMOEA/DC and HREA were proposed. We think that developing algorithms to obtain the local PSs is a more practical and general direction in the MMO community. 

2) Another important issue is that there lacks a multi-modality detection method for a given problem. That is, for DMs to deal with a certain real-world problem, there is no information on whether this problem is an MMOP or not. According to some previous works, the convergence ability of the MMEAs is worse than the state-of-the-art MOEAs. Thus, MMEAs will not be the first choice. Developing an effective and efficient tool or method to detect the multi-modality of an MOP is significantly important and urgent.

3) As we discussed in \sref{sec_testsuite}, limitations exist for the proposed benchmarks. There are no discrete optimization test problems for now. The decision variables for real-world problems usually contain several kinds, e.g., continuous, discrete, and binary. Although it's easy to transform the existing test suites into discrete optimization problems, there is no work that systematically analyses the performance of the existing MMEAs. 

4) In addition, as results in \sref{sec_polygonresult} show, the existing MMEAs face huge challenges in solving MMOPs with many decision variables. Many proposed MMOPs are relatively simple to be solved, e.g., MMF1-8. An important reason is that multiple PSs can not be directly observed for multi-dimension problems. Therefore, the searching ability and efficiency can not be well evaluated by the existing test suites. Moreover, the drawbacks of utilizing the diversity-maintaining technique as the first-selection strategy have not been well studied yet. A comprehensively MMOP test suite that has difficulty in searching PF is needed for the MMOP community.

\begin{figure}[tbph]
	\begin{center}
		\includegraphics[width=3.5in]{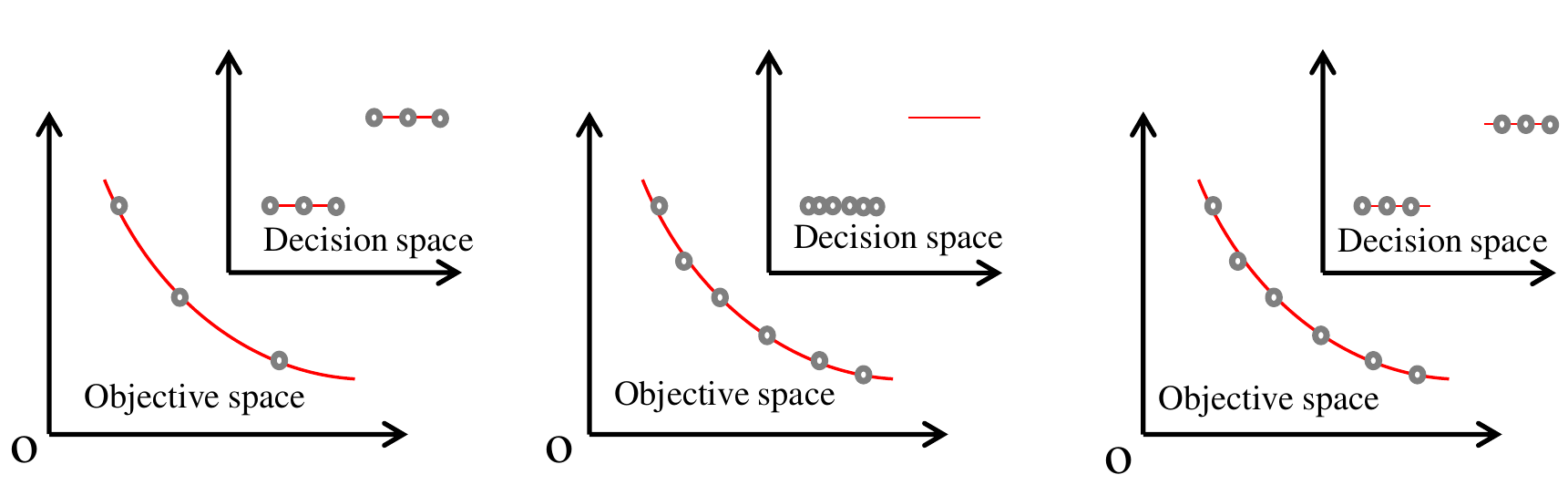}
		\caption{Illustration of solution distributions in the objective and decision spaces, where the red line and gray circles indicate the true PF(PS) and the solutions respectively. Notably, two different PSs correspond to the same PF.}
		\label{fig_crowddis}
	\end{center}
\end{figure}

5) As for the performance metrics, there lacks a comprehensive indicator that can measure both diversity and convergence in the objective and decision spaces, which are both important for solving an MMOP \cite{tanabe2019review,lin2020multimodal,li2022hierarchy}. \fref{fig_crowddis} presents the relationship between the diversity in the objective space and the diversity in the decision space of a two-objective two-decision-variable MMOP, where two different PSs correspond to the same PF. The left figure indicates the situation for primitive MMEAs, where the distribution in the objective is not considered. The middle figure presents the distribution of solutions for normal MOEAs, which only considers the objective space. The aim of MMEAs is shown in the right figure, which takes two space diversity into account. However, most of the existing performance metrics for MMEAs only consider solution quality in the decision space. $IGDM$ made some efforts in comprehensively evaluating the quality. Intuitively, to overall evaluate the quality of the two spaces, weights should be given, which is hard for the DMs to balance. In addition, true PS and PF data are necessary for the existing performance metrics, which is hard for real-world problems.

\section{Conclusion}
\label{sec_conclusion}
Multimodal multi-objective optimization problems are common in the real world and receive more and more attention. In this work, we first made a review of the proposed MMOP test suites and discussed their properties. Then, we introduced several state-of-the-art MMEAs with different diversity-maintaining techniques. Next, we comprehensively compared the performance of 12 popular MMEAs on the chosen benchmark problems. Our experimental results indicate that there is no algorithm that performs more overwhelmingly than all other compared MMEAs on all test suites. However, considering to obtain local PSs is apparently effective in dealing with most of the MMOPs.

Many multi-objective real-world optimization problems show multi-modality characteristic, e.g., configuration of hybrid renewable energy system \cite{li2021sizing}, job-shop scheduling problem \cite{li2019knee}, satellite mission planning problems \cite{gabrel2003mathematical}, etc. However, few of them choose to utilize MMEAs to obtain multiple solutions. For problems with or without multi-modality, obtaining all global and local optimal solutions can help DMs understand the implicit properties. Therefore, future works include applying MMEAs to more real-world problems. In addition, research on general diversity-maintaining techniques can significantly improve the ability to jump out of the local optimal region. Thus, embedding multimodal techniques into other MOEAs to help enhance the performance could be a useful research topic, e.g., enhancing the existing constraint multi-objective evolutionary algorithms (CMOEAs) \cite{tian2020coevolutionary,liu2019hand,wang2021dual} by improving the diversity of solutions.

\bibliographystyle{IEEEtran}
\bibliography{reference}

\begin{IEEEbiography}[{\includegraphics[width=1in,height=1.25in]{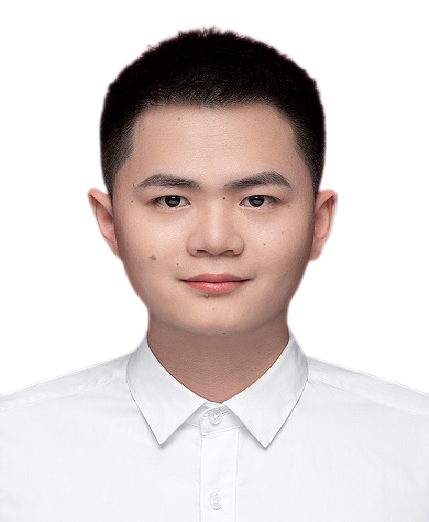}}]{Wenhua Li} received his B.S. and M.S. degrees in 2018 and 2020, respectively, from 	National University of Defense Technology (NUDT), Changsha, China. He is now a Ph.D. student in Management Science and Technology. His current research interests include multi-objective evolutionary algorithms, energy management in microgrids and artificial intelligence.
\end{IEEEbiography}
\vspace{-20 pt}


\begin{IEEEbiography}[{\includegraphics[width=1in,height=1.25in]{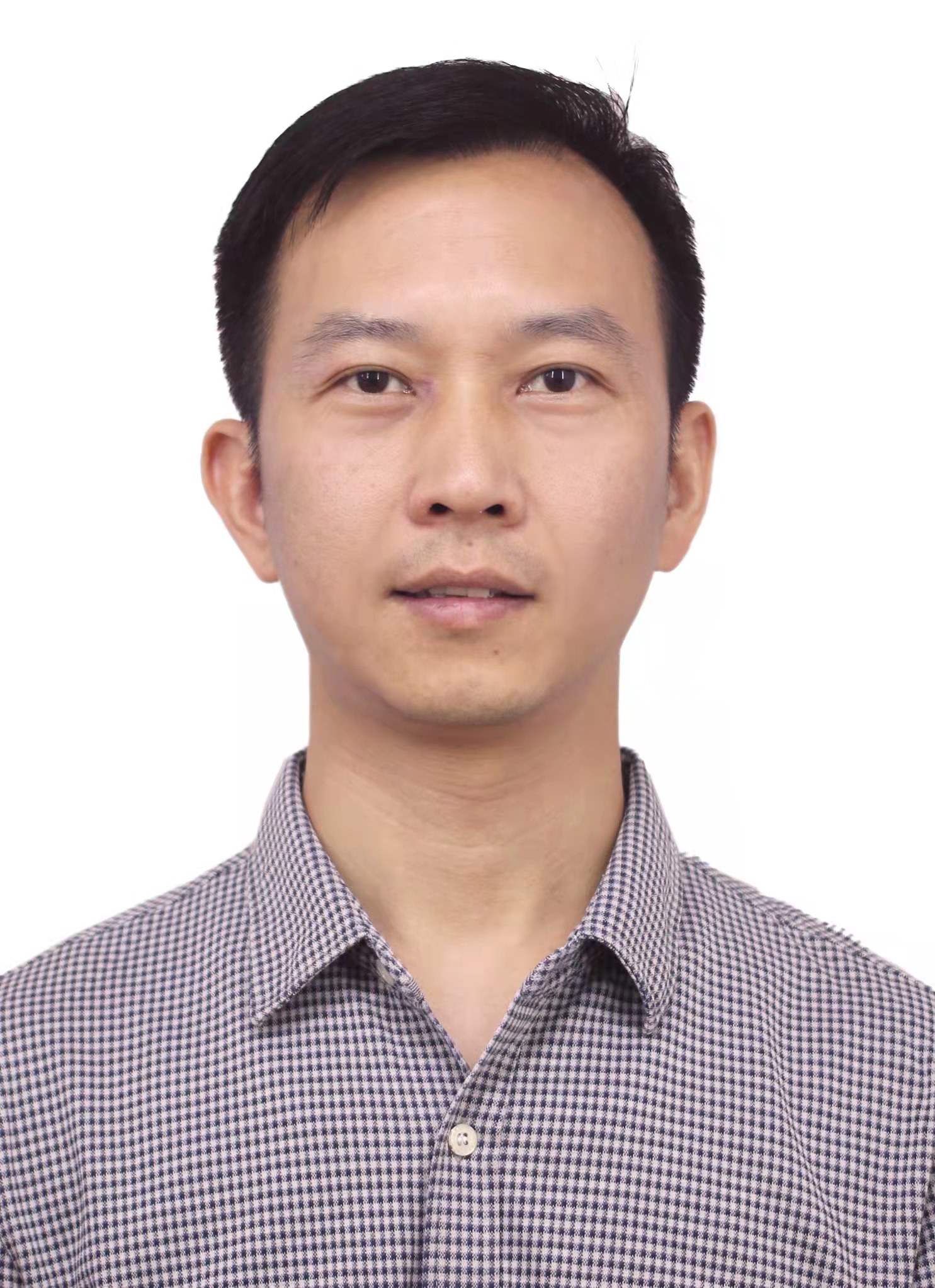}}]{Tao Zhang} received B.S., M.S., and Ph.D. degrees from National University of Defense Technology(NUDT), Changsha, China, in 1998, 2001, and 2004, respectively.

He is a Full Professor with the College of Systems Engineering, NUDT. He is also the Director of the Hunan Key Laboratory of Multi-Energy System Intelligent Interconnection Technology (HKL-MSI2T). His current research interests include optimal scheduling, data mining, and optimization methods on the energy internet. He was the recipient of the Science and Technology Award of Provincial Level (first place in 2020 and 2021, and second place in 2015 and 2018).
\end{IEEEbiography}

\begin{IEEEbiography}[{\includegraphics[width=1in,height=1.25in]{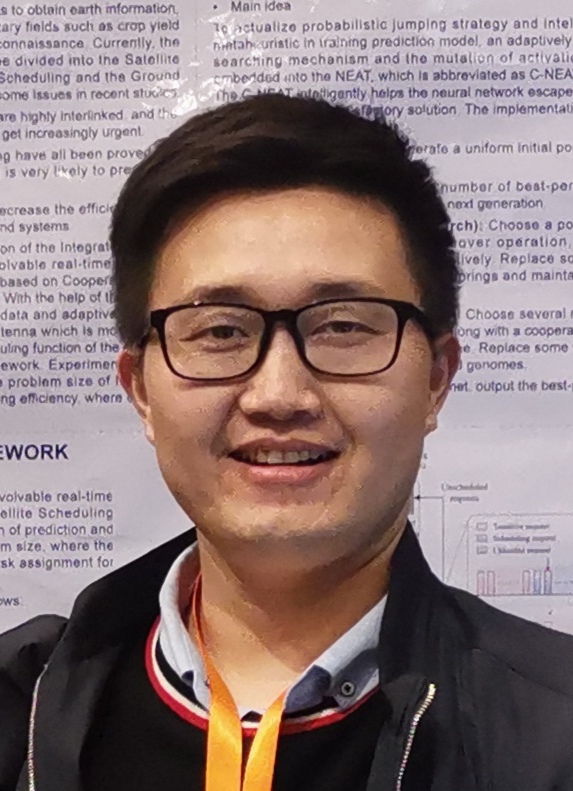}}]{Rui Wang} (Senior Member, IEEE) received his Bachelor degree from the National University of Defense Technology, P.R. China in 2008, and the Doctor degree from the University of Sheffield, U.K in 2013. Currently, he is an Associate professor with the National University of Defense Technology. His current research interest includes evolutionary computation, multi-objective optimization and the development of algorithms applicable in practice.
	
Dr. Wang received the Operational Research Society Ph.D. Prize at 2016, and the National Science Fund for Outstanding Young Scholars at 2021. He is also an Associate Editor of the Swarm and Evolutionary Computation, the IEEE Trans on Evolutionary Computation.
	
\end{IEEEbiography}

\begin{IEEEbiography}[{\includegraphics[width=1in,height=1.25in,clip,keepaspectratio]{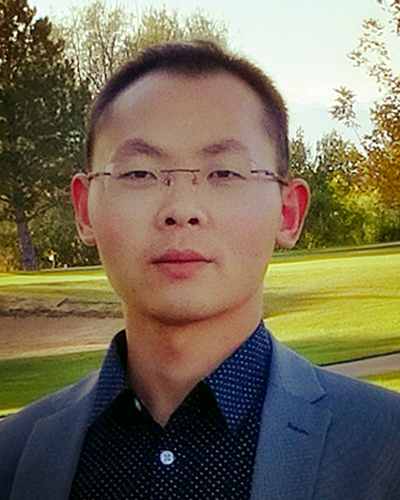}}]{Shengjun Huang} received the B.S. and M.S. degrees in management science and engineering from the National University of Defense Technology, Changsha, China, in 2011 and 2013, and the Ph.D. degree in energy systems from the University of Alberta, Canada, in 2018. He is currently an Associate Professor with the College of Systems Engineering, National University of Defense Technology, Changsha, China. His research interests include mixed-integer linear programming, robust optimization algorithms, large-scale power systems, and microgrid clusters.
\end{IEEEbiography}

\begin{IEEEbiography}[{\includegraphics[width=1in,height=1.25in,clip,keepaspectratio]{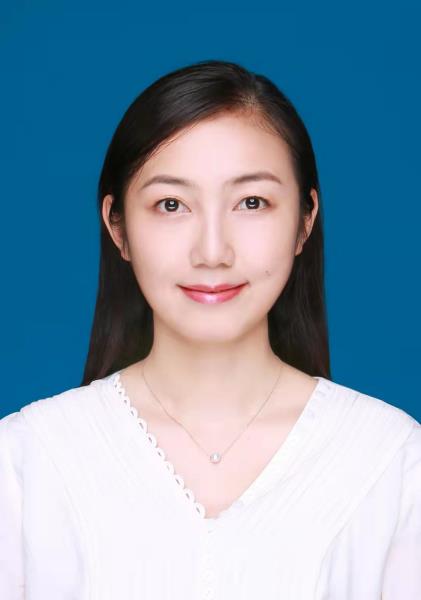}}]{Jing Liang} (Senior Member, IEEE) received the B.E. degree in automation from Harbin Institute of Technology, Harbin, China, in 2003, and the Ph.D. degree in Electrical and Electronic Engineering From Nanyang Technological University, Singapore, in 2009.

She is currently a Professor with the School of Electrical Engineering, Zhengzhou University, Zhengzhou, China. Her main research interests are evolutionary computation, swarm intelligence, multi-objective optimization, and neural network. She currently serves as an Associate Editor for the IEEE Transactions on Evolutionary Computation, and the Swarm and Evolutionary Computation.
\end{IEEEbiography}

\end{document}